\documentclass[runningheads]{llncs}

\usepackage{eccv}
\usepackage{eccvabbrv}

\usepackage{graphicx}
\usepackage{float}
\usepackage{booktabs}
\usepackage{multirow}
\usepackage{wrapfig}
\usepackage{listings}
\usepackage{xcolor}
\usepackage{pgfplots}
\pgfplotsset{compat=1.18}
\usepackage{microtype}
\usepackage[accsupp]{axessibility}
\usepackage{minitoc}
\doparttoc

\definecolor{myblue}{HTML}{3E3EE9}
\definecolor{myorange}{HTML}{FF9000}
\definecolor{mygreen}{HTML}{4B9F2D}
\definecolor{myyellow}{HTML}{D1CA00}
\definecolor{claudebg}{RGB}{244,236,226}
\definecolor{LightGrey}{rgb}{0.92,0.92,0.92}
\definecolor{Myred}{rgb}{1.00,0.12,0.36}
\definecolor{Myblue}{rgb}{0,0.60,0.87}
\usepackage[table]{xcolor}
\newcommand{\mypar}[1]{\paragraph{#1}}

\usepackage[breaklinks,colorlinks,citecolor=eccvblue]{hyperref}
\hypersetup{
  colorlinks=true,
  linkcolor=myblue,
  citecolor=myblue,
  urlcolor=black
}

\usepackage{orcidlink}

\begin{document}

\title{PPTArena: A Benchmark for PowerPoint Editing}

\titlerunning{PPTArena}

\author{
    Michael Ofengenden\inst{1,2}\thanks{Work done during a FoDOMMaT internship at UIUC.}\orcidlink{0009-0007-8418-2831} \and
    Yunze Man\inst{1}\orcidlink{0000-0002-2357-2883} \and
    Ziqi Pang\inst{1}\orcidlink{0009-0007-2846-5618} \and\\
    Liang-Yan Gui\inst{1}\orcidlink{0009-0005-8204-3577} \and
    Yu-Xiong Wang\inst{1}\orcidlink{0000-0003-4414-0198}
}

\authorrunning{M. Ofengenden et al.}

\institute{
University of Illinois Urbana-Champaign \and
University of California, Berkeley
}

\maketitle
\begin{center}
    \captionsetup{type=figure}
    \includegraphics[trim=0 0 0 0, clip=True, width=0.92\textwidth]{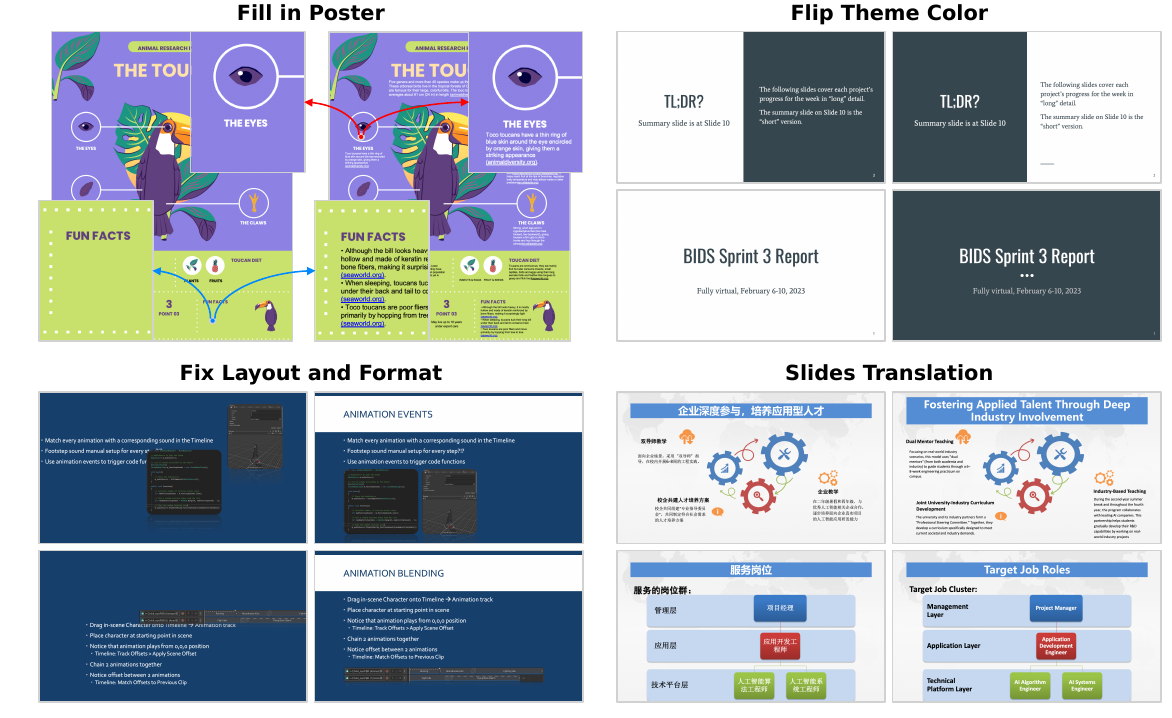}
    \captionof{figure}{Examples showing structure-aware, multimodal, and cross-slide reasoning tasks.}
    \label{fig:teaser}
\end{center}%
\begin{abstract}
We introduce PPTArena, a benchmark for PowerPoint editing that evaluates how agents modify real slides from natural-language instructions. Unlike benchmarks that rely on image-PDF renderings or text-to-slide generation, PPTArena features 100 decks with over 1,300 human-curated edits across 2,125 slides, spanning text, charts, animations, and professional master styles. Each edit pairs a ground-truth deck with a target rubric and is scored by two Vision-Language Model (VLM) judges: one rates instruction following from structural diffs, the other visual quality from slide images. On top of this benchmark, we present PPTPilot, a structure-aware agent that plans semantic edit sequences, routes between programmatic tools and deterministic XML operations, and verifies each result in an iterative plan-edit-check loop. PPTPilot outperforms strong VLM-based agents by more than 10 percentage points on compound, layout-sensitive, and cross-slide edits, with large gains in visual fidelity and deck-wide consistency. Despite this, all agents still struggle on long-horizon, document-scale tasks, underscoring how hard reliable PowerPoint editing remains. We publicly release our code: {\hypersetup{urlcolor=myblue}\url{https://github.com/michaelofengend/PPTArena}}.
\end{abstract}
\section{Introduction}\label{sec:intro}

As Vision-Language Model (VLM) agents begin to operate software, the capability that matters most in everyday use—editing existing PowerPoint (PPT) decks with precision—remains largely unverified and under-supported~\cite{openai2025chatgptagent}. Image- or PDF-based formulations discard deck semantics (formats, shape trees, master layouts), while text-to-slides pipelines emphasize generation and ignore edit-in-place constraints~\cite{jung2025talkslideslanguagedrivenagents, ge2025autopresentdesigningstructuredvisuals}. This gap matters because most decks are refined through revision, not from scratch, making reliable layout reasoning and non-destructive modification the realistic bar for agentic PPT capability. Yet we still lack a benchmark that asks the practical question: \emph{can today’s multimodal agents reliably edit existing decks with high instruction fidelity and visual quality?}

Reliable PPT editing is intrinsically hard. Rasterized ``image editing'' views discard the object- and style-level structure that makes editing precise: fonts and paragraphs, shape geometries, z-order, theme colors, slide masters, and cross-references are all lost once a deck is treated as a bitmap. The same instruction (\eg, ``make the subtitle 18pt and align the two logos to the grid'') can require multiple coordinated actions across several slides, conditioned on the existing layout and theme. Evaluation is equally subtle: a change can be syntactically valid yet semantically wrong or aesthetically poor. These failure modes are systematically invisible to benchmarks that only check final text strings, API-level diffs, or pixel similarity, and they motivate a benchmark that treats PPT editing as a structured program over deck semantics, with explicit scoring of both instruction following and visual quality.

Our motivation is grounded in how presentations are actually made and maintained. In professional and academic settings, most decks do not begin from a completely blank canvas; they evolve through continuous revision: merging slides from prior talks, adapting templates for new audiences, and polishing visual hierarchy. Editing reveals whether an agent truly understands the structure already present: the agent must locate the correct element, reason about its relationships (alignment, grouping, z-order), and modify it without collateral changes elsewhere. To capture this reality, we introduce \textbf{PPTArena}, a benchmark designed for PowerPoint editing on real decks. PPTArena assembles 100 real-world source decks and 2{,}125 slides into 1{,}300+ discrete, human-specified edits that range from local text updates to compound, cross-slide transformations, such as deck-wide theme flips, accessibility passes, and multi-step layout repairs. 

Unlike generation datasets that can easily output thousands of static templates, agentic editing requires causal, element-level ground-truths for in-place modifications. Creating these high-fidelity edit pairs is an order of magnitude more complex; consequently, our 1{,}300+ tasks provide substantial statistical power, exceeding the scale of recent multimodal agent benchmarks like OSWorld (369 tasks)~\cite{xie2024osworld} or SWE-bench Multimodal (517 visual issues)~\cite{yang2025swebenchmultimodal}. Furthermore, to ensure rigorous task depth and structural variety, our 100 source decks were meticulously filtered from over 18{,}000 candidates across academic, corporate, creative, and multilingual domains. Each edit bundles an initial deck, a fully specified target deck, and a \emph{style-aware rubric} that disambiguates correctness at the level of content, typography, layout, and color roles. Representative tasks are illustrated in \Cref{fig:teaser}, including filling in missing poster content while preserving hierarchy, flipping theme color roles consistently across a deck, fixing layout and format across related slides, and translating slide content while maintaining charts and structure. To our knowledge, PPTArena is the \textit{first} benchmark that (i) treats PPT editing as a structured, causal program over deck semantics, (ii) ships element-level ground-truths, style targets, and error rubrics to disambiguate correctness, and (iii) uses dual instruction-following and visual judges that are extensively validated against human expert ratings to allow diverse configurations of edits while maintaining task coherence.

Complementing the benchmark, we present \textbf{PPTPilot}, a structure-aware pilot agent for robust, fine-grained PPT editing. PPTPilot decomposes each natural-language instruction into a sequence of semantic operations, chooses between high-level APIs (\eg, \texttt{python-pptx}) and direct XML patching, and validates the outcome against task-specific targets. Two design choices are key: \emph{structure-aware planning}, where the agent parses slide masters, placeholders, shape trees, text, and visual data before editing, and \emph{deterministic execution}, where XML-level patches and strict schemas give exact control over fonts, theme color slots, positions, and master-level changes while programmatic tools handle repetitive global operations (\eg, translation, bulk normalization). An iterative plan-edit-verify loop, coupled with XML validation and visual checks, improves robustness on visually demanding and long-horizon edits.

We evaluate a broad spectrum of baselines on PPTArena, including strong proprietary PPT agents (\eg, ChatGPT Agent~\cite{openai2025chatgptagent} and MiniMax Agent~\cite{minimax_m2_2025}), extended-thinking VLM configurations, and ablations of PPTPilot. Even with generous prompting and tool access, existing systems struggle to balance instruction fidelity with visual quality on compound, multi-step edits, and they frequently fail on cross-slide dependencies and master-level changes surfaced by our benchmark. In contrast, PPTPilot achieves substantially higher scores, improving over strong proprietary agents and frontier VLM systems by more than 10 percentage points on compound, layout-sensitive, and cross-slide edits, while maintaining competitive performance on simpler tasks. Nonetheless, PPTPilot and all evaluated agents still fail on hard, visually dependent tasks, suggesting both the difficulty of PPTArena and the headroom for future research.

Our key contributions are threefold:
(1) \textbf{PPTArena}, a benchmark for agentic PowerPoint editing that (i) operates on deck-native structure rather than rasterized slides, (ii) offers a taxonomy of single- and multi-edit tasks that stress structural grounding, cross-slide consistency, accessibility, and narrative intent, and (iii) pairs each edit with element-level ground-truths, style targets, and a dual-judge protocol (rigorously aligned with human perception) that separately measures instruction fidelity and visual/layout quality, extending beyond prior PPT evaluation setups~\cite{guo-etal-2024-pptc,ge2025autopresentdesigningstructuredvisuals}. (2) \textbf{PPTPilot}, a structure-aware pilot agent that plans edits over semantic elements and executes them via a hybrid of high-level programmatic tools and deterministic XML patching, with routing, strict schemas, and iterative verification designed for controllability, reliability, and transparency.
(3) \textbf{A comprehensive empirical study} of proprietary agents, open VLMs, and PPTPilot variants on PPTArena, revealing that the benchmark is challenging even for state-of-the-art systems. Ultimately, we hope our work bridges the gap between static visual perception and actionable agentic AI, demonstrating that true multimodal understanding requires agents not just to interpret visual layouts, but to causally manipulate their underlying structures.

\section{Related Work}
\label{sec:related}
PowerPoint editing sits at the intersection of autonomous agent platforms, productivity automation, and evaluation tooling. We review related work highly relevant to PPTArena across multimodal agentic benchmarks, presentation editing systems, industrial agent frameworks, and LLM-as-judge evaluation~\cite{sheetagentsurvey2024,ge2023openagi,kurin2025scalingagents,agenticse2025}.  

\noindent\textbf{Multimodal agentic and slide benchmarks.} General-purpose multimodal benchmarks ensure agents possess the perceptual and reasoning depth required for high-quality edits. A line of benchmarks demands grounding and knowledge integration beyond literal reading~\cite{vqartbench2025,reasonvqa2025,schwenk2022aokvqa}. Other datasets stress robustness and preference alignment~\cite{framesvqa2025,rewardbench2025}. More challenging benchmarks aggregate expert-level tasks across multiple disciplines~\cite{yue2023mmmu,mmmupro2024}. PowerPoint-centered benchmarks expose how fragile instruction fidelity remains for today's VLMs. PPTC and PPTC-R evaluate multi-turn editing sessions~\cite{guo-etal-2024-pptc,zhang-etal-2024-pptc}, and SlideAudit offers a structured look at design quality, while ANA probes temporal comprehension ignored by static evaluations~\cite{slideaudit2025,animationattention2025}. Concurrent and independent work~\cite{pptarena_cua2026} also benchmarks agents on PowerPoint, but evaluates GUI-based creation and editing across only 12 source files; in contrast, PPTArena targets in-place editing at an order-of-magnitude larger scale (100 decks, 2{,}125 slides, 1{,}300+ edits) with deterministic structural diffs and per-sample style targets.

Broader agentic benchmarks have progressed from controlled settings toward realistic, multimodal environments---spanning the web, native GUIs and code-driven OS control, and industrial tasks with UI randomization~\cite{zhou2024webarena,NEURIPS2023_5950bf29,koh-etal-2024-visualwebarena,browserarena2025,bonatti2025windowsagentarena,xie2024osworld,kapoor2024omniact,moteki2025fieldworkarena,tian-etal-2025-mmina,crab2024,bmoca2024}---while aggregate leaderboards highlight how evaluator design, scaling laws, and benchmark rigor shape reported capability~\cite{liu2024agentbench,DBLP:conf/iclr/MialonF0LS24,ge2023openagi,chainofagents2025,practices2025agenticbenchmarks,hendrycks2025definition,kurin2025scalingagents,agenticse2025}.

\noindent\textbf{Presentation editing.} Agentic presentation pipelines blend planning, content synthesis, and low-level manipulation. A line of work explores combinations of generation and refinement, yet each reports brittleness in object targeting, template bias, or cascading errors~\cite{ge2025autopresentdesigningstructuredvisuals,pptagent2025,jung2025talkslideslanguagedrivenagents,docrefine2025,autoslides2025}. Other work highlights the cost of over-reliance on generic templates~\cite{jung2025talkslideslanguagedrivenagents}. Comparative studies confirm API-driven execution outperforms GUI-based approaches for fine-grained control~\cite{apivsgui2025,agenticvideoediting2025}, motivating PPTArena's XML-level enforcement and pixel-grounded targets.

\noindent\textbf{Industrial agent and tool-calling.} Progress in agent infrastructure illustrates how self-supervised tools expand computer-use competence~\cite{toolformer2023,gur2023webagent,li2024oscopilot,webvoyager2025} and perception-driven agents~\cite{sheetmind2025,spreadsheetbench2024,cheng2022screenqa}. Open-source orchestration stacks make it easier to compose planners, memory modules, and tool executors, while industrial reports detail production deployments~\cite{autogen2023,langgraph2024,agenticframeworks2025,openai2025chatgptagent,openai2025gpt5systemcard,gemini2025v25report,adobe2025expressassistant}. Concurrent analyses of scaling curves and structured agent engineering~\cite{kurin2025scalingagents,agenticse2025} warn that larger models alone do not guarantee reliability, reinforcing PPTArena's focus on judge audits that make agent improvements interpretable.

\noindent\textbf{LLM- and VLM-as-judge evaluation.} Reliable evaluation remains a bottleneck as agent tasks grow more open-ended. A line of work studies the feasibility of delegating assessment to specialized VLMs~\cite{xu2024reliablejudge,zheng2023judge,mllmasajudge2024}. Follow-up work exposes risks of judge bias, prompting recommendations such as no-free-label baselines, adversarial judge detection, multi-judge audits, and alignment across heterogeneous tasks~\cite{nofreelabels2025,whosjudge2025,surveyllmasjudge2024,rewardbench2025}. Our work adopts these lessons through a dual-judge protocol, separating instruction compliance from visual and layout grading, ensuring that progress measured on PPTArena reflects genuine improvements rather than exploitation of single-model biases.

\section{PPTArena Benchmark}
\label{sec:benchmark}
PPTArena contains 100 decks with real-world editing instructions spanning 2{,}125 slides. Each edit bundles an initial deck, a target deck with human-generated ground-truth references, structured textual instructions, and a rubric capturing layout, typography, color, and content requirements. We group the edits into sixteen topical buckets (detailed in \Cref{tab:edit_taxonomy}), ensuring that the benchmark stresses both semantic reasoning and low-level formatting fidelity.

\subsection{Benchmark Composition and Difficulty}
\label{sec:benchmark_composition}
\textbf{Data sourcing and coverage.} We web-scrape over 18{,}000 PowerPoints (including slides from SlidesCarnival~\cite{slidescarnival}, Zenodo~\cite{zenodo}, SlideShare~\cite{slideshare})—the largest open-sourced dataset of PPTs—and convert them into structured JSON traces that capture layout, styling, and content metadata.

\begin{table}[htpb]
    \caption{Taxonomy of 16 editing operations in PPTArena. The five major categories: Content, Layout, Styling, Interactivity, and Structure, encompass operations from basic text manipulation to advanced master-level edits and accessibility compliance.}
    \centering
    \begin{tabular}{@{}p{0.21\columnwidth}p{0.78\columnwidth}@{}}
    \toprule
    \textbf{Category} & \textbf{Edit Types} \\
    \midrule
    \rowcolor{gray!8}
    \textbf{Content} & 
    \textbf{(1)}~Text \& Typography • \textbf{(2)}~Shapes \& Drawing • \textbf{(3)}~Images \& Pictures • \textbf{(4)}~Tables • \textbf{(5)}~Charts • \textbf{(6)}~SmartArt \& Diagrams • \textbf{(7)}~Audio \& Video \\[2pt]
    
    \textbf{Layout} & 
    \textbf{(8)}~Alignment, Distribution, Grid, Grouping, Z-order • \textbf{(9)}~Slide Layouts \& Placeholders \\[2pt]
    
    \rowcolor{gray!8}
    \textbf{Styling} & 
    \textbf{(10)}~Themes (colors, fonts, effects), Background • \textbf{(11)}~Master-level edits (Slide/Notes Masters) \\[2pt]
    
    \textbf{Interactivity} & 
    \textbf{(12)}~Animations (entrance, emphasis, exit, paths, timing) • \textbf{(13)}~Slide Transitions • \textbf{(14)}~Hyperlinks \\[2pt]
    
    \rowcolor{gray!8}
    \textbf{Structure} & 
    \textbf{(15)}~Slide/Section/Order Mgmt., Slide Numbers, Headers/Footers, Notes • \textbf{(16)}~Comments/Review, Accessibility (alt text, reading order, contrast) \\
    \bottomrule \\
    \end{tabular}
    
    \label{tab:edit_taxonomy}
\end{table}

Automated filtering retains decks with diverse multimodal assets, which we further combine with a curated internal corpus contributed by literature analysts, biology researchers, and art and design students. From more than 500 hand-reviewed candidates, including 25 decks created from scratch by us, we select the 100 PowerPoints that best span professional, academic, multi-lingual, and art/design genres, ensuring every topic bucket in~\Cref{tab:edit_taxonomy} contains challenging exemplars.

\noindent\textbf{Taxonomy-driven task design.} Drawing from established principles of presentation design~\cite{duarte2008slide,reynolds2011presentation,williams2015non,alley2005scientific}, we define five parent categories that decompose into 16 concrete edit types, guaranteeing coverage from low-level typography to master-edit workflows. Following~\cite{pmlr-v235-maia-polo24a}, we favor fewer but richer edits, so the taxonomy is used to balance high-difficulty scenarios rather than to inflate totals. Tasks range from simple text replacements to multi-edit, multimodal reasoning problems. For instance, \Cref{fig:teaser} illustrates a cross-modal editing task that requires matching images to captions across slides using both visual and textual cues; many editing tasks straddle multiple taxonomy buckets, so category counts exceed 100.

PPTArena distinguishes itself from prior benchmarks through its emphasis on \textit{edit difficulty} across four key dimensions: multi-step reasoning depth, cross-slide dependencies, semantic understanding requirements, and long-horizon planning complexity. While existing resources such as PPTC-R~\cite{zhang-etal-2024-pptc} and T2US~\cite{jung2025talkslideslanguagedrivenagents} focus on short, template-bound edits, PPTArena intentionally concentrates difficulty into fewer but richer scenarios to expose real-world failures.

\newcommand{\goodmark}[1]{\textcolor{green!60!black}{\checkmark}\,#1}
\newcommand{\badmark}[1]{\textcolor{red!70!black}{\ensuremath{\times}}\,#1}

\begin{table}[t]
\centering

\begin{minipage}[t]{0.36\textwidth}
\centering
\caption{\textbf{Challenge distribution in PPTArena.} Per-category averages of atomic edit operations and slides touched, with the share of \emph{cross-slide} and \emph{high-diff} cases.}
\label{tab:difficulty_by_category_main}

\scriptsize
\setlength{\tabcolsep}{2.2pt}
\renewcommand{\arraystretch}{1.05}
\resizebox{\linewidth}{!}{%
\begin{tabular}{@{}l r c c c c@{}}
\toprule
\textbf{Category} &
\multicolumn{1}{c}{\rotatebox{90}{\textbf{Cases}}} &
\multicolumn{1}{c}{\rotatebox{90}{\textbf{Avg. ops}}} &
\multicolumn{1}{c}{\rotatebox{90}{\textbf{Avg. slides}}} &
\multicolumn{1}{c}{\rotatebox{90}{\textbf{Cross-slide}}} &
\multicolumn{1}{c}{\rotatebox{90}{\textbf{High-diff.}}} \\
\midrule
Content        & 67 & 11.3 & 6.8  & 26\% & 41\% \\
Layout         & 29 & 15.0 & 8.9  & 34\% & 52\% \\
Styling        & 29 & 13.4 & 9.4  & 30\% & 48\% \\
Structure      & 15 & 17.8 & 11.2 & 56\% & 71\% \\
Interactivity  & 4  & 18.9 & 10.5 & 70\% & 75\% \\
\midrule
All tags & 144 & 13.4 & 8.3 & 32\% & 49\% \\
\bottomrule
\end{tabular}
}
\end{minipage}
\hfill
\begin{minipage}[t]{0.6\textwidth}
\centering
\caption{\textbf{Comparison with prior benchmarks.} PPTArena emphasizes richer, multi-operation, cross-slide, and multimodal tasks.}
\label{tab:edit_difficulty_comparison}

\scriptsize
\setlength{\tabcolsep}{3pt}
\renewcommand{\arraystretch}{1.0}
\resizebox{\linewidth}{!}{%
\begin{tabular}{@{}l*{3}{c}@{}}
\toprule
\textbf{Metric} & \textbf{PPTC-R} & \textbf{T2US} & \textbf{PPTArena} \\
\midrule
\rowcolor{gray!10}
\multicolumn{4}{@{}l}{\textit{Task Complexity}} \\
\quad Avg. operations per edit & 2.9 & 1.2 & \textbf{13.4} \\
\quad Avg. slides per edit & 1.3 & 1.2 & \textbf{8.3} \\
\quad Cross-slide dependencies & 21\% & 5\% & \textbf{32\%} \\
\quad Text-visual dependencies & \badmark{} & 1.3\% & \textbf{28\%} \\
\midrule
\rowcolor{gray!10}
\multicolumn{4}{@{}l}{\textit{Benchmark Design}} \\
\quad Human-created decks & \badmark{} & \goodmark{} & \goodmark{} \\
\quad Python-created decks & \goodmark{} & \badmark{} & \goodmark{} \\
\quad Ground-truth provided & \goodmark{} & \badmark{} & \goodmark{} \\
\quad Cross-platform compatibility & \goodmark{} & \badmark{} & \goodmark{} \\
\midrule
\rowcolor{gray!10}
\multicolumn{4}{@{}l}{\textit{Advanced Requirements}} \\
\quad Accessibility constraints & \badmark{} & \badmark{} & \goodmark{} \\
\quad External knowledge & \badmark{} & \badmark{} & \goodmark{} \\
\quad Complex multimodal tasks & \badmark{} & \badmark{} & \goodmark{} \\
\bottomrule
\end{tabular}
}
\end{minipage}

\end{table}

\noindent\textbf{Difficulty distribution.} Table~\ref{tab:difficulty_by_category_main} quantifies this concentration of difficulty. We flag a case as \emph{cross-slide} when it requires coordinated edits spanning at least two slides, and as \emph{high-diff} when it involves cross-slide dependencies or strong visual-textual reasoning such as chart remapping or translation. Structure and Interactivity edits require the longest programs (17.8--18.9 ops) over the most slides (10--11), driving the highest cross-slide (56--70\%) and high-diff (71--75\%) rates, while even Content, Layout, and Styling keep roughly a third of cases cross-slide and 41--52\% high-diff. PPTArena thus targets multi-slide, multimodal reasoning rather than inflating counts with single-slide tweaks.

\begin{figure}[htbp] 
    \centering
    
    \colorbox{yellow!12}{
        \begin{minipage}{0.95\linewidth}
            \small 
            \textbf{\normalsize Cross-Slide Image-Caption Correlation}\\[-0.2em]
            \rule{\linewidth}{0.5pt}\\[0.1em] 
            \textcolor{blue!70!black}{\texttt{"Category"}}: \textit{Layout} \\
            \textcolor{blue!70!black}{\texttt{"Prompt"}}: \textit{1. For every slide, match each image to its correct caption. 2. After pairing them, rearrange the pairs to create a balanced layout. 3. Update each slide's title correspondingly. 4. Finally, make sure all images are the same size, 3.2 inches wide by 2.4 inches high.}\\[0.1em]
            \rule{\linewidth}{0.5pt}\\[0.3em]
            
            \begin{minipage}[t]{0.48\linewidth}
                \centering
                \fbox{\includegraphics[width=0.85\linewidth]{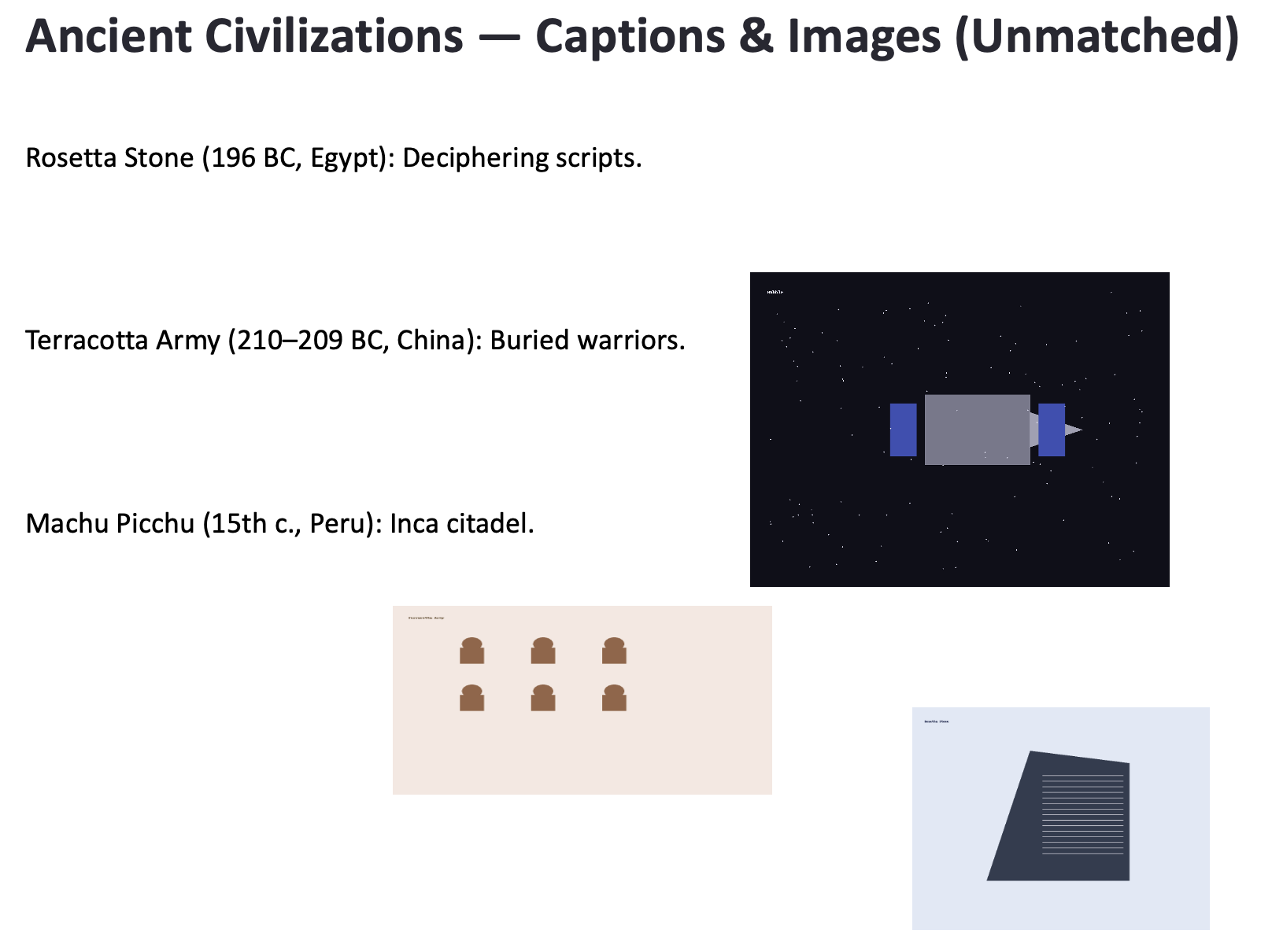}}\\[0.1em]
                {\footnotesize\textbf{(a) Original}}
            \end{minipage}
            \hfill
            \begin{minipage}[t]{0.48\linewidth}
                \centering
                \fbox{\includegraphics[width=0.85\linewidth]{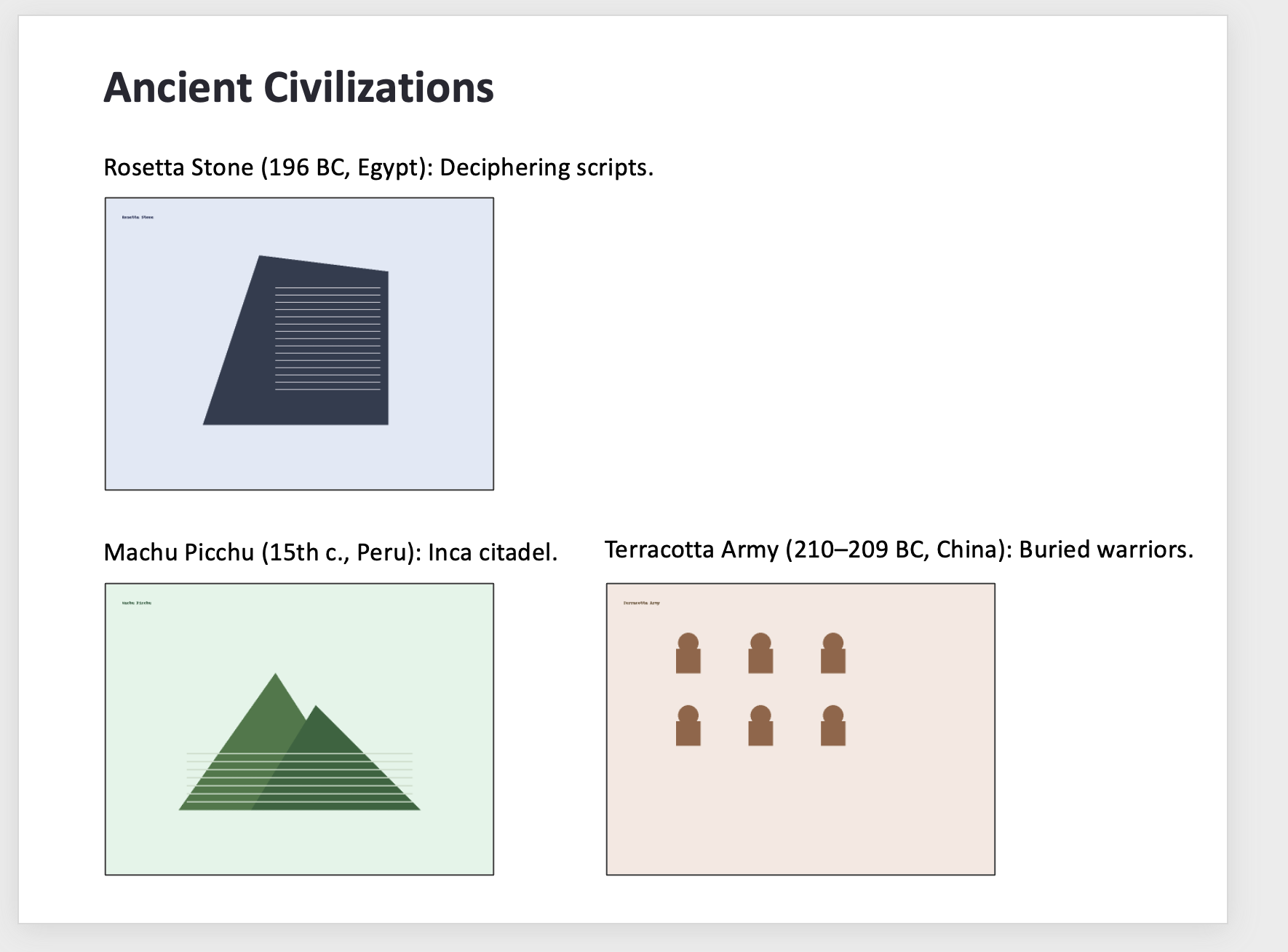}}\\[0.1em]
                {\footnotesize\textbf{(b) Ground-Truth}}
            \end{minipage}
        \end{minipage}
    }
    
     \par\smallskip
    
    \colorbox{yellow!12}{
        \begin{minipage}{0.95\linewidth}
            \small
            \textbf{\normalsize Configure Speaker Notes}\\[-0.2em]
            \rule{\linewidth}{0.5pt}\\[0.1em]
            \textcolor{blue!70!black}{\texttt{"Category"}}: \textit{Content, Structure} \\
            \textcolor{blue!70!black}{\texttt{"Prompt"}}: \textit{Slide 2 contains speaker notes for the other slides. Please move the text from the text boxes on slide 2 to the speaker notes of the appropriate slides, then delete slide 2.}\\[0.1em]
            \rule{\linewidth}{0.5pt}\\[0.3em]
            
            \begin{minipage}[t]{0.48\linewidth}
                \centering
                \fbox{\includegraphics[width=0.65\linewidth]{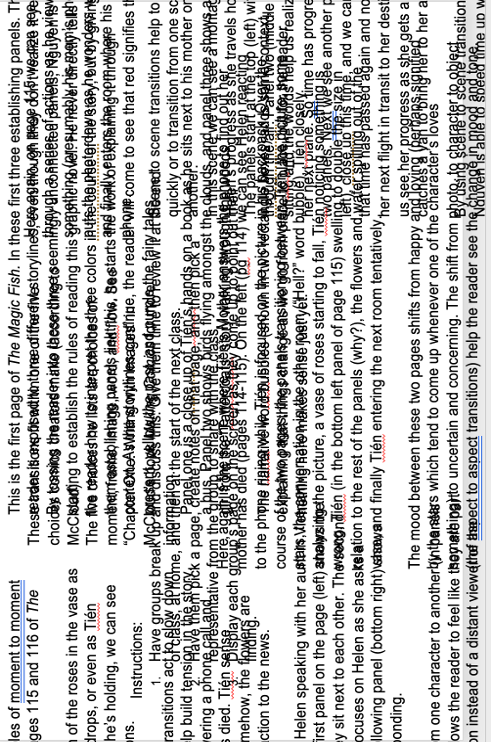}}\\[0.1em]
                {\footnotesize\textbf{(c) Original slide 2}}
            \end{minipage}
            \hfill
            \begin{minipage}[t]{0.48\linewidth}
                \centering
                \fbox{\includegraphics[width=0.65\linewidth]{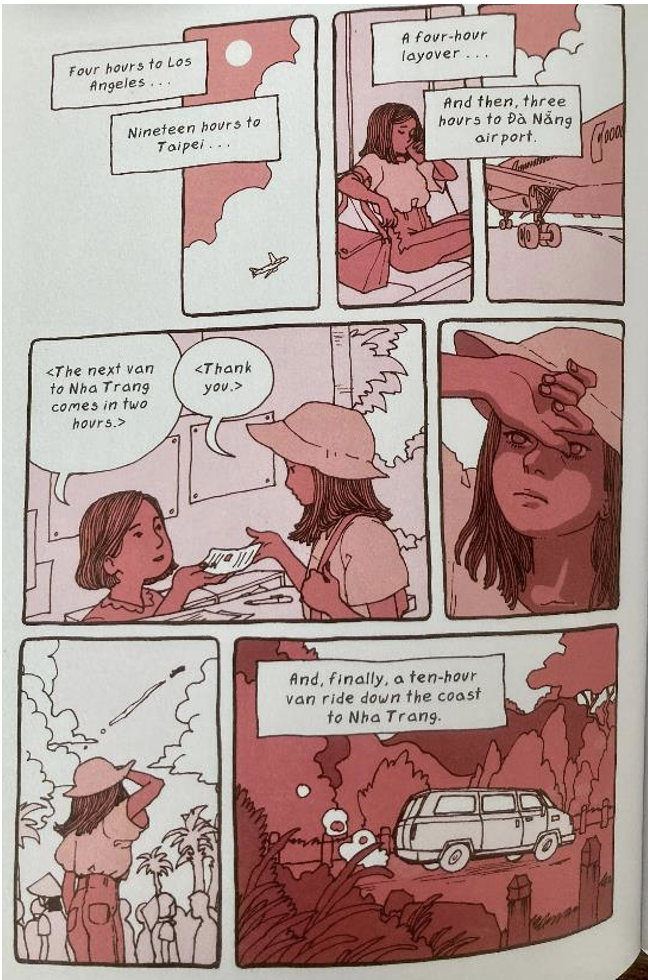}}\\[0.1em]
                {\footnotesize\textbf{(d) Ground-Truth slide 6}}
            \end{minipage}
        \end{minipage}
    }
    
    \caption{Representative edits from PPTArena. \textbf{Top:} Cross-slide image-caption matching requires semantic understanding to correlate visual and textual content. \textbf{Bottom:} Configuring speaker notes requires aligning narration, speaker notes, and visual panels across slides, forcing the agent to reason about cross-slide correspondences before deleting the staging slide.}
    \label{fig:representative_cases}
\end{figure}

\subsection{Comparison with Prior Benchmarks}

Table~\ref{tab:edit_difficulty_comparison} quantitatively compares PPTArena against prior benchmarks. PPTArena demonstrates substantially higher complexity with an average of 13.4 operations per edit and 8.3 slides per edit, with 32\% of edits involving cross-slide dependencies and 28\% requiring visual-textual reasoning. Prior benchmarks suffer from key limitations: PPTC-R relies on synthetically generated decks that lack real-world visual richness and complexity, while T2US artificially inflates its size by rewording prompts for identical tasks rather than introducing genuine task diversity~\cite{jung2025talkslideslanguagedrivenagents, zhang-etal-2024-pptc, lee-etal-2024-prometheus}.

In contrast, PPTArena combines both human-created and Python-generated decks, provides ground-truth for deterministic evaluation, and maintains cross-platform compatibility. It uniquely incorporates accessibility constraints, external knowledge requirements, and complex multimodal tasks that emphasize multi-step reasoning, cross-slide dependencies, and long-horizon planning. This design exposes failure modes that remain hidden on simpler tasks and provides a rigorous testbed for evaluating the next generation of agentic systems. We demonstrate two representative samples in \Cref{fig:representative_cases}.

\subsection{VLM-as-Judge Evaluation Protocol}\label{sec:eval}

\noindent\textbf{Instruction following (\textbf{IF}) and visual quality (\textbf{VQ}).} Our evaluation framework is designed to move beyond simple pixel- or code-level diffs, which fail to capture the semantic and aesthetic goals of PPT editing. We instead measure an agent's performance on two fundamental axes: \emph{Instruction Following} and \emph{Visual Quality} ~\cite{sim-etal-2025-vlms}, both scored by expert VLM judges on an integer scale from 0 (Failure) to 5 (Perfect).

\noindent(1) \emph{Instruction Following (IF)} measures the agent's semantic and logical adherence to the user's prompt. It assesses \textit{what} was done, such as correctly identifying and moving content, applying the right formatting, or fulfilling all sub-tasks in a complex command.

\noindent(2) \emph{Visual Quality (VQ)} measures the \textit{aesthetic and professional polish} of the resulting slide. It assesses \textit{how} the changes were implemented, focusing on layout, alignment, typography, color harmony, and overall visual appeal, independent of the instruction's logical fulfillment.

By combining the two metrics together, our PPTArena covers the common user requirements from either content aspects (IF) or aesthetic aspects (VQ).

\noindent\textbf{Per-sample rubric: style target.}
A core challenge in PPT editing is the immense variation across decks, layouts, and design habits. There is no universal rubric that can reliably score every instruction, which separates our setting from common LLM judges used in question-answering. Our solution is to generate a fine-grained, per-sample style target that specifies all crucial structural and visual requirements for that editing task. For example, if the user prompt is ``Overhaul this rock-cycle presentation with ..., reorganizing the rock types into three columns on slide 2, and replace slide 3's wall of text with ...,'' then the corresponding style-target rubric would spell out the exact ground-truth with hyper-specific constraints, such as: \textit{``... must have a geology photo occupying ... Slide 2 columns must be labeled ... Slide 3 centers a fully labeled cycle diagram linking the three rock types with ...''}.

To provide trustworthy style targets, we combine automatic generation together with exhaustive human verification. Specifically, we send the JSON summaries and screenshots of the PPT's ground-truth and original decks to GPT-5~\cite{openai2025gpt5systemcard} to generate style targets. Both ground-truth and the original slide are provided as input so that the generation of the style target can precisely understand the desired outcome. Then each style target is manually verified for correctness and faithfulness to the editing instructions and PPT context.

\begin{figure}[t]
    \centering
    \includegraphics[width=1.0\columnwidth]{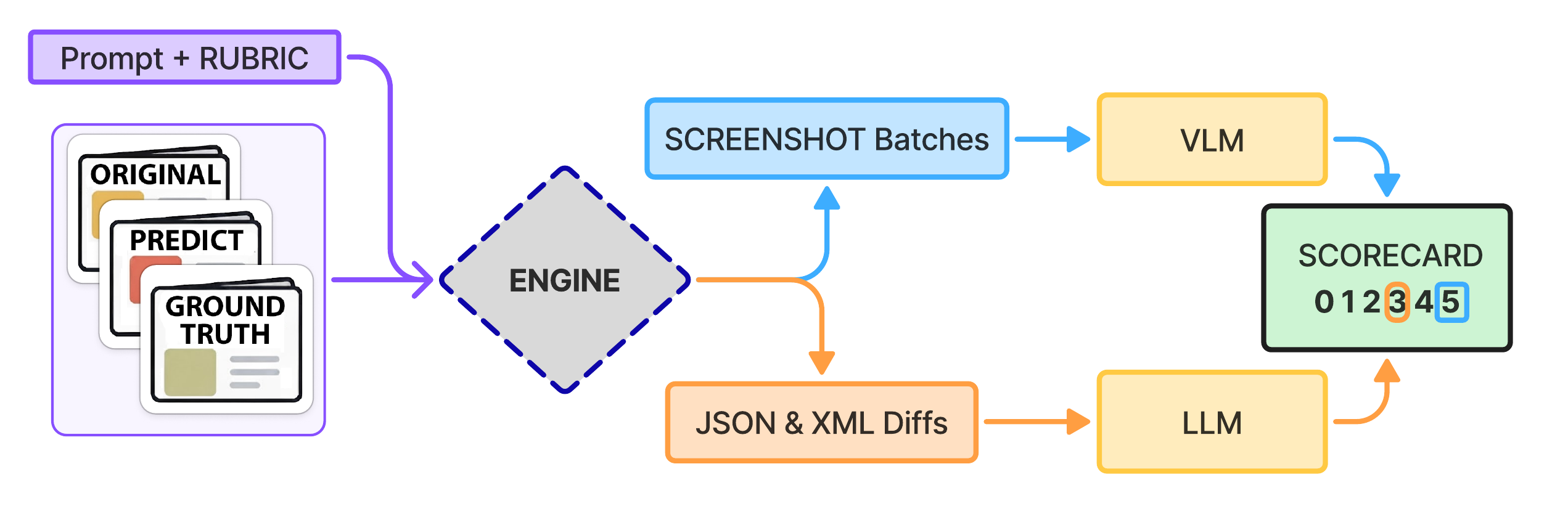}
    \caption{\textbf{Our VLM-as-judge paradigm.} To maximize the reliability of existing VLMs, we employ two separate judges. The visual quality (VQ) judge primarily comprehends the PPT screenshots for visual understanding, while the instruction-following (IF) judge focuses on structured data to analyze the contents.}
    \label{fig:vlm_judge_pipeline}
\end{figure}

\noindent\textbf{Dual-judge framework for reliable evaluation.}
To ensure reliable scoring, we employ a dual-judge framework that separately conducts the evaluation for instruction following and visual quality (as in \Cref{fig:vlm_judge_pipeline}), each implemented as a separate VLM, \eg, GPT-5.2~\cite{openai2025gpt5systemcard}. To further enhance the judge's capability for IF and VQ, respectively, we selectively provide them contexts that align best with the evaluation target, in addition to the style target mentioned above.

\noindent(1) \emph{Instruction Following Judge}: This judge receives \textit{only} the structured data diffs (\eg, JSON and XML summaries) between the original, predicted, and ground-truth slides. Such inputs force the judge to concentrate at the content level and carefully inspect whether the desired content changes are correct.

\noindent(2) \emph{Visual Quality Judge}: This judge is responsible for visual aesthetics, so it receives \textit{only} the rendered screenshots of the predicted and ground-truth slides. Its context is engineered to focus purely on aesthetics, comparing the visual execution of alignment, layout, and style against the rubric.

To further improve the reliability, especially for multi-slide edits, we use Structural Similarity Index Measure (SSIM) screening to forward only the slides with salient changes to the visual judge, so that it concentrates better on the edits without being overwhelmed by the enormous contexts.

\noindent\textbf{Comparison with judges in related benchmarks.}
Our evaluation methodology marks a significant advance over existing benchmarks. Prior work, such as PPTC-R~\cite{guo-etal-2024-pptc}, primarily relies on API-level ``diffs''. While useful, this approach is brittle and cannot detect critical semantic errors, such as an agent copying the correct text but to the \textit{wrong slide}. Our IF Judge, by operating on structured summaries, is explicitly designed to capture such logical failures. Furthermore, while other benchmarks like T2US~\cite{jung2025talkslideslanguagedrivenagents} also use VLM-as-judge, their reliance on prompting without a strong rubric leads to noisy and unreliable ratings. PPTArena designs the style target to mitigate such issues by providing a rigorous and reproducible foundation for scoring. 

To conclude, our dual-judge, rubric-grounded architecture is essential for measuring the complex, multi-step reasoning required for PPT editing and for analyzing the subtle failure modes that simpler benchmarks would miss.

\section{An Effective PPT Editing Agent: PPTPilot}
\label{sec:pptpilot}

We introduce PPTPilot, an agent for presentation editing, whose simple architecture yields surprisingly effective results, even outperforming proprietary products like OpenAI's ChatGPT Agent~\cite{openai2025chatgptagent} (details in Sec.~\ref{sec:main_comparison}). Our design is built on two key insights. \textbf{(1)} The primary challenge in this domain is not solely intent recognition, but reliability and precision. PowerPoint files are built on the brittle Office Open XML (OOXML) format, which is highly intolerant to malformed or ``hallucinated'' VLM outputs, and therefore requires specialized formats and contexts suitable for PPT editing. \textbf{(2)} We found that no single editing modality, \eg, purely relying on the XML format, is sufficient. A robust agent must be capable of intelligently selecting the optimal tool and editing interface for a given task.

\begin{figure}[!tbp]
    \centering
    \includegraphics[width=1.00\linewidth]{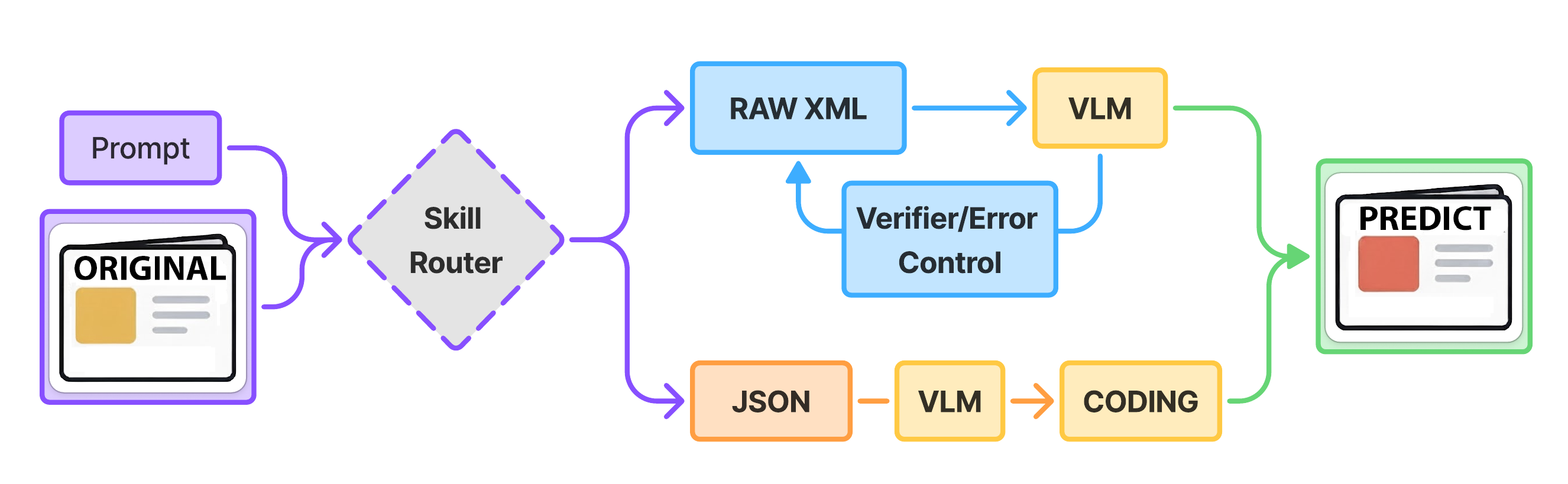}
    \caption{\textbf{Our PPTPilot paradigm.} Our key insight highlights a combination of two different editing skills: functional code and direct XML edits to control the fine-grained structural elements. A ``Skill Router'' determines which skill is more suitable for a query. And then the corresponding VLM executes the edits via either route. Notably, the editing operations can also be enhanced with reflection, trading time for more reliable editing. For the router we use a fast LLM (GPT-5 nano or Gemini-3.0 flash), and then GPT-5.2 for our VLM edit calls.}
    \label{fig:pptpilot_diagram}
\end{figure}

PPTPilot's core design is a \textit{dual-path} architecture, emphasizing the ability to handle editing queries via either programmatic tools or direct XML editing (Figure~\ref{fig:pptpilot_diagram}). This hybrid design enables our agent to address a wide range of editing queries reliably and in a principled way.

\noindent\textbf{Programmatic editing.}
Utilizing \texttt{python-pptx} to edit PPTs programmatically---a method commonly adopted in prior work~\cite{jung2025talkslideslanguagedrivenagents,pptagent2025,ge2025autopresentdesigningstructuredvisuals,autoslides2025,pang2025paper2postermultimodalposterautomation}---scripts the edits by generating code (bottom of Figure~\ref{fig:pptpilot_diagram}). This approach is highly effective for repetitive, well-defined, and content-centric operations, such as performing a ``find-and-replace'' across all slides or translating text. However, it lacks the fine-grained control required for complex structural modifications (\eg, altering slide masters, themes, or specific layout geometries).

\noindent\textbf{Direct XML editing.}
To address the limitations of the programmatic path in structural and visual editing scenarios, we have equipped PPTPilot with a second skill: the ability to directly read, parse, and rewrite raw OOXML files (\eg, \texttt{slide1.xml}, \texttt{theme.xml}), as shown in the top half of Figure~\ref{fig:pptpilot_diagram}. This approach provides the precision required for structured contexts, as the VLMs can directly manipulate fine-grained properties like the specific positions of elements. Since OOXML encodes most of the information in a PPT, the XML path provides a unified interface well-aligned with existing VLMs for PPT editing. However, the long context and strict format requirements of XML make it challenging to perform precise edits, especially when modifications span a large number of slides, in which case the programmatic approach is significantly more reliable.

\noindent\textbf{Skill routing.} To determine which editing skills to adopt for a specific user query, we employ an LLM that routes the query to the proper editing skills, as the beginning of branching in Figure~\ref{fig:pptpilot_diagram}. Upon receiving a user instruction, this decider analyzes the prompt combined with the presentation's structure, including the screenshots and contents. Based on this analysis, it routes the task to either the programmatic path or the direct XML editing path. Across PPTArena, the router dispatches 66\% of edits to the programmatic path and 34\% to the direct XML path (Supp.\ Sec.~G.1).

\noindent\textbf{Self-correction with reflection.} Finally, we acknowledge the complexity of PPT editing, which indicates the challenge of correct edits in a single try. Inspired by representative agents such as ReACT~\cite{yao2022react}, we introduce an iterative reflection path into the PPTPilot, so that it can gradually refine its predictions. The agent proposes an edit to the XML files, which is rendered temporarily to a PPT file. Then a verifier model assesses the output PPT according to the original instructions and provides feedback for failures. In this way, the agent produces an updated PPT edit based on the feedback and is able to correct for its own errors.

Despite the simplicity of our design, we find it effective and efficient for PPT editing when compared against existing agent products and frameworks. We hope our PPTPilot can serve as a baseline for research into PPT editing.

\section{Experiments}
\label{sec:experiments}

\noindent\textbf{VLM-as-Judge.} We evaluate with the strongest vision-language models, Gemini 3.1 Pro~\cite{gemini2025v25report} and GPT-5.2~\cite{openai2025gpt5systemcard}. The main comparison reports the GPT-5.2 judge (Tables~\ref{tab:pptarena_subset25_gpt5} and~\ref{tab:pptarena_full_gpt5}); a cross-judge robustness check under Gemini~3.1~Pro is summarized in Sec.~\ref{sec:judge_human_alignment}, with the full per-system breakdown and implementation details in the supplementary (Supp.\ Table~A).

\noindent\textbf{Agent Baselines.} We evaluate a wide range of PPT agents against PPTPilot: extended-thinking ChatGPT~\cite{openai2025gpt5systemcard}, Gemini-CLI~\cite{gemini2025v25report}, ChatGPT-Agent~\cite{openai2025chatgptagent} (which drives a simulated desktop and terminal), and MiniMax Agent, marketed as a multimodal ``PPT helper.'' We additionally include the open-weight Kimi-K2.6~\cite{kimi2025k2} as a non-proprietary reference point. PPTAgent~\cite{pptagent2025} and Paper2Poster~\cite{pang2025paper2postermultimodalposterautomation} both fail on all tasks.

\noindent\textbf{Subset Evaluation.} Because several proprietary products enforce strict rate limits and high costs (\eg, ChatGPT and MiniMax Agent~\cite{minimax_m2_2025}), we also evaluate on a budget-limited subset of 25 decks: the 20 hardest cases plus 5 added for breadth. Subset details, per-system budgets, and the ${\approx}\,\$700$ total cost are in the supplementary (Supp.\ Secs.~D and~I).

\subsection{Performance on PPTArena}\label{sec:main_comparison}

\begin{table*}[!t]
\caption{{\bf PPTArena matched 25-case subset evaluation.} All systems here are evaluated on the same 25-deck, 206-edit subset.}
  \centering
  \footnotesize
  \renewcommand{\arraystretch}{0.95}
  \setlength{\tabcolsep}{4.2pt}
  \resizebox{\linewidth}{!}{%
  \begin{tabular}{l r
      >{\columncolor{cyan!9}}c >{\columncolor{cyan!9}}c
      >{\columncolor{gray!12}}c >{\columncolor{gray!12}}c
      >{\columncolor{orange!12}}c >{\columncolor{orange!12}}c
      >{\columncolor{green!8}}c >{\columncolor{green!8}}c
      >{\columncolor{purple!10}}c >{\columncolor{purple!10}}c
      >{\columncolor{blue!8}}c >{\columncolor{blue!8}}c
      c c}
  \toprule
  & & \multicolumn{2}{c}{\textbf{PPTPilot}}
    & \multicolumn{2}{c}{\textbf{Gemini CLI}}
    & \multicolumn{2}{c}{\textbf{ChatGPT}}
    & \multicolumn{2}{c}{\textbf{ChatGPT Agent}}
    & \multicolumn{2}{c}{\textbf{MiniMax Agent}}
    & \multicolumn{2}{c}{\textbf{Kimi-K2.6}}
    & \multicolumn{2}{c}{\textbf{PPTAgent}} \\
  \cmidrule(lr){3-4}\cmidrule(lr){5-6}\cmidrule(lr){7-8}
  \cmidrule(lr){9-10}\cmidrule(lr){11-12}\cmidrule(lr){13-14}\cmidrule(lr){15-16}
  \textbf{Category} & \textbf{Decks}
  & IF$\uparrow$ & VQ$\uparrow$ & IF$\uparrow$ & VQ$\uparrow$
  & IF$\uparrow$ & VQ$\uparrow$ & IF$\uparrow$ & VQ$\uparrow$
  & IF$\uparrow$ & VQ$\uparrow$ & IF$\uparrow$ & VQ$\uparrow$
  & IF$\uparrow$ & VQ$\uparrow$ \\
  \midrule
  Content       & 19 & \textbf{2.00} & \textbf{1.95}          & 1.84 & 1.53          & 1.05 & 1.35          & 1.80 & 1.50          & 1.10 & 0.75 & 1.90 & 1.15 & 0.00 & 0.00 \\
  Layout        & 7  & 1.86 & \textbf{2.29}          & 1.57 & 1.57          & \textbf{2.00} & \textbf{2.29} & 1.14 & 0.71          & 0.71 & 0.71 & 1.86 & 0.57 & 0.00 & 0.00 \\
  Styling       & 5  & \textbf{2.40} & 1.60          & 2.00 & 2.40          & 1.83 & \textbf{2.67} & 0.83 & 1.67          & 1.00 & 1.00 & 1.33 & 1.17 & 0.00 & 0.00 \\
  Structure     & 3  & 1.47 & 2.00                   & 0.67 & 2.00          & 0.33 & 1.33          & \textbf{2.00} & \textbf{2.33} & 0.67 & 1.33 & 1.00 & 0.33 & 0.00 & 0.00 \\
  Interactivity & 1  & \textbf{3.00} & \textbf{2.00} & 0.00 & 0.00          & 1.00 & 1.00          & 0.00 & 0.00          & 2.00 & 1.00 & 0.00 & 0.00 & 0.00 & 0.00 \\
  \midrule
  Total         & 25 & \textbf{1.87} & \textbf{1.91} & 1.78 & 1.71          & 1.12 & 1.56          & 1.68 & 1.60          & 1.04 & 0.84 & 1.68 & 1.00 & 0.00 & 0.00 \\
  \bottomrule
  \end{tabular}}
  
  \label{tab:pptarena_subset25_gpt5}
\end{table*}

\begin{table}[!ht]
    \caption{{\bf PPTArena full-benchmark evaluation}}
  \centering
  {\footnotesize
  \renewcommand{\arraystretch}{0.9}
  \setlength{\tabcolsep}{7pt}
  \begin{tabular}{l r r
      >{\columncolor{cyan!9}}c >{\columncolor{cyan!9}}c
      >{\columncolor{gray!12}}c >{\columncolor{gray!12}}c
      >{\columncolor{orange!12}}c >{\columncolor{orange!12}}c}
  \toprule
  & & & \multicolumn{2}{c}{\textbf{PPTPilot}}
    & \multicolumn{2}{c}{\textbf{Gemini CLI}}
    & \multicolumn{2}{c}{\textbf{ChatGPT}} \\
  \cmidrule(lr){4-5}\cmidrule(lr){6-7}\cmidrule(lr){8-9}
  \textbf{Category} & \textbf{Decks} & \textbf{Edits}
  & IF$\uparrow$ & VQ$\uparrow$
  & IF$\uparrow$ & VQ$\uparrow$
  & IF$\uparrow$ & VQ$\uparrow$ \\
  \midrule
  Content       & 67  & 887  & \textbf{2.54} & \textbf{2.75} & 1.30 & 2.54 & 2.03 & 2.20 \\
  Layout        & 29  & 504  & \textbf{2.64} & \textbf{2.38} & 1.07 & 1.78 & 2.08 & 2.19 \\
  Styling       & 29  & 413  & 2.32 & \textbf{2.72} & 0.91 & 1.89 & \textbf{2.41} & 2.44 \\
  Structure     & 15  & 182  & \textbf{2.43} & \textbf{2.95} & 1.32 & 2.27 & 1.73 & 1.93 \\
  Interactivity & 4   & 30   & 3.00 & \textbf{3.00} & 0.00 & 0.00 & \textbf{3.25} & 2.75 \\
  \midrule
  Total         & 100 & 1340 & \textbf{2.57} & \textbf{2.69} & 1.21 & 1.98 & 2.07 & 2.22 \\
  \bottomrule
  \end{tabular}
  \label{tab:pptarena_full_gpt5}}
\end{table}

Table \ref{tab:pptarena_full_gpt5} summarizes system-level performance across the full benchmark. PPTPilot attains the strongest overall results with an instruction-following score of 2.57 and a visual-quality score of 2.69.

\noindent\textbf{Baseline system performance.} ChatGPT performs well on straightforward content edits and light styling, yet drops on tasks requiring visual-text alignment, cross-slide reasoning, or deck-wide structural constraints. ChatGPT Agent shows somewhat stronger visual performance but struggles with multi-step logical instructions. Its relative edge on \emph{Structure} edits stems from GUI-level control: PowerPoint natively handles slide masters, footers, and section dividers, whereas PPTPilot can over-edit master XML when a change is slide-local. MiniMax Agent, despite being marketed as a presentation tool, underperforms consistently: it is notable only in the visual-layout category, trailing PPTPilot and ChatGPT Agent elsewhere. PPTAgent highlights the fragility of one-shot, generation-driven pipelines: its outputs diverge substantially from the original structure, breaking preservation requirements and failing all tasks under our rubric.

\noindent\textbf{Runtime efficiency, latency, and key design principles.}
We report end-to-end wall-clock latency per task (Table~\ref{tab:latency_main}); runs that fail to produce a valid
\setlength{\intextsep}{2pt}
\begin{wraptable}{r}{0.5\textwidth}
    \centering
    \footnotesize
    \caption{End-to-end latency per task under each method's original environment.}
    \renewcommand{\arraystretch}{0.95}
    \setlength{\tabcolsep}{6pt}
    \begin{tabular}{lc}
        \toprule
        \textbf{Method} & \textbf{Latency (min)} $\downarrow$ \\
        \midrule
        ChatGPT Agent & 4--30 \\
        MiniMax Agent & 3 \\
        PPTPilot (1-pass) & 1.5 \\
        PPTPilot (3-loop) & 3 \\
        \bottomrule
    \end{tabular}
    \label{tab:latency_main}
\end{wraptable}
PPTX within our cap are timeouts. PPTPilot completes edits quickly in a single pass, while ChatGPT Agent often enters long verification loops and can hit the cap; a 3-iteration reflection loop stays competitive in latency while improving reliability. Together, these results indicate that reliable PPT editing hinges on three properties embodied in PPTPilot: (i) structure-aware planning over deck semantics, (ii) a hybrid execution model routing between programmatic APIs and deterministic OOXML operations, and (iii) an iterative refine-and-verify loop that stabilizes long-horizon edits. We break down these techniques in Table~\ref{tab:pptpilot_ablation_main} (full study in Supp.\ Sec.~E.3).

\begin{table}[t]
  \centering
  \footnotesize
  \setlength{\tabcolsep}{10pt}
  \renewcommand{\arraystretch}{1.05}
  \caption{\textbf{PPTPilot ablation on PPTArena.} Average instruction fidelity (IF) and visual quality (VQ) across judge configurations (top block) and executor variants (bottom block). Hybrid routing plus the reflection loop is essential; the dual judge with structured diffs is the only stable evaluation regime.}
  \label{tab:pptpilot_ablation_main}
  \begin{tabular}{lcc}
  \toprule
  \textbf{Configuration} & \textbf{IF}$\uparrow$ & \textbf{VQ}$\uparrow$ \\
  \midrule
  \rowcolor{LightGrey}
  \multicolumn{3}{l}{\textit{Judge configuration}} \\
  Single VLM judge (all signals) & 2.31 & 4.26 \\
  Dual judge (no diffs) & 3.76 & 4.54 \\
  Dual judge with diffs & 2.36 & 2.40 \\
  \midrule
  \rowcolor{LightGrey}
  \multicolumn{3}{l}{\textit{PPTPilot executor variants}} \\
  XML-only & 0.95 & 2.85 \\
  \texttt{python-pptx}-only & 2.06 & 2.73 \\
  Hybrid (no refinement) & 2.36 & 2.69 \\
  Hybrid + Loop (3$\times$) & \textbf{2.84} & \textbf{3.21} \\
  \bottomrule
  \end{tabular}
\end{table}

\noindent\textbf{Ablating PPTPilot's design.} Table~\ref{tab:pptpilot_ablation_main} isolates the three properties above. Forcing a single execution path collapses performance: XML-only reaches only 0.95 IF (it struggles with deck-wide operations), and \texttt{python-pptx}-only reaches 2.06 IF (it lacks fine-grained structural control). Hybrid routing recovers most of the gap (2.36 IF), and the iterative refine-and-verify loop lifts the full system to 2.84 IF / 3.21 VQ, with most corrections landing by the second pass. The top block validates the dual-judge protocol: a single judge over all signals is over-lenient (VQ 4.26, a 33\% IF/VQ gap), and dropping structured diffs inflates scores further (3.76/4.54), whereas the dual judge with diffs yields the calibrated, internally consistent scores we report throughout.

\subsection{VLM-as-Judge Reliability and Human Alignment}
\label{sec:judge_human_alignment}

Code-level metrics fail to capture nuanced visual semantics such as occlusion, alignment, and layout balance, so we evaluate with two specialized VLM judges---one for \emph{Instruction Following} (IF) and the other for \emph{Visual Quality} (VQ). To validate that these track human perception, we run a blind study on a stratified 10\% subset across all editing categories, with 25 human experts rating IF and VQ on the judges' ordinal scale. Table~\ref{tab:human_corr_main} shows strong human-judge rank and linear correlations, confirming the judges' reliability.

\setlength{\intextsep}{2pt}
\begin{wraptable}{r}{0.50\textwidth}
    \centering
    \footnotesize
    \caption{Human alignment and judge consistency (IF/VQ) on a stratified 10\% subset with 25 expert raters.}
    \renewcommand{\arraystretch}{0.95}
    \setlength{\tabcolsep}{5pt}
    \begin{tabular}{lcc}
        \toprule
        \textbf{Metric} & \textbf{IF} & \textbf{VQ} \\
        \midrule
        Pearson ($r$) $\uparrow$      & 0.72 & 0.81 \\
        Spearman ($\rho$) $\uparrow$  & 0.65 & 0.80 \\
        Kendall ($\tau$) $\uparrow$   & 0.52 & 0.71 \\
        \multicolumn{3}{l}{\textit{Consistency (5 runs per example)}} \\
        Majority agreement $\uparrow$ & 78\%  & 95\% \\
        Std.\ dev.\ $\downarrow$      & 15.4\% & 9\% \\
        \bottomrule
    \end{tabular}
    \label{tab:human_corr_main}
\end{wraptable}
We also assess pipeline stability by running each judge five times per example on identical inputs, reporting the fraction of examples with majority agreement and the standard deviation of the discrete scores (as percentages). Table~\ref{tab:human_corr_main} shows that VQ judging is exceptionally consistent, and both metrics give stable, reproducible signals. A page-wise control study confirms batching does not distort VQ: an independent Gemini~3.1~Pro judge scoring 105 matched pairs one page at a time agrees within one point on 8 of 10 cases (mean VQ $1.88$ page-wise vs.\ $2.00$ batched; Pearson $r{=}0.917$).

\noindent\textbf{Cross-judge robustness.} To confirm the leaderboard is not an artifact of a single judge, we re-score every system with an independent Gemini~3.1~Pro judge (full per-category table in Supp.\ Table~A). Absolute values shift slightly, but the ranking is stable: PPTPilot retains its lead in both IF (2.45) and VQ (2.74) on the full benchmark, ahead of ChatGPT (1.97/2.03) and Gemini CLI (1.92/2.15), and keeps its lead on the hard subset.

\subsection{Human Evaluation of PPTPilot vs.\ Proprietary Agents}
\label{sec:human_eval_pptpilot}
\setlength{\intextsep}{0pt}
\begin{wraptable}[7]{r}{0.45\textwidth}
    \centering
    \footnotesize
    \caption{Blind human scores (25 expert raters) on the stratified subset.}
    \renewcommand{\arraystretch}{0.95}
    \setlength{\tabcolsep}{6pt}
    \begin{tabular}{lcc}
        \toprule
        \textbf{Method} & \textbf{IF} $\uparrow$ & \textbf{VQ} $\uparrow$ \\
        \midrule
        MiniMax Agent & 1.75 & 1.85 \\
        ChatGPT Agent & 3.17 & 2.82 \\
        PPTPilot      & \textbf{3.86} & \textbf{3.81} \\
        \bottomrule
    \end{tabular}
    \label{tab:human_eval_main}
\end{wraptable}
Evaluators consistently rate PPTPilot above the proprietary ChatGPT Agent and
MiniMax Agent baselines; the newly rated MiniMax Agent (IF/VQ $=1.75/1.85$) remains far below PPTPilot ($3.86/3.81$, Table~\ref{tab:human_eval_main}). These human-grounded results corroborate our automated findings (Sec.~\ref{sec:main_comparison}).

\noindent\textbf{What the human study adds.} Under direct preference the gaps widen: PPTPilot leads ChatGPT Agent by $0.69$ IF / $0.99$ VQ and MiniMax by over two points on both axes (Table~\ref{tab:human_eval_main}), with raters flagging the deck-wide-constraint and text--image alignment failures our rubric targets (Supp.\ Sec.~H).

\section{Conclusion}
\label{sec:conclusion}
We introduce PPTArena and PPTPilot for evaluating and improving PowerPoint editing in agentic multimodal systems. PPTArena grounds evaluation in real decks with structured rubrics and dual judges for instruction fidelity and visual quality. PPTPilot’s dual-path plan-edit-verify design improves reliability over strong proprietary agents and frontier VLMs. Yet agents still fail on complex, long-horizon, multi-modal tasks indicating that robust PPT editing remains far from solved. We hope PPTArena and PPTPilot seed future work on reliable editing of the documents people actually use.

\noindent\textbf{Limitations and future work.} Open directions include conversational refinement for under-specified intent, cross-application workflows (live charts, document-to-deck synthesis), and hyper-specialized domains beyond our current taxonomy; agents also remain brittle on edits coupling visual, spatial, and cross-slide reasoning (Supp.\ Secs.~H and~K).

\subsubsection*{Acknowledgements.}
This work was supported in part by NSF under Grants 2106825 and 2519216; the DARPA Young Faculty Award; the ONR Grant N00014-26-1-2099; the NIFA Award 2020-67021-32799; the NSF REU Site Program under Award 2447843 (FoDOMMaT: The Future of Discovery: Training Students to Build and Apply Open-Source Machine Learning Models and Tools); the Apple AIML Academic Research Program; the Amazon-Illinois Center on AI for Interactive Conversational Experiences; the Capital One Illinois Center for Generative AI Safety, Knowledge Systems, and Cybersecurity; and the Toyota Research Institute. This work used computational resources, including the NCSA Delta and DeltaAI supercomputers through allocations CIS230012, CIS230013, CIS240133, and CIS240387 from the Advanced Cyberinfrastructure Coordination Ecosystem: Services \& Support (ACCESS) program; the TACC Frontera supercomputer, Amazon Web Services (AWS), and OpenAI API through the National Artificial Intelligence Research Resource (NAIRR) Pilot; and Google Cloud research credits through the Gemini Academic Program.

\bibliographystyle{splncs04}
\bibliography{main}

\clearpage
\renewcommand\thesection{\Alph{section}}
\renewcommand\thetable{\Alph{table}}
\renewcommand\thefigure{\Alph{figure}}
\renewcommand\theequation{\Alph{equation}}
\setcounter{section}{0}
\setcounter{table}{0}
\setcounter{figure}{0}
\setcounter{equation}{0}


\begingroup
\makeatletter
\setlength{\@fptop}{0pt}
\makeatother
\setlength{\fboxrule}{0.4pt}
\let\oldincludegraphics\includegraphics
\renewcommand{\includegraphics}[2][]{%
  \setlength{\fboxsep}{0pt}%
  \fbox{\oldincludegraphics[#1]{#2}}}

\section{Video Demo}
\label{app:demo-screenshots}

We have created a web-app based on PPTPilot, which can convert a given PPT according to the user instructions. Please refer to the \textbf{videos in our supplementary materials} for their process. In our Fig.~\ref{fig:pptpilot_demo_screenshots}, we have illustrated the webapp conducting chat-based editing for users' provided PPTs and how PPTArena runs the evaluation by comparing the ground truth and prediction slides. We also provide a video of how PPTPilot reasons and conducts edits via xml manipulation.
  
  \begin{figure}[!t]
      \centering
      \begin{minipage}{0.95\linewidth}
          \centering
          \includegraphics[width=\linewidth]{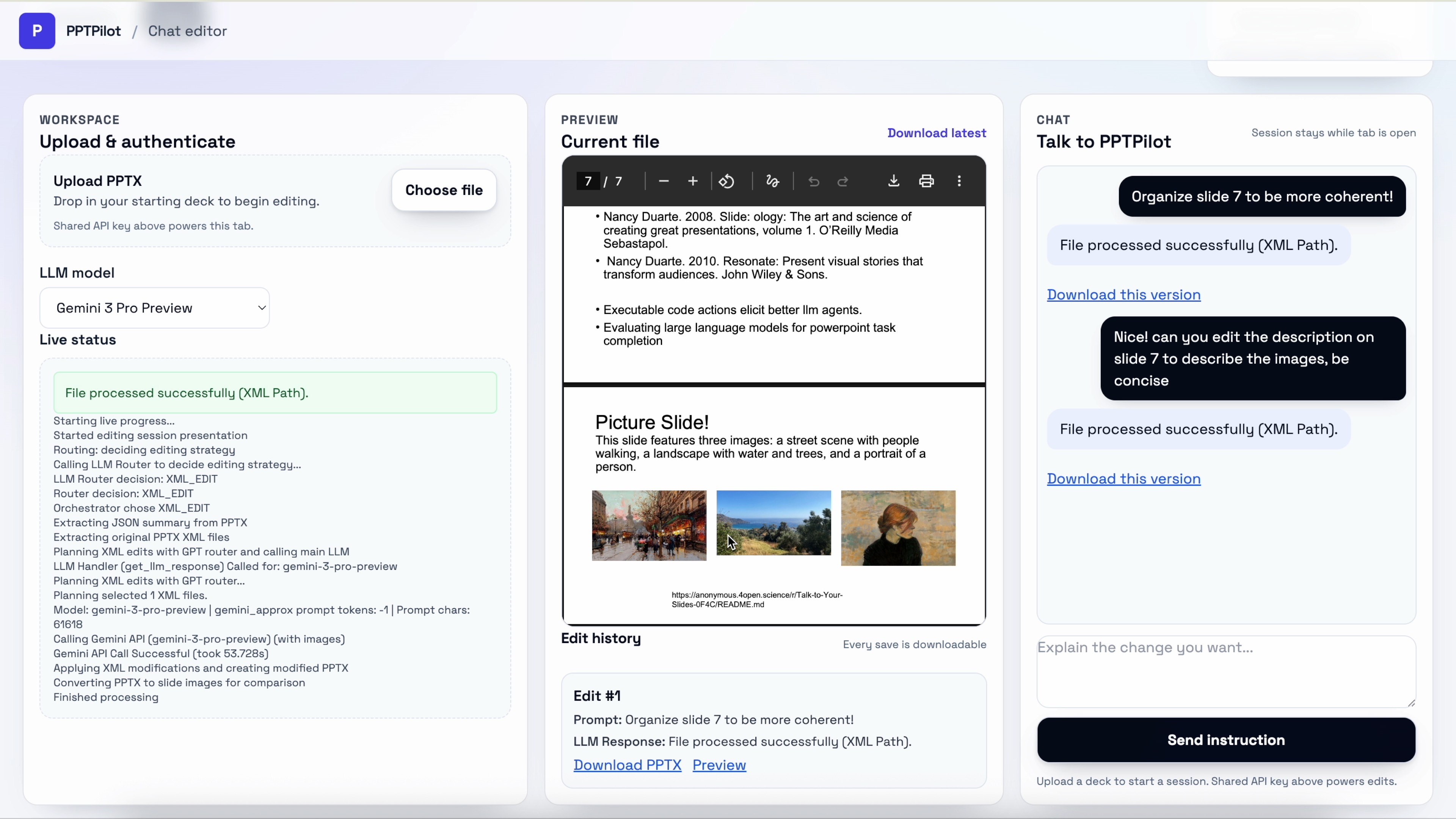}
          
          {\small Chat-based Editing}
      \end{minipage}
      \hfill
      \begin{minipage}{0.95\linewidth}
          \centering
          \includegraphics[width=\linewidth]{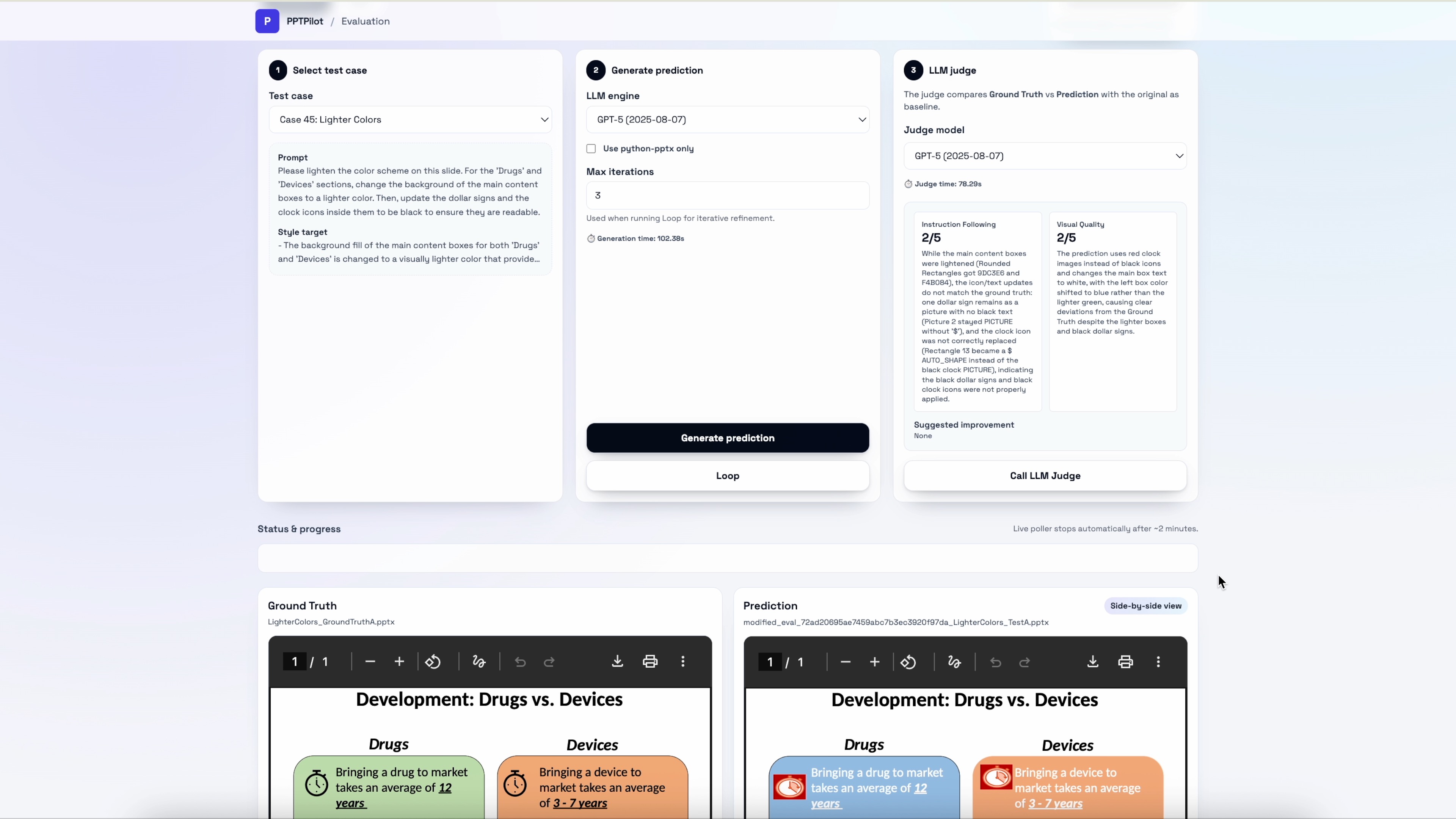}
          
          {\small PPTArena Evaluation}
      \end{minipage}
      \caption{Our WebApp Demo for PPTPilot in chat-based editing (Top) and PPTArena evaluation (bottom). See demo videos for better visualization.}
      \label{fig:pptpilot_demo_screenshots}
  \end{figure}

\begin{table*}[!htbp]
\centering
\small
\setlength{\tabcolsep}{6pt}
\renewcommand{\arraystretch}{1.2}
\caption{{\bf VLM-as-Judge Evaluation Results (Gemini-3.1-Pro)}. Scores report instruction following (IF) and visual quality (VQ) with VLM-as-judge. Columns marked with $^{*}$ are only run on a 25-case subsample for cost and rate-limit reasons. Bracketed values in the PPTPilot column report scores on that same subsample.}

\label{tab:gemini2.5proasEvals}
\resizebox{\linewidth}{!}{%
\begin{tabular}{l r
    >{\columncolor{cyan!9}}c >{\columncolor{cyan!9}}c  
    >{\columncolor{gray!12}}c   >{\columncolor{gray!12}}c    
    >{\columncolor{orange!12}}c   >{\columncolor{orange!12}}c    
    >{\columncolor{green!8}}c   >{\columncolor{green!8}}c    
    >{\columncolor{purple!10}}c >{\columncolor{purple!10}}c  
    c c  
    c c  
}
\toprule
& & \multicolumn{2}{c}{\textbf{PPTPilot}}
  & \multicolumn{2}{c}{\textbf{Gemini CLI}}
  & \multicolumn{2}{c}{\textbf{ChatGPT}}
  & \multicolumn{2}{c}{\textbf{ChatGPT Agent$^{*}$}}
  & \multicolumn{2}{c}{\textbf{MiniMax Agent$^{*}$}}
  & \multicolumn{2}{c}{\textbf{PPTAgent$^{*}$}}
  & \multicolumn{2}{c}{\textbf{Paper2Poster$^{*}$}} \\
\cmidrule(lr){3-4}\cmidrule(lr){5-6}\cmidrule(lr){7-8}\cmidrule(lr){9-10}\cmidrule(lr){11-12}\cmidrule(lr){13-14}\cmidrule(lr){15-16}
\textbf{Category} & \textbf{Cases} & IF$\uparrow$ & VQ$\uparrow$ & IF$\uparrow$ & VQ$\uparrow$ & IF$\uparrow$ & VQ$\uparrow$ & IF$\uparrow$ & VQ$\uparrow$ & IF$\uparrow$ & VQ$\uparrow$ & IF$\uparrow$ & VQ$\uparrow$ & IF$\uparrow$ & VQ$\uparrow$ \\
\midrule
Content      & 67 &
\textbf{2.51}\hspace{0.15em}[\textbf{2.10}] & \textbf{2.68}\hspace{0.15em}[\textbf{2.35}] &
1.93 & 2.14 &
1.95 & 2.00 &
1.45 & 1.65 &
0.90 & 0.80 &
0.00 & 0.00 &
0.00 & 0.00 \\
Layout       & 29 &
\textbf{2.18}\hspace{0.15em}[\textbf{2.00}] & \textbf{2.64}\hspace{0.15em}[\textbf{2.14}] &
1.89 & 2.07 &
1.96 & 2.00 &
0.86 & 0.86 &
0.86 & 0.57 &
0.00 & 0.00 &
0.00 & 0.00 \\
Styling      & 29 &
\textbf{2.54}\hspace{0.15em}[\textbf{2.17}] & \textbf{2.50}\hspace{0.15em}[\textbf{2.33}] &
2.17 & 2.07 &
2.37 & 1.85 &
0.83 & 0.83 &
1.00 & 0.83 &
0.00 & 0.00 &
0.00 & 0.00 \\
Structure    & 15 &
\textbf{2.47}\hspace{0.15em}[1.67] & \textbf{3.60}\hspace{0.15em}[\textbf{2.33}] &
1.87 & 2.33 &
1.80 & 1.53 &
\textbf{2.00} & 1.67 &
0.67 & 0.67 &
0.00 & 0.00 &
0.00 & 0.00 \\
Interactivity& 4 &
2.00\hspace{0.15em}[\textbf{2.00}] & 3.00\hspace{0.15em}[\textbf{2.00}] &
1.00 & 2.25 &
\textbf{2.25} & \textbf{4.25} &
1.00 & \textbf{2.00} &
0.00 & 0.00 &
0.00 & 0.00 &
0.00 & 0.00 \\
\midrule
All Cases    & 100 &
\textbf{2.45}\hspace{0.15em}[\textbf{2.04}] & \textbf{2.74}\hspace{0.15em}[\textbf{2.60}] &
1.92 & 2.15 &
1.97 & 2.03 &
1.44 & 1.32 &
0.92 & 0.80 &
0.00 & 0.00 &
0.00 & 0.00 \\
\bottomrule
\end{tabular}}

\end{table*}

\section{PPTPilot Analysis}

\subsection{Gemini 3.1 Pro as Judge}

In our main paper, we utilize GPT as the VLM judge for PPTArena. To validate the robustness of our PPTPilot, we conduct a further experiment using Gemini 3.1 Pro as the VLM judge. As shown in Table~\ref{tab:gemini2.5proasEvals}, we conduct the same set of experiments as in Table~\ref{tab:pptarena_full_gpt5} and highlight the following observations. As clearly shown, PPTPilot still shows a significant advantage over other approaches, including the proprietary agents. This suggests that our PPTPilot is simple yet effective. \Cref{fig:gemini_barplots} visualizes the alignment between scores assigned by different VLM judges. The comparison demonstrates that while absolute score values may vary slightly between judges, the relative ranking of agents remains consistent, with PPTPilot maintaining its lead across both the full benchmark and the challenging subset.

\subsection{PPTPilot on T2US}

In Table~\ref{tab:pppilot-vs-t2us}, we compare PPTPilot on the T2US~\cite{jung2025talkslideslanguagedrivenagents} benchmark. We adopt \textit{gemini-3.0-flash} as our backbone model, the same as T2US, for a fair comparison. As clearly demonstrated, our PPTPilot has a significant advantage over T2US, achieving better quality, exceeding a score of 4 out of 5 in multiple categories. In fact, this also demonstrates the need for our PPTArena to introduce more challenging PPT editing scenarios. 

\begin{table}[t]
\centering
\caption{\textbf{PPTPilot vs.\ T2US on the T2US benchmark.}
We report success rate and judge ratings for different quality dimensions.}

\label{tab:pppilot-vs-t2us}
\resizebox{0.7\columnwidth}{!}{%
\begin{tabular}{lcc}
\toprule
\multirow{2}{*}{Metric} & \multicolumn{2}{c}{T2US Benchmark} \\
\cmidrule(lr){2-3}
 & T2US~\cite{jung2025talkslideslanguagedrivenagents} & PPTPilot (ours) \\
\midrule
Success rate (\%)              & 96.83          & \textbf{100.00} \\
Instruction following $\uparrow$ & 2.21           & \textbf{4.05}   \\
Image quality $\uparrow$         & 2.17           & \textbf{4.44}   \\
Layout quality $\uparrow$        & 2.58           & \textbf{4.40}   \\
Color quality $\uparrow$         & 2.57           & \textbf{4.20}   \\
Text quality $\uparrow$          & 2.48           & \textbf{3.92}   \\
\bottomrule
\end{tabular}%
}

\end{table}

\section{PPTArena Analysis}

\subsection{Comparison with Related Benchmarks}

In Table~\ref{tab:benchmark_landscape}, we compare our PPTArena with existing PPT-related benchmarks and highlight its uniqueness. As clearly demonstrated, our PPTArena provides broader coverage of scenarios, does not rely on predefined APIs or COMs, and is based on a larger number of source slides. In addition, our evaluation protocol is carefully designed to enable rigorous, reliable evaluation by providing detailed ground truth, predictions, and per-sample style-targets for reproducible scoring.

Notably, prior benchmarks, \emph{e.g.}, PPTC-R, might not be challenging enough given the progress of VLMs. Specifically, we evaluate GPT-5 and Gemini 3.1 Pro using PPTC-R's pipeline and achieved the success rates of 92\% and 88\%, respectively. These observations further suggest the necessity of building a challenging PPT editing benchmark like our PPTArena.

\begin{table*}[t]
    \centering
    \small
    \setlength{\tabcolsep}{6pt}
    \renewcommand{\arraystretch}{1.1}
    \caption{\textbf{Benchmark and evaluation coverage comparison.}
    We contrast PPTArena with prior PowerPoint editing / generation benchmarks in terms of scenario focus and what aspects of evaluation they make observable vs.\ leave under-specified.}
    
    \label{tab:benchmark_landscape}
    \resizebox{\linewidth}{!}{%
    \begin{tabular}{l p{0.30\linewidth} p{0.47\linewidth}}
    \toprule
    \textbf{Benchmark} & \textbf{Scenario focus} & \textbf{Evaluation coverage and gaps} \\
    \midrule
    PPTC-R~\cite{zhang-etal-2024-pptc}
    & Macro playback of $<\!100$ API functions on templated decks
    & Regenerates decks via Office macros and checks API diffs, but does not release PPTX pairs or manifests. Outputs can vary by Office build, and there is no direct semantic or visual quality judgment. \\
    \addlinespace[2pt]
    TSBench (Talk to Your Slides)~\cite{jung2025talkslideslanguagedrivenagents}
    & COM-scripted instruction following on 379 prompts over corporate templates
    & Lacks dual scoring and public reference slides; edits rarely exceed four operations per case, and 379 prompts are extrapolated from 56 underlying edit tasks, limiting diversity and compositional difficulty. \\
    \addlinespace[2pt]
    Paper2Poster~\cite{pang2025paper2postermultimodalposterautomation}
    & Single-slide poster generation from research papers
    & Uses a QA pipeline plus a single VLM-as-judge. Scoring is coarse, with no structured manifests or per-operation fidelity, and the benchmark is not designed for multi-step editing or layout-preserving updates. \\
    \addlinespace[2pt]
    AutoPresent~\cite{ge2025autopresentdesigningstructuredvisuals}
    & SlidesBench: prompts derived from existing slides for PPT generation from scratch
    & Reports spatial/text/color metrics and heuristic checks, but reference decks and judge rationales are not publicly released (subscription only), and the reported test set contains just 10 PowerPoint decks. \\
    \addlinespace[2pt]
    PPTAgent~\cite{pptagent2025}
    & Document-to-slide synthesis via a hierarchical presentation agent
    & PPTEval uses a single LLM to rate content/design/coherence, without deterministic manifests, accessibility checks, or explicit visibility into which elements were inserted, modified, or preserved. \\
    \midrule
    \textbf{PPTArena (ours)}
    & 100-case, 2{,}125-slide multi-edit benchmark covering 16 edit types across 5 categories
    & Releases Original / Prediction / Ground-truth PPTX triplets, style-target manifests, and dual VLM judges: IF (instruction-following via structured JSON/XML diffs) and VQ (visual quality via screenshots). Supports reproducible, traceable scoring, hybrid API/XML evaluations, and iterative verification under realistic multi-edit workloads. \\
    \bottomrule
    \end{tabular}}
\end{table*}

\section{PPTArena Dataset Construction}
\label{app:dataset}

We detail the multi-stage construction of PPTArena, covering data sourcing, curation, and the extraction pipeline to derive rubrics for evaluation illustrated in \Cref{lst:pptx_to_json}.

\subsection{Data Sources and Licensing}

\begin{itemize}
    \item \textbf{Zenodo}: We sourced open-access presentations from Zenodo repositories.
    \item \textbf{Government and Educational Sources}: We collected PPTs from .gov and .edu domains to ensure a variety of formal and academic styles.
    \item \textbf{Web Scraping}: We utilized targeted search queries to find relevant financial and business presentations, such as:
    \begin{quote}
        \texttt{`quarter revenues' filetype:pptx}
    \end{quote}
    \item \textbf{Creative Licenses}: We included high-quality templates and decks from Slideshare and SlidesCarnival that are available under Creative Commons Licenses.
    \item \textbf{Student Contributions}: We curated presentations from students in diverse fields like Biology, Art Practice, and Chemical Engineering to capture different disciplinary norms.
\end{itemize}

\subsection{PPT and Scenario Curation}

\subsubsection{Dataset Scale and Diversity}
Agentic editing benchmarks differ fundamentally from generation datasets. In text-to-slide or slide generation settings, one can often scale data through prompt variation or template expansion. PPTArena instead requires causal, element-level ground truths for in-place modifications, where each case must specify not only the source deck and instruction, but also a precise edited target that preserves document structure and supports reliable diff-based evaluation. This makes benchmark construction substantially more demanding than collecting static presentation decks or synthetic prompt completions.

PPTArena therefore emphasizes meaningful task scale rather than raw template volume. The benchmark contains 100 curated decks with 2,125 slides and more than 1,300 human-curated edits, covering 16 edit types across 5 major categories, including text, chart, animation, layout, structural, and master-style changes. These cases include both local edits and long-horizon scenarios with cross-slide dependencies, deck-wide consistency requirements, and style transfer constraints. We view this depth as especially important for agentic evaluation, since realistic presentation editing depends less on isolated single-slide changes and more on multi-step reasoning over structured documents.

Diversity in PPTArena also operates at multiple levels. The final benchmark spans professional, academic, creative, and multilingual materials, with substantial variation in visual structure, multimodal content, and editing goals. Rather than scaling through repeated templates or paraphrased prompts, PPTArena scales through heterogeneous source decks, realistic edit operations, and challenging document-level workflows.

\subsubsection{Scenario Curation}

\mypar{PPT Selection.} In our initial sourcing, we gathered a database with more than 18k PPTs. To identify decks suitable for a high-quality evaluation benchmark, we first convert each PPT file into a structured JSON representation and retain only decks with multimodal elements such as images, tables, charts, and customized themes. We then rank the remaining PPTs by file size, number of slides, and content variety, yielding the top 500 most diverse slide decks. Finally, a team of human annotators manually inspects these candidates and selects the final 100 PPTs based on three criteria: (1) \emph{Quality}: rejecting decks with broken layouts, low-resolution images, or unreadable text; (2) \emph{Privacy}: removing decks containing personal information; and (3) \emph{Complexity}: prioritizing slides with challenging elements such as complex charts or grouped shapes. This process produces a high-quality and structurally diverse set of source decks for evaluating realistic editing tasks.

\mypar{Editing Query Creation.} We construct queries through a combination of AI-generated prompts and human-inspired edits. For each category, we prompt multiple AI models to generate high-fidelity, multi-step, and reasoning-intensive tasks. We manually validate more than 200 candidate prompts, then iteratively match and refine them against selected PPTs, resulting in a curated set of 100 benchmark cases.

\mypar{Distribution.} We ensure that all categories in our benchmark are well represented, with at least two cases for every specific edit type. The benchmark is intentionally weighted toward edit types that are both common and challenging in real-world presentation workflows. As shown in Fig.~\ref{fig:pptarencases}, PPTArena covers a broad range of requests and visual elements.

\mypar{Difficulty Labeling.} For each case, we record four aligned signals: (1) the number of atomic operations in the ground-truth program, (2) the number of distinct slides edited, (3) whether the instruction or style target imposes cross-slide dependencies, and (4) a \emph{high-diff} flag. We mark a case as high-diff when it involves cross-slide dependencies or strong visual-textual semantic reasoning, such as chart remapping or translation. These labels produce the difficulty distribution summarized in Table~\ref{tab:difficulty_by_category_main} of the main paper.

\subsection{Ground-Truth Deck Creation}
Our benchmark is derived from real-world slide decks (domain-specific presentations or web-sourced) with 20\%  containing \texttt{python-pptx} manipulations to provide a controlled baseline. For the real-world decks, expert human annotators manually perform edits to create the ground-truth target, with each case typically requiring more than two hours of work. This ensures that the target state reflects high-quality, human-level design decisions. For the 20 synthetic cases, the ground-truth PPTs are generated programmatically, providing a reliable reference.

\subsection{Subset Evaluation Data Setup}
By default, we evaluate agents on the full benchmark. However, because several proprietary agents are constrained by strict rate limits and high usage costs, we also curate a subset for budget-sensitive evaluation. Specifically, we define a ``hard'' subset of 25 cases intended to capture the most difficult scenarios in PPTArena while preserving breadth across task types. We select 20 cases that we believe represent the most difficult and complex cases, containing large diffs, many operations, and xml-based edits that agents generally struggle the most on, and an additional 5 for breadth. In total, this matched subset comprises 206 edits across the 25 decks, and every subset-only system reported in the matched comparison of the main paper (Table~\ref{tab:pptarena_subset25_gpt5}) is evaluated on exactly these 206 edits.

\section{Evaluation and VLM Judges}

This section provides further details for our evaluation of VLM judge.

\subsection{Per-Sample Style-Target Generation}

The essential part of our reliable evaluation is the generation of a per-sample style-target to capture the varied requirements of different edit queries. We first use GPT-5 to generate initial style targets, along with detailed JSON summaries of the original and ground-truth PPTs. Then, human annotators manually refined the style targets to ensure they accurately captured the nuances of the transformation. In our style targets, they emphasize the following rubrics: (1) \emph{Content}: Accuracy of text updates and data; (2) \emph{Layout}: Alignment, spacing, and grid adherence; (3) \emph{Typography}: Font consistency and hierarchy; (4) \emph{Global Constraints}: Theme application and master slide usage. The exact prompt templates of prompting the language models are shown in \Cref{lst:style_prompt,lst:smart_diff}.

\subsection{VLM-as-Judge Evaluation}
\label{app:judge}

We employ a dual-model judging system to evaluate agent performance, designed to balance deterministic evaluation with semantic understanding. The primary judge is \textbf{GPT-5.2}, configured with temperature 0.2 and top-$k$=1 to ensure consistent outputs. For robustness and cross-verification, we also utilize \textbf{Gemini 3.1 Pro} as a secondary judge, as shown in Table~\ref{tab:gemini2.5proasEvals}. The evaluation pipeline constructs a composite prompt by concatenating the user's original instruction with the explicit style target. This prompt is provided to the VLM judge, along with two distinct modalities of the presentation state, to assess instruction-following and visual quality. Full prompt templates are provided in \Cref{lst:if_prompt,lst:vq_prompt}.

\mypar{Instruction Following.} To evaluate instruction following, we build the contexts for the VLM judge so that it concentrates on the contents. Accordingly, the inputs include: (1) \emph{Structured Diffs}: This allows the judge to precisely verify if specific requested actions (\emph{e.g.}, ``change font to Arial'') were executed; (2) \emph{XML \& JSON}: We enable the judge to understand the XML and JSON diffs. Depending on the types of editing queries, we select the optimal way to leverage the VLM judge's context lengths. We do this by calculating the percentage of the diff between the XMLs and JSONs, and then sending the one corresponding to the higher percentage change to the LLM. We see that generally master and theme edits flag XML diffs, while many textual replacements flag the JSONs.

\mypar{Visual Quality.} The input for evaluating visual quality aims at providing the multi-modal information for the VLM judge. The primary input is the set of high-resolution \emph{visual screenshots}. To improve the reliability of the VLM judge, we implement several heuristic mechanisms: (1) \emph{SSIM Screening}: An editing query might only influence a small portion of slides, making the evaluation of all of the slides redundant. Therefore, we can utilize the Structural Similarity Index (SSIM) to automatically mark certain slides as correct if the SSIM between the prediction and ground-truth is high. (2) \emph{Context Lengths}: To avoid the challenges of long context lengths, we group the screenshots into batches of 5 slides per VLM judge inference, enabling them to look closely into every screenshot.

\mypar{Validating batched VQ scoring.}\label{app:pagewise-vq} To confirm that batching screenshots (five per inference) does not distort visual-quality scores relative to judging slides one at a time, we ran a page-wise control study: an independent Gemini~3.1 Pro VQ judge scored 105 matched slide pairs one page at a time, which we compare against the batched scores on the same pairs. The two regimes agree within one point on 8 of 10 cases (mean VQ $1.88$ page-wise vs.\ $2.00$ batched; Pearson $r=0.917$), indicating that batching preserves the VQ signal while substantially reducing the number of judge calls.


\subsection{Ablation Study}
\label{sec:ablation}

We conduct a systematic ablation study of PPTPilot's key components to understand their individual contributions to overall performance. We evaluate each variant using the metrics of task success rate, PPTX validity, instruction fidelity (IF), visual quality (VQ), and computational cost.

\mypar{Experimental setup.} Our baseline configuration uses the full PPTPilot pipeline: JSON snapshot construction, intelligent routing between XML editing and programmatic \texttt{python-pptx} paths, strict output schemas for LLM responses, and post-hoc XML validation with automatic repair. We systematically disable or replace individual components while holding all other factors constant.

\mypar{Ablation variants.} We evaluate four key configurations:

\noindent\textit{XML-only:} Forces all edits through the direct XML manipulation path. This approach excels at precise, slide-local modifications (\eg, adjusting a single shape's geometry) but struggles with repetitive deck-wide operations such as bulk text normalization or global theme application, resulting in increased latency and reduced task success on global transformations.

\noindent\textit{\texttt{python-pptx}-only:} Routes all requests through the programmatic API. This variant handles global changes effectively (uniform typography, batch renaming, translation) but underperforms on precise structural fixes requiring fine-grained control over individual elements.

\noindent\textit{Hybrid (no refinement):} Uses intelligent routing but disables iterative refinement. This reduces latency but sacrifices robustness on multi-step transformations where a second pass typically resolves residual formatting issues.

\noindent\textit{Hybrid + Loop (3$\times$):} The full system with up to three refinement iterations.

\mypar{Results and observations.} Table~\ref{tab:pptpilot_ablation_main} presents the ablation results. The hybrid routing strategy proves essential: forcing a single execution path substantially degrades task success, particularly on workloads mixing local and global edits. The XML-only variant achieves an IF score of 0.95, while \texttt{python-pptx}-only reaches 2.06. The hybrid approach without refinement achieves 2.36, and enabling iterative refinement further improves performance to 2.84 IF and 3.21 VQ. Case-by-case analysis reveals that providing slide images to PPTPilot improves layout reasoning: visual grounding enhances decisions about z-ordering, spacing, and alignment without compromising text editing quality. Iterative refinement stabilizes complex edits, with one additional pass often resolving formatting drift after re-theming or batch rewrites, at modest computational overhead. Empirically, most corrections occur by the second pass while later iterations plateau, which is why we cap refinement at three passes as a favorable accuracy--latency trade-off.

\mypar{Judge configuration analysis.} The top block of Table~\ref{tab:pptpilot_ablation_main} compares three evaluation strategies: a single-call VLM judge receiving all modalities simultaneously, a single-call judge using only slide images, and our dual-judge configuration with explicit diff analysis. The dual-judge approach with structural diffs proves substantially more stable on multi-edit cases, as the instruction-following judge can directly reason over structured manifests rather than inferring changes from pixel-level comparisons alone. Other configurations are either inconsistent or over-lenient in one or more aspects. In a single judge format, a difference of 33 percent between visual quality and instruction following is un-reliable and shows that there isn't coherence between the two. Without presenting diffs, the judges are overloaded with long-horizon tokens and their context-windows are often maxed and they miss details and errors more easily.

\section{Examples of PPTArena's Challenging Cases}
\label{app:benchmark-comparison}

\begin{figure*}[!tbp]
    \centering
    \begin{minipage}{0.98\linewidth}
    \setlength{\fboxsep}{0pt}
    
    \rule{0.98\linewidth}{0.5pt}

    \begin{tabular}{@{}p{0.48\linewidth}p{0.48\linewidth}@{}}

    \begin{minipage}[t]{\linewidth}
    \centering
        \textbf{\small Object Distance (16 slides)}\\
        \begin{minipage}[t]{0.48\linewidth}
            \centering
            {\small Original}\\[0.3em]
            \includegraphics[page=1,width=\linewidth]{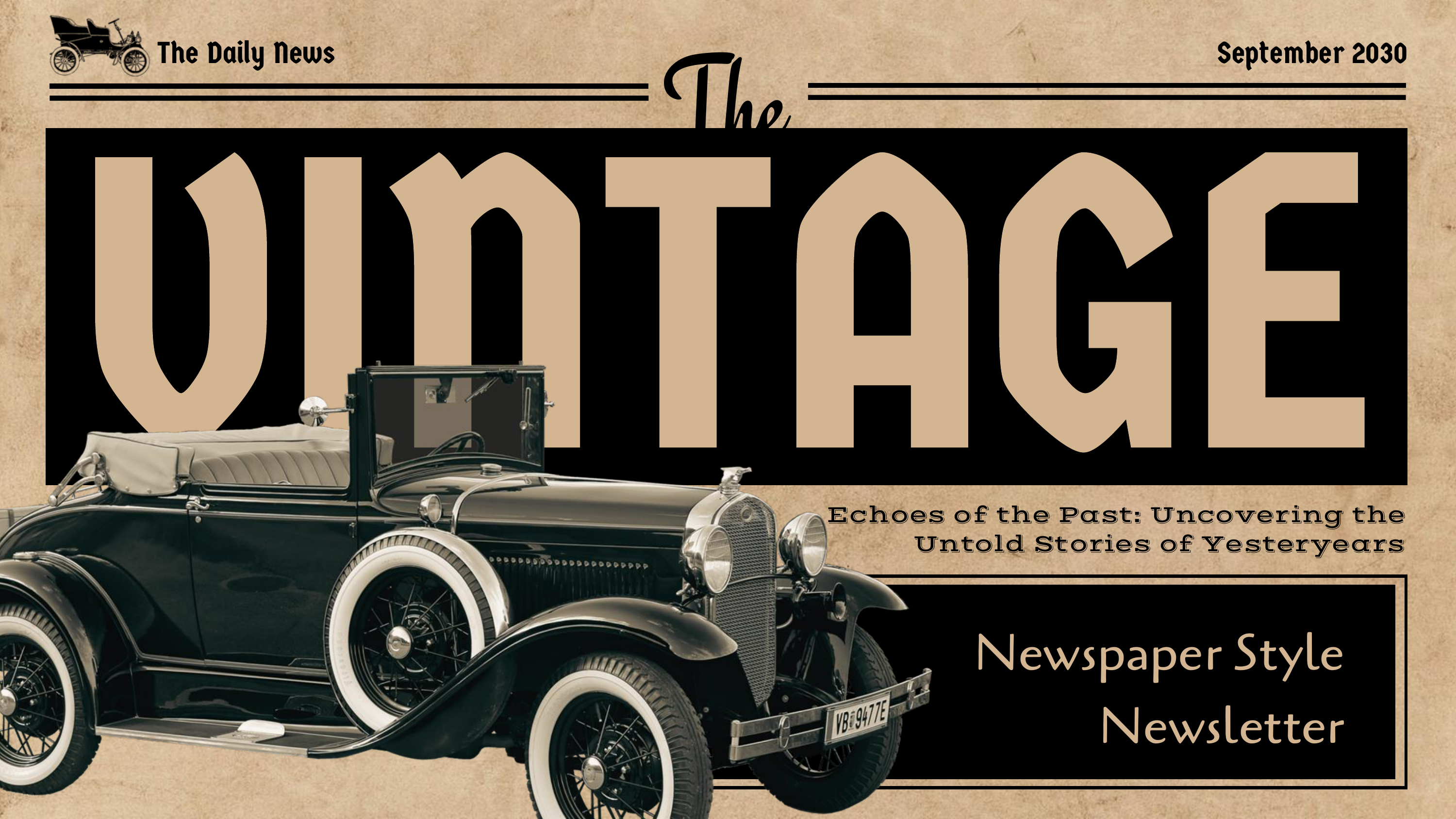}\\[0.2em]
            \includegraphics[page=14,width=\linewidth]{figures/ObjectDistance_TestA.pdf}
        \end{minipage}%
        \hfill
        \begin{minipage}[t]{0.48\linewidth}
            \centering
            {\small Ground Truth}\\[0.5em]
            \includegraphics[page=1,width=\linewidth]{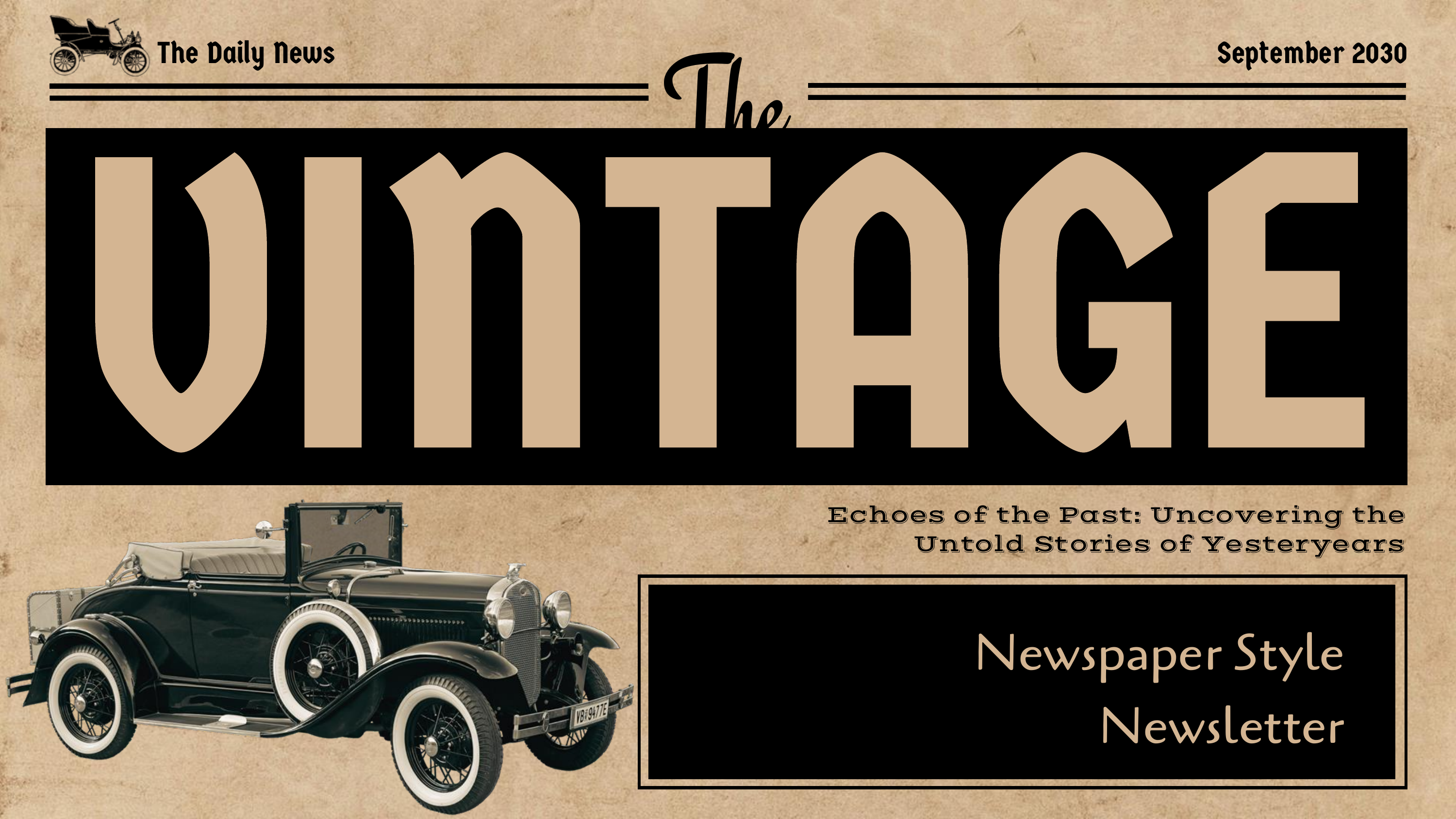}\\[0.2em]
            \includegraphics[page=14,width=\linewidth]{figures/ObjectDistance_GTA.pdf}
        \end{minipage}
    \end{minipage}
    &
    \begin{minipage}[t]{\linewidth}
    \centering
        \textbf{\small Update Theme (28 slides)}\\
        \begin{minipage}[t]{0.48\linewidth}
            \centering
            {\small Original}\\[0.3em]
            \includegraphics[page=22,width=\linewidth]{figures/updatetheme_TestA.pdf}\\[0.2em]
            \includegraphics[page=27,width=\linewidth]{figures/updatetheme_TestA.pdf}
        \end{minipage}%
        \hfill
        \begin{minipage}[t]{0.48\linewidth}
            \centering
            {\small Ground Truth}\\[0.5em]
            \includegraphics[page=22,width=\linewidth]{figures/updatetheme_GroundTruthA.pdf}\\[0.2em]
            \includegraphics[page=27,width=\linewidth]{figures/updatetheme_GroundTruthA.pdf}
        \end{minipage}
    \end{minipage}
    \\[1.2em]

    \begin{minipage}[t]{\linewidth}
    
    \centering
        \textbf{\centering\small Chart Conversions (5 slides)}\\
        \begin{minipage}[t]{0.48\linewidth}
            \centering
            {\small Original}\\[0.3em]
            \includegraphics[page=1,width=\linewidth,trim=0cm 1.8cm 0cm 0cm,clip]{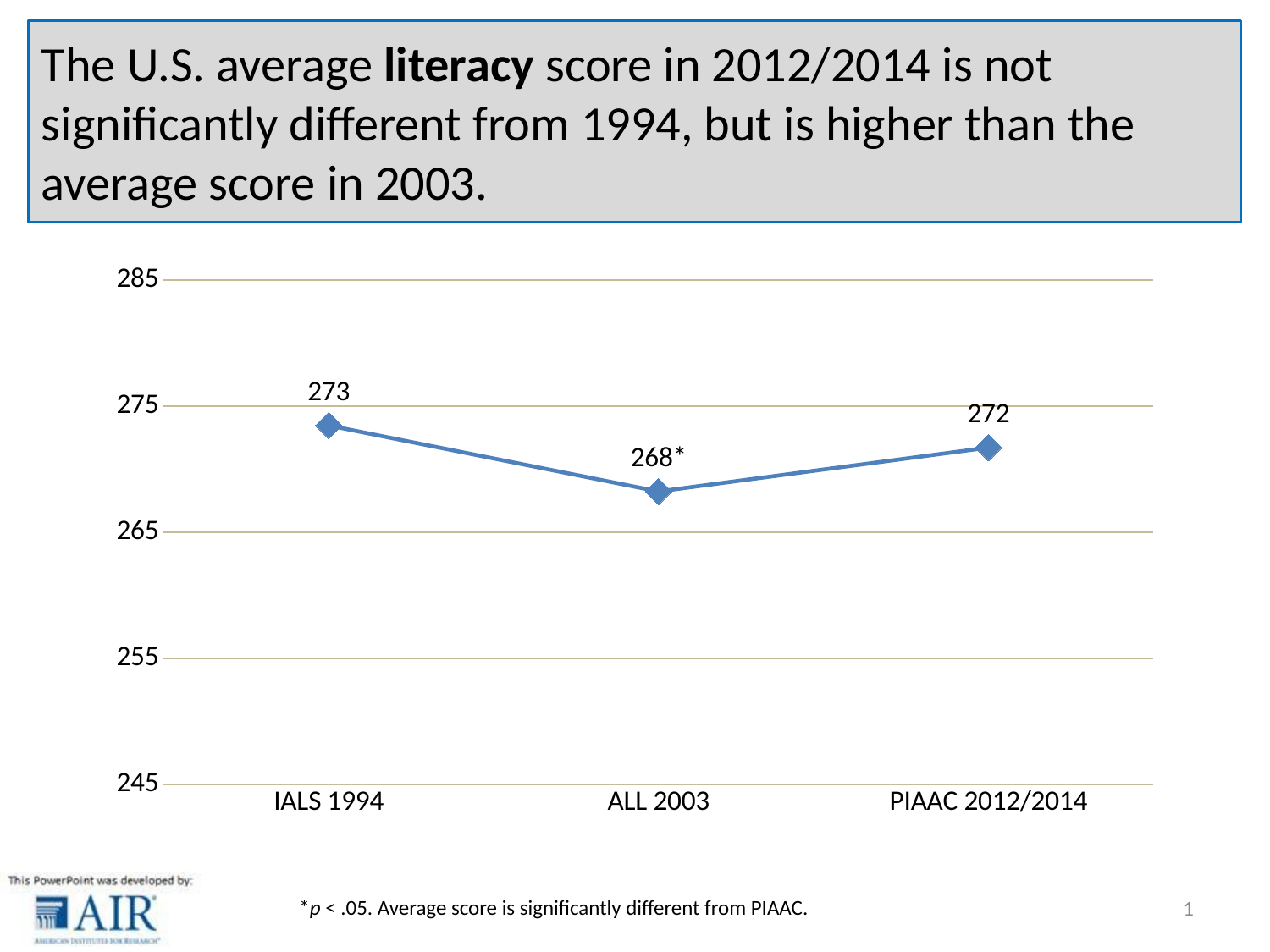}\\[0.2em]
            \includegraphics[page=3,width=\linewidth,trim=0cm 1.5cm 0cm 1cm,clip]{figures/PieChart_TestA.pdf}
        \end{minipage}%
        \hfill
        \begin{minipage}[t]{0.48\linewidth}
            \centering
            {\small Ground Truth}\\[0.3em]
            \includegraphics[page=1,width=\linewidth,trim=0cm 1.8cm 0cm 0cm,clip]{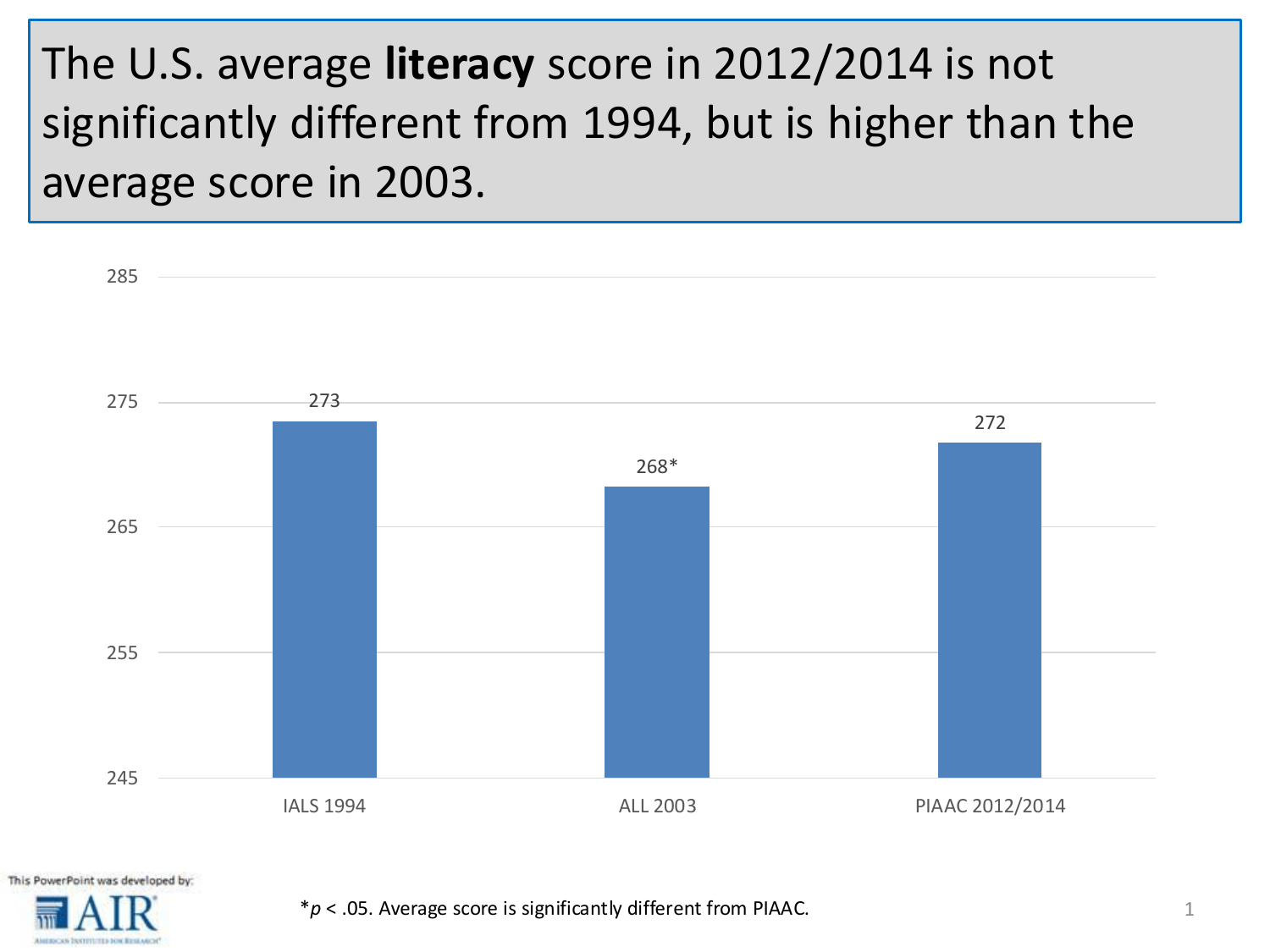}\\[0.2em]
            \includegraphics[page=3,width=\linewidth,trim=0cm 1.5cm 0cm 1cm,clip]{figures/PieChart_GroundTruthA.pdf}
        \end{minipage}
    \end{minipage}
    &
    \begin{minipage}[t]{\linewidth}
    
    \centering
        \textbf{\small Translate Arabic LTR (22 slides)}\\
        \begin{minipage}[t]{0.48\linewidth}
            \centering
            {\small Original}\\[0.3em]
            \includegraphics[page=5,width=\linewidth]{figures/TranslateArabicLTR_TestB.pdf}\\[0.2em]
            \includegraphics[page=17,width=\linewidth]{figures/TranslateArabicLTR_TestB.pdf}
        \end{minipage}%
        \hfill
        \begin{minipage}[t]{0.48\linewidth}
            \centering
            {\small Ground Truth}\\[0.3em]
            \includegraphics[page=5,width=\linewidth]{figures/TranslateArabicLTR_GroundTruthB.pdf}\\[0.2em]
            \includegraphics[page=17,width=\linewidth]{figures/TranslateArabicLTR_GroundTruthB.pdf}
        \end{minipage}
    \end{minipage}
    \end{tabular}
    
    \rule{0.98\linewidth}{0.5pt}
    \end{minipage}
    
    \caption{Example edit queries in PPTArena, including correcting object distances, changing plot styles, updating PPT themes, and translating the contents. As demonstrated, our PPTArena contains a wide range of slide decks with varied visual elements and content. PPT editing operations are also typically spread across multiple pages, underscoring the challenge of reliable PPT editing in the real world.}
    \label{fig:pptarencases}
    
\end{figure*}



\subsection{Multi-Step Reasoning Depth}

Unlike PPTC-R's focus on adversarial variations of single instructions (e.g., translating text with noisy phrasing or API constraints), PPTArena requires agents to decompose complex instructions into multiple interdependent sub-tasks. Consider the case \textbf{Multi-Edit Cascade} in figure~\ref{fig:appendix_cases_74_93_98}. This single instruction requires: (1) global theme application across all slides, (2) image-to-chart conversion with data extraction, (3) layout optimization on a specific slide, (4) programmatic progress bar generation with positional calculations, and (5) bibliography synthesis from scattered references. By contrast, the longest PPTC-R long-turn we measured strings together 29 deterministic API invocations across nine slides (largely repeating \texttt{move\_to\_slide} $\rightarrow$ \texttt{set\_font\_*} / \texttt{set\_background\_color}) without ever mixing modalities or reconciling content semantics. T2US similarly focuses on isolated operations like typo correction or translation, where the reasoning depth rarely exceeds two steps. 

Another example is \textbf{12-Column Grid \& Baseline Rhythm Canonicalization} which exemplifies another level of complexity with the prompt and style target shown in~\Cref{fig:appendix_case75_example}. This case requires the agent to: (1) infer an implicit 12-column grid structure from messy layouts, (2) compute baseline alignments across heterogeneous text boxes, (3) resolve chart-legend collisions through spatial reasoning, (4) establish image-caption groupings while maintaining visual balance, and (5) perform self-referential text updates that describe the corrections. The last requirement—updating descriptive text to match the new state—introduces a meta-cognitive challenge absent from prior benchmarks, where ground truth never depends on the agent's own edits.

\subsection{Cross-Slide Dependencies and Global Constraints}
Some tasks in PPTArena force long-horizon planning because mistakes in early slides cascade to the remaining ones. The 
\textbf{Cross-Slide Data Consolidation} example in figure~\ref{fig:appendix_cases_74_93_98} pushes this further: the agent must (1) parse and merge structured data from two slides, (2) delete slides while updating all subsequent slide references, (3) apply typographic transformations globally, and (4) perform layout normalization on a different slide. This creates a dependency graph where earlier actions (slide deletion) affect later operations (slide indexing). Although the PPTC-R release contains 386 turns (21\% of 1,808) that touch more than one slide, each loop simply replays the same formatting adjustment after a \texttt{move\_to\_slide} call, so there is no dependency between slides or data flow to maintain.

Another example, \textbf{Cross-Slide Conditional Formatting}, in figure~\ref{fig:appendix_cases_74_93_98} introduces conditional logic across slides. This requires: (1) parsing tabular data on one slide, (2) establishing semantic correspondences between table entries and timeline elements on another slide, and (3) applying conditional formatting based on extracted attributes. No comparable case exists in PPTC-R or T2US, whose multi-slide turns remain independent formatting loops without shared semantics.

\subsection{Semantic Understanding and Multi-Modal Reasoning}

PPTArena includes 18 cases requiring deep semantic understanding of content. The example shown in the main paper, \textbf{Fill in the Animal Research Poster} shown on the first paper illustrates this: the agent must (1) semantically parse section headers to understand topical context (e.g., ``THE EYES'' expects information about visual systems), (2) retrieve factual knowledge about specific animals, (3) produce context-appropriate summaries that fit space constraints, (4) optimize font sizes to prevent overflow, and (5) format academic citations. Another example is a case that asks for a format change;
\textbf{Screenshot-to-Editable Text} requires vision-language integration. The agent must: (1) perform OCR on the screenshot, (2) extract visual elements (the university crest), (3) reconstruct the layout semantically rather than pixel-perfectly, and (4) ensure all elements are editable PowerPoint objects, not images.

\subsection{Long-Horizon Planning and Accessibility}

Our case that asks to \textbf{Update Theme and Slides Backgrounds} shown in Figure~\ref{fig:pptarencases} spans 27 slides and requires: creating custom master layouts, classifying slides into categories, applying different templates, enforcing contrast, and preserving existing content during a theme change. The longest PPTC-R released turn touches nine slides but only issues local typography commands, never rethinking layout intent or accessibility structure.

Accessibility requirements introduce another layer of difficulty. One of our non-visual dependent examples titled \textbf{WCAG Accessibility \& Master Cleanup} asks the agent to (1) generate semantically meaningful slide titles based on content analysis, (2) reorder shape z-indices to match logical reading flow, (3) remove font overrides while preserving visual appearance, and (4) validate against WCAG 2.1 AA standards.

\section{PPTPilot Implementation Details}
\label{app:pptpilot}

This section provides the implementation details of our simple yet effective PPTPilot. The prompt templates and example code for the parts below are in \Cref{lst:pptx_to_json,lst:smart_diff}.


\subsection{Skill Router and Editing Pipeline}

\mypar{Skill Router.} Different PPTs might require different kinds of edits. To tackle this, we first implement a lightweight router. This router is a small, fast LLM such as GPT-5 nano or Gemini 3.0-flash. The router takes in the prompt and a high-level JSON summary of the PPT and then decides: (1) whether to route the editing to direct XML editing or programmatic editing; (2) which slides are the editing target required for the VLMs to operate on. In our prompts, we use in-context examples to provide guidance and reference for the router. For example, editing content across many slides, \emph{e.g.}, more than 5, is better for programmatic editing, while visual contents, structures, and layouts are better suited for XML edits for better fidelity. Empirically, across all PPTArena edits the router dispatches $66\%$ of queries to the programmatic (\texttt{python-pptx}) path and $34\%$ to the direct XML path\label{app:router-dist}, reflecting that bulk, content-centric operations dominate real-world decks while a substantial minority of edits demand the fine-grained structural control of the XML path.


\mypar{Direct XML Editing.} 
If the router chooses to tackle the task using XML patching, the LLM digests the XML structure, JSON summary, and prompts, then returns which XML files and patches to edit. Such information is further provided to a stronger reasoning model to refine and produce the final changes.

\mypar{Programmatic Editing.} 
When the router flags large-scale content requests, the system follows a programmatic path with two sequential LLM invocations. The first produces a structured content plan or rewritten instruction with more specified objectives. This is further combined with the full JSON summary of the slides and relevant screenshots as input for the second LLM inference. The second step generates executable editing code (\emph{e.g.}, via pptx python libraries) that applies the updates to the targeted slides. Separating content synthesis from code generation improves reliability for operations like slide creation, translation, and summarization.

\mypar{Verifier and Error Controls.}
We implement heuristic checks to verify the output XML produces a valid PPT and that code generation runs without errors. The functionality of this step is similar to that of a PPT compiler, returning the errors to the VLMs for well-formed XML for valid slides. 

\begin{figure*}[!t]
    \centering
    \small
    \begin{tabular}{@{}ccc@{}}
        \begin{minipage}[t]{0.31\linewidth}
        \textbf{Conduct a profile crop of each face.}
            \centering
            \includegraphics[page=1,width=\linewidth, trim={2cm 0cm 2cm 4cm}, clip]{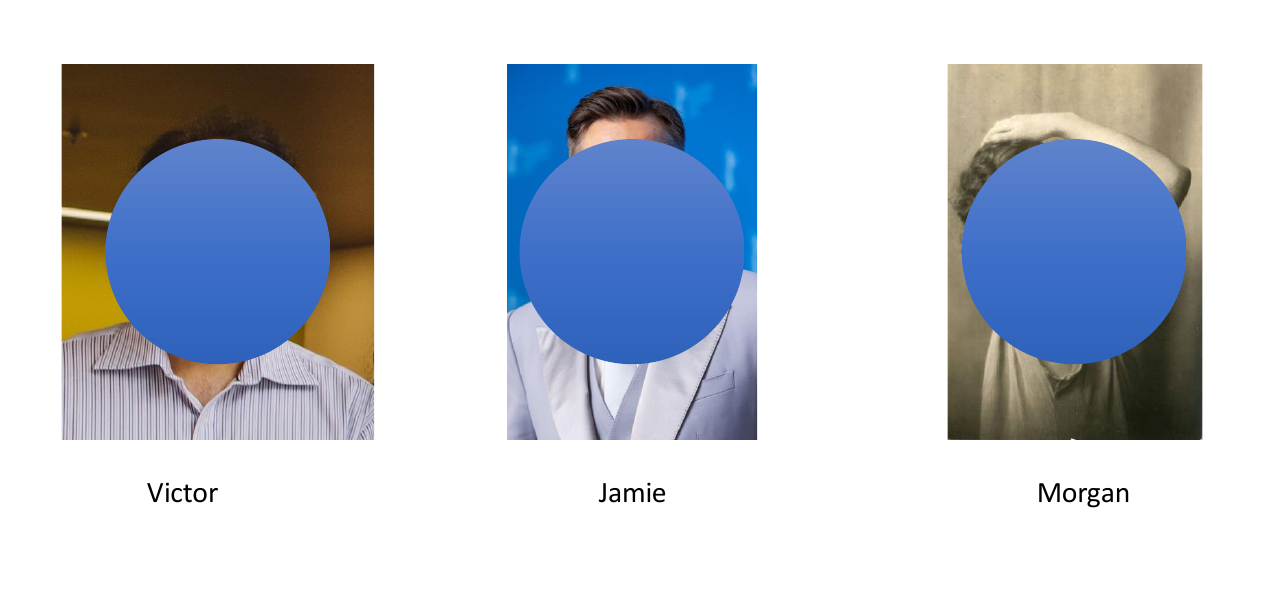}\\[-0.2em]
            \textit{\scriptsize ChatGPT Failure}\\
            
        \end{minipage} &
        \begin{minipage}[t]{0.31\linewidth}
        \textbf{Create a Rock Cycle diagram}
            \centering
            \includegraphics[page=3,width=0.75\linewidth]{figures/ChatGPTBad2.pdf}\\[-0.2em]
            \textit{\scriptsize ChatGPT Failure}\\
            
        \end{minipage} &
        \begin{minipage}[t]{0.31\linewidth}
        \textbf{Match image size \& correct caption}
            \centering
            \includegraphics[page=1,width=\linewidth, trim={2cm 2.4cm 2cm 5cm}, clip]{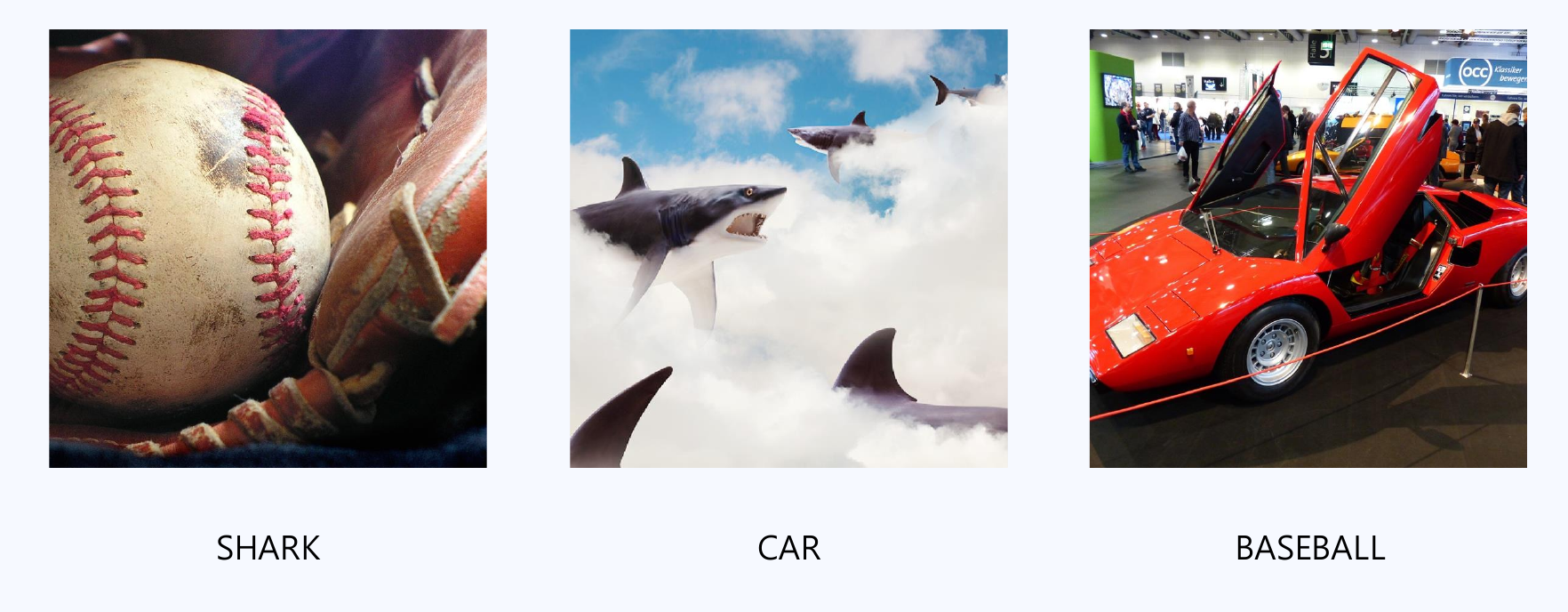}\\[-0.2em]
            \textit{\scriptsize PPTPilot \& ChatGPT Agent Failure}\\[0.5em]
        \end{minipage}
        \\[0.9em]
        \begin{minipage}[t]{0.31\linewidth}
        \textbf{Second Quarter $\rightarrow$ Q2}\\[0.2em]
            \centering
            \includegraphics[page=5,width=\linewidth]{figures/ChatgptAgentFailure.pdf}\\[-0.2em]
            \textit{\scriptsize ChatGPT Agent Failure}\\[0.5em]
        \end{minipage} &
        \begin{minipage}[t]{0.31\linewidth}
        \textbf{SmartArt timeline from Bullet Points}\\[0.2em]
            \centering
            \includegraphics[page=2,width=\linewidth, trim={0cm 4cm 8cm 0}, clip]{figures/dynamicPPTFailure.pdf}\\[-0.2em]
            \textit{\scriptsize PPTPilot Failure }\\[-0.2em]
        \end{minipage} &
        \begin{minipage}[t]{0.31\linewidth}
        \textbf{Kazakh $\rightarrow$ English; keep French}\\[0.2em]
            \centering
            \includegraphics[page=5,width=\linewidth, trim={3cm 3cm 10cm 5cm}, clip]{figures/chatgptAgentPredictionbad.pdf}\\[-0.2em]
            \textit{\scriptsize ChatGPT Agent Failure}\\[-0.2em]
        \end{minipage}
    \end{tabular}
    \caption{Malformed prediction artifacts from ChatGPT Extended Thinking mode, ChatGPT Agent, \& PPTPilot.}
    \label{fig:bad_predictions_grid}
\end{figure*}

\subsection{Self-correction and Reflection.} 
To enable more reliable PPT editing, we employ a self-correction reflective loop in PPTPilot. After each round of editing, PPTPilot formats the updated screenshot(s) of the changed slide(s) and the prompts back into the same editing skill for an additional round of editing. As shown in Table~\ref{tab:pptpilot_ablation_main}, such a technique steadily improves the quality of editing. In future work, we will explore incorporating an explicit VLM judge in the loop, without relying on the self-reflection capability of direct XML editing or programmatic editing branches.

\section{Performance Analysis and Failure Modes}
\label{app:quantitative}

We provide granular performance metrics across our five taxonomy categories; visualized benchmark tables are shown in \Cref{tab:gemini2.5proasEvals}.
Our results show that \textbf{Structure}, \textbf{Layout}, and \textbf{Interactivity} are the most challenging categories; often requiring strong XML path executions for success. Some of the most challenging edit types are ones require reasoning across multiple slides and multiples modalities incuding visual, text, and diagrams. We provide examples in \Cref{fig:bad_predictions_grid,fig:pptpilot_examples,fig:moreimages_cases}

\subsection{Comparatives}
In Figure~\ref{fig:bad_predictions_grid} we see how state-of-the-art models fail at tasks that contain both visual and textual dependencies. ChatGPT fails to crop the faces, instead placing a circle on top of the profiles. It also is unable to construct a well-formed simple rock cycle diagram. All of the models, including ChatGPT \& PPTPilot fail at reshaping and ordering images correctly. Once reshaped, agents lose track of the image content, resulting in mismatches with the captions. This case is particularly tricky as it requires alignment, image classification, and layout reasoning. In other cases, ChatGPT Agent fails but PPTPilot succeeds, for example when asked to translate Kazakh but keep French from a second-language acquisition slidedeck, ChatGPT kept the Kazakh and translated French into English—the exact opposite of the requested task. PPTPilot handles this correctly as shown in Figure~\ref{fig:pptpilot_examples}. However, PPTPilot failed when asked to generate a dynamic SmartArt timeline from a set of bullet points. While some visuals were included, they don't correspond to the text nor is the timeline dynamic and interactive as SmartArt is in PPTs. ChatGPT Agent also failed to correctly layout "Second Quarter" to "Q2" and instead overlaps it with other objects. We find that these agents struggle with tasks that are fundamental to powerpoint editing and require both visual, spatial, and semantic analysis and configuration. Although PPTPilot is capable of performing many tasks correctly as shown in Figure~\ref{fig:pptpilot_examples}, there is still a lot of progress to be made.

\begin{figure*}[!t]
    \centering
    \small
    
    \begin{minipage}[t]{0.32\textwidth}
        \centering
        \textbf{Order images to the right slide}\\[0.35em]
        \includegraphics[width=0.95\linewidth]{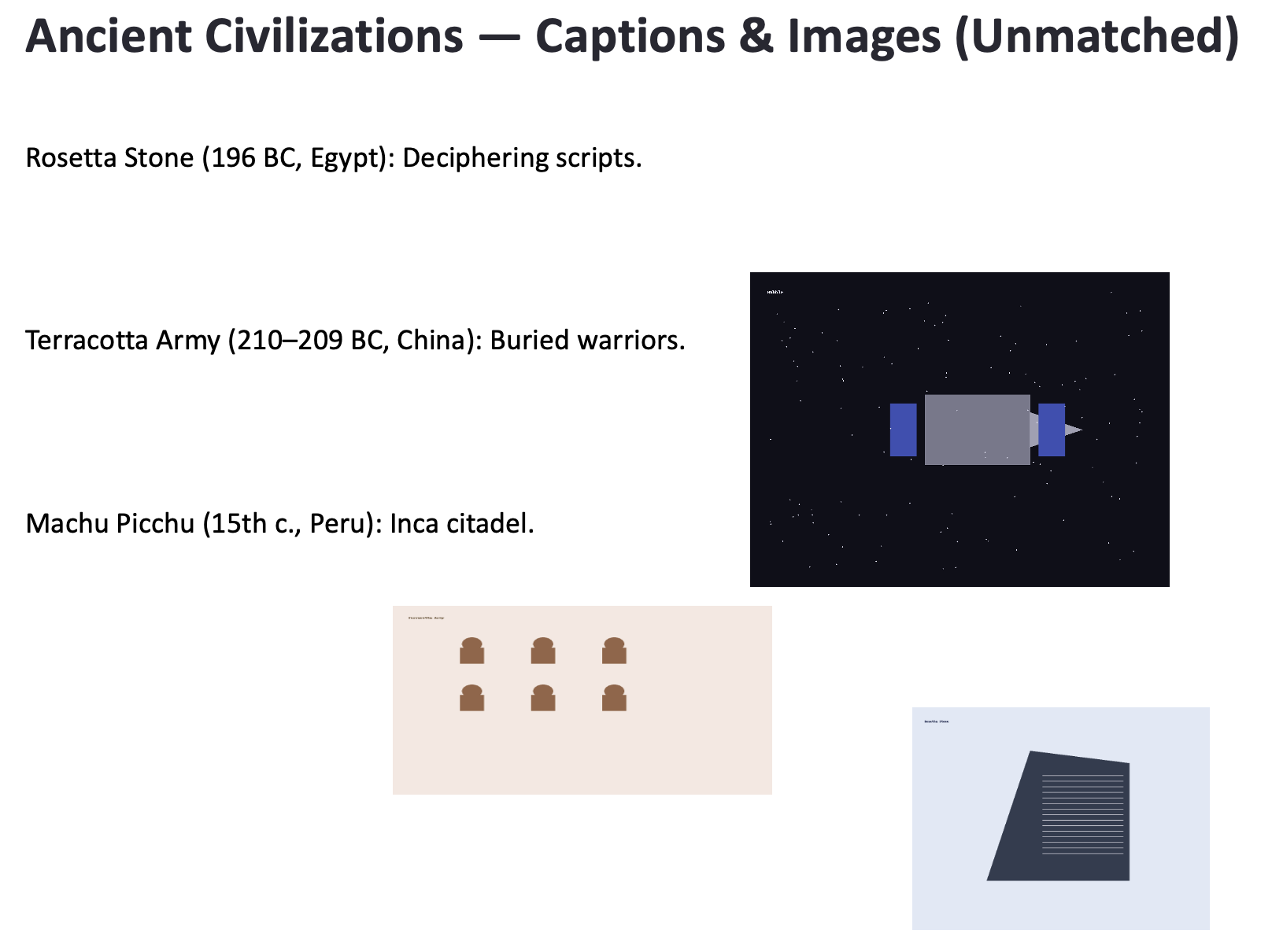}\\[-0.15em]
        \textit{\scriptsize Original}\\[0.35em]
        \includegraphics[width=0.95\linewidth]{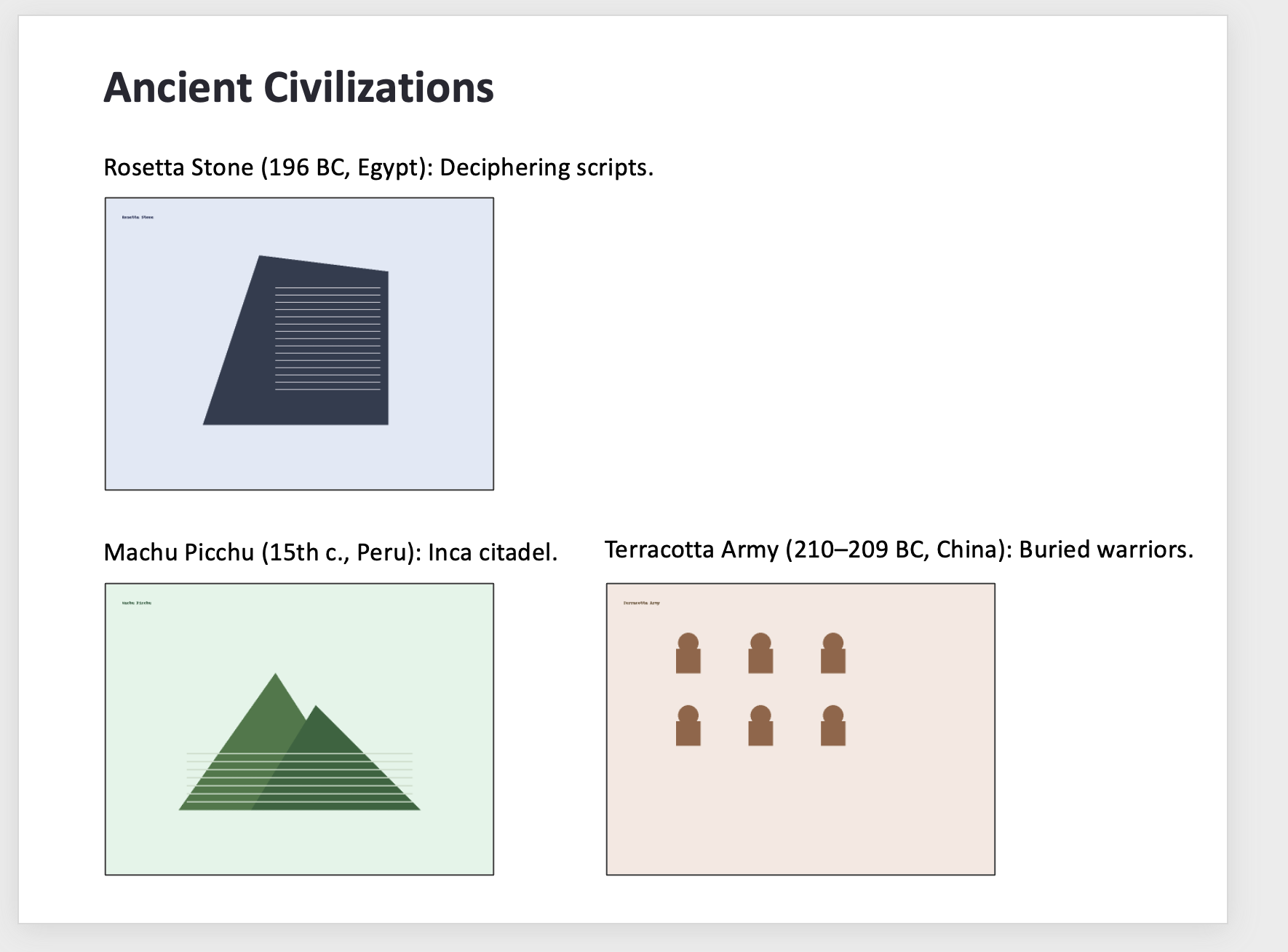}\\[-0.15em]
        \textit{\scriptsize PPTPilot Correction}
    \end{minipage}\hfill
    \begin{minipage}[t]{0.32\textwidth}
        \centering
        \textbf{Kazakh $\rightarrow$ English; keep French}\\[0.35em]
        \includegraphics[page=5, width=0.95\linewidth, trim=0.0cm 0.0cm 0.6cm 0.0cm, clip]{figures/translatekazak.pdf}\\[-0.15em]
        \textit{\scriptsize Original}\\[0.35em]
        \includegraphics[page=5, width=0.95\linewidth]{figures/PPTPIlotcorrect.pdf}\\[-0.15em]
        \textit{\scriptsize PPTPilot Correction}
    \end{minipage}\hfill
    \begin{minipage}[t]{0.32\textwidth}
        \centering
        \textbf{Convert Math from Latex to OOML}\\[0.35em]
        \includegraphics[page=3, width=0.95\linewidth]{figures/LaTeXtoOMML_TestA.pdf}\\[-0.15em]
        \textit{\scriptsize Original}\\[0.35em]
        \includegraphics[page=3, width=0.95\linewidth]{figures/LaTeXtoOMML_GroundTruthA.pdf}\\[-0.15em]
        \textit{\scriptsize PPTPilot Correction}
    \end{minipage}

    \caption{PPTPilot agent predictions on challenging benchmark cases. The top row shows original slides, while the bottom row displays PPTPilot's automated corrections. These cases demonstrate content reordering while preserving semantic meaning, semantic-reasoning for text replacements, and powerpoint integration with formatting dependencies. These examples highlight PPTPilot's ability to interpret complex instructions and execute precise modifications.}
    \label{fig:pptpilot_examples}
\end{figure*}

\section{Baseline Configurations and Budgets}
\label{app:baseline-configs}

For reproducibility, Table~\ref{tab:baseline_configs} reports the interface, backbone model and its maximum context window, the refinement/retry budget, and per-task latency for every system we evaluate. The total cost of running all baselines and both VLM judges across PPTArena is approximately \$700.

\mypar{ChatGPT (extended thinking).} We use the ChatGPT webchat interface in extended-thinking mode. Each task provides the source \texttt{.pptx} and the natural-language instruction; the model returns an edited \texttt{.pptx}, with up to three retries when the output is malformed or fails to open. We do not grant tool access or screenshots beyond the file itself, isolating the model's intrinsic editing ability.

\mypar{Other baselines.} ChatGPT Agent operates the desktop application through GUI/computer-use control and can enter long verification loops (we cap a single task at 30 minutes). Gemini CLI edits through a command-line agent, MiniMax Agent through its hosted web agent, and Kimi-K2.6 is run as an open-weight chat baseline. PPTAgent and Paper2Poster are one-shot generation pipelines. The fixed $3\times$ refinement cap for PPTPilot was selected from validation behavior: most corrections occur by the second pass, while later iterations plateau (Sec.~\ref{sec:ablation}).

\begin{table}[!htbp]
  \centering
  \footnotesize
  \setlength{\tabcolsep}{5pt}
  \renewcommand{\arraystretch}{1.2}
      \caption{Baseline configurations and budgets on PPTArena. \emph{Context} is the backbone's maximum input context window per vendor documentation. Latencies are end-to-end wall-clock per task; dashes denote values not separately metered or not publicly reported. The open-source PPTAgent and Paper2Poster default to a configurable GPT-4o backbone. Total evaluation cost (all systems $+$ both judges) ${\approx}\,\$700$.}
  \label{tab:baseline_configs}
  \resizebox{\linewidth}{!}{%
  \begin{tabular}{lllccl}
  \toprule
  \textbf{System} & \textbf{Interface} & \textbf{Backbone} & \textbf{Context} & \textbf{Refine/Retry} & \textbf{Latency} \\
  \midrule
  PPTPilot (ours)          & Hybrid (code+XML)  & GPT-5.2 (router: GPT-5 nano) & 272K & $\le 3$ loops   & 1.5--3 min \\
  ChatGPT (ext.\ thinking) & Webchat            & GPT-5.2                      & 272K & $\le 3$ retries & --         \\
  ChatGPT Agent            & GUI / computer-use & GPT-5.2 agent                & 272K & --              & 4--30 min (30-min cap) \\
  Gemini CLI               & CLI agent          & Gemini 3.1 Pro               & 1M   & --              & --         \\
  MiniMax Agent            & Hosted web agent   & MiniMax M2                   & 205K & --              & 3 min      \\
  Kimi-K2.6                & Open-weight chat   & Kimi-K2.6                    & 256K & --              & --         \\
  PPTAgent                 & One-shot gen.      & GPT-4o (configurable)        & 128K & --              & --         \\
  Paper2Poster             & One-shot gen.      & GPT-4o / Qwen-2.5            & 128K & --              & --         \\
  \bottomrule
  \end{tabular}}
\end{table}

\section{Reproducibility and Release Details}
\label{app:release}

We release our code, PPTArena benchmark, PPTPilot Agent (public webapp), detailed prompts and benchmark files to ensure reproducibility. Concretely, our release includes the per-judge prompt templates, rendered slide screenshots, the structured (XML/JSON) diffs used by the instruction-following judge, and the raw judge outputs for every system, so that all scores in this paper can be independently recomputed.

\section{Limitations and Future Work}
\label{app:limitations-future}

We have introduced PPTArena and PPTPilot to establish presentation editing as a rigorous, measurable domain for multi-modal agents. Moving beyond pixel-level generation to structure-aware editing, we demonstrate that reliable automation requires distinct planning, routing, and verification steps. While PPTPilot sets a new state-of-the-art, the complexity of real-world presentation design offers a vast landscape for future research. We outline several key directions where we envision expanding the scope of agentic presentation editing.

\mypar{Collaborative editing.} Our current evaluation relies on explicitly stated instructions. In practice, user intent is often under-specified (\eg, ``make this slide less cluttered'' or ``highlight the content here''). Future work should explore conversational refinement, where agents are evaluated not just on the final edit, but on their ability to ask clarifying questions, propose options, and engage in multi-turn dialogue to resolve ambiguity before executing changes.

\mypar{Cross-application workflows.} PPTArena evaluates editing within the closed environment of a slide deck. However, professional workflows frequently involve migrating data between applications, such as embedding live Excel charts or synthesizing Word documents into slides. An exciting frontier is extending the benchmark to support cross-application manifests, testing an agent’s ability to maintain semantic consistency as it shuttles content between diverse file formats and software ecosystems.

\mypar{Hyper-Specialized Domain Coverage.} While PPTArena spans a diverse taxonomy from business to biology, certain hyper-specialized domains impose unique constraints not yet fully captured. We plan to extend our dataset to include technical and scientific edge cases, such as editing complex LaTeX equations in engineering decks or managing strict regulatory compliance disclosures in financial presentations. This will test the limits of an agent's external knowledge retrieval and its ability to adhere to rigid industry standards.

\clearpage

\begin{figure}[!ht]
    \centering
    \colorbox{yellow!12}{%
    \begin{minipage}{0.96\linewidth}

        \begin{minipage}{\linewidth}
            \begin{minipage}{0.48\linewidth}
                \centering\textbf{\large Original}
            \end{minipage}%
            \hfill
            \begin{minipage}{0.48\linewidth}
                \centering\textbf{\large Ground Truth}
            \end{minipage}
        \end{minipage}
        
        \rule{\linewidth}{1pt}

        \textbf{\normalsize Multi-Edit Cascade}\\[0.2em]
        \begin{minipage}{0.48\linewidth}
            \centering
            \includegraphics[page=4,width=0.85\linewidth]{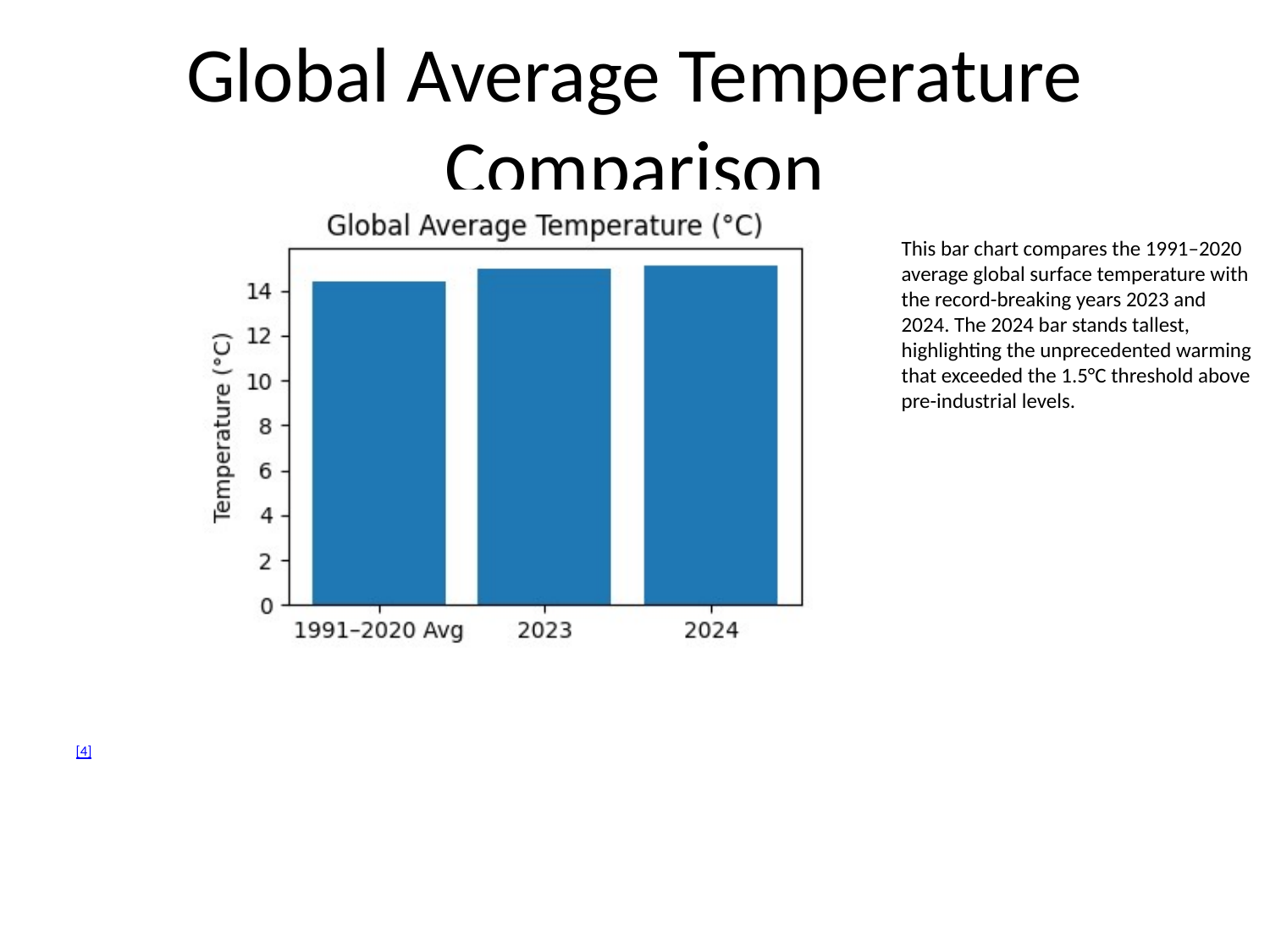}\\[0.2em]
            \includegraphics[page=6,width=0.85\linewidth]{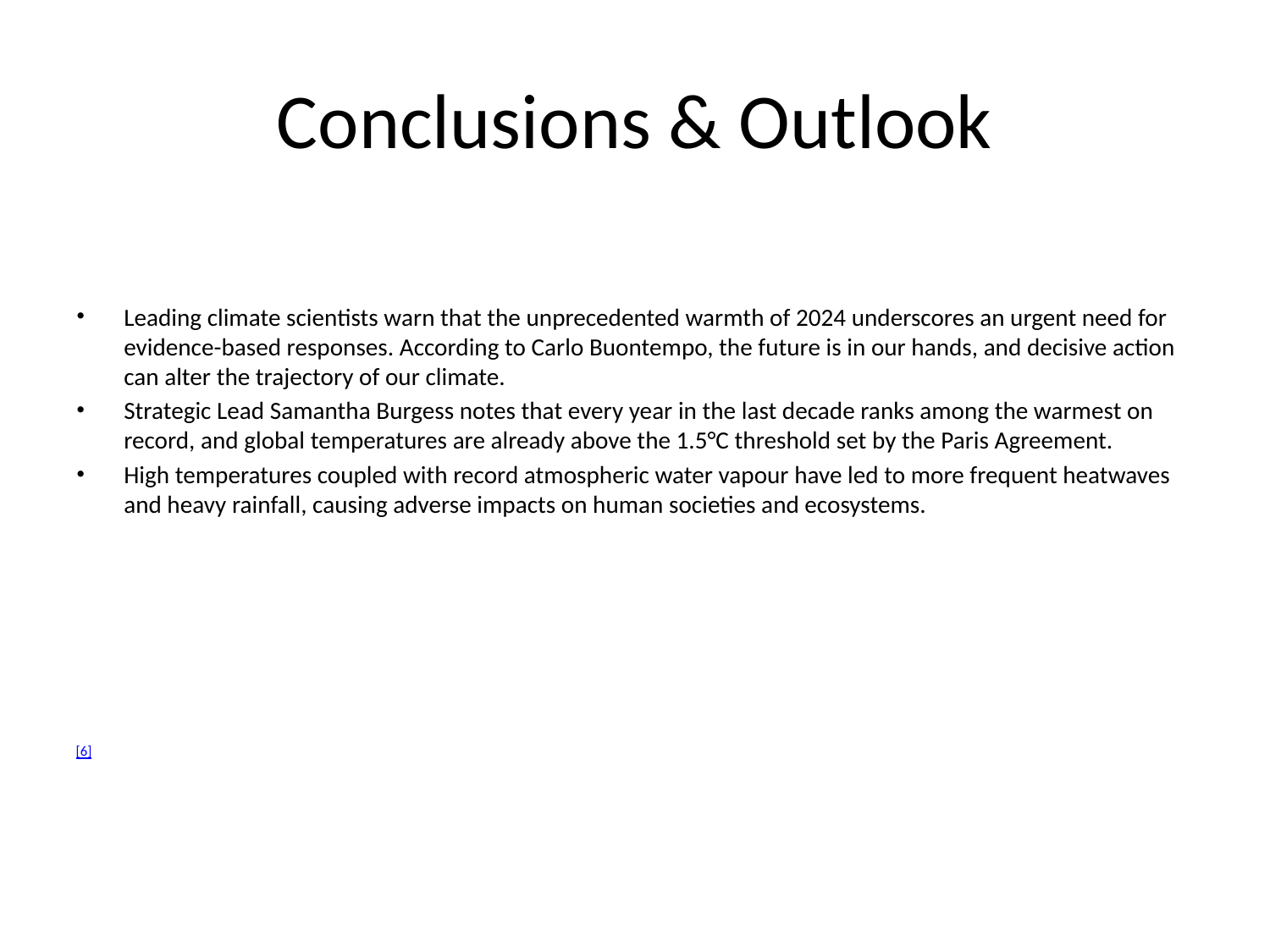}
        \end{minipage}%
        \hfill
        \begin{minipage}{0.48\linewidth}
            \centering
            \includegraphics[page=4,width=0.85\linewidth]{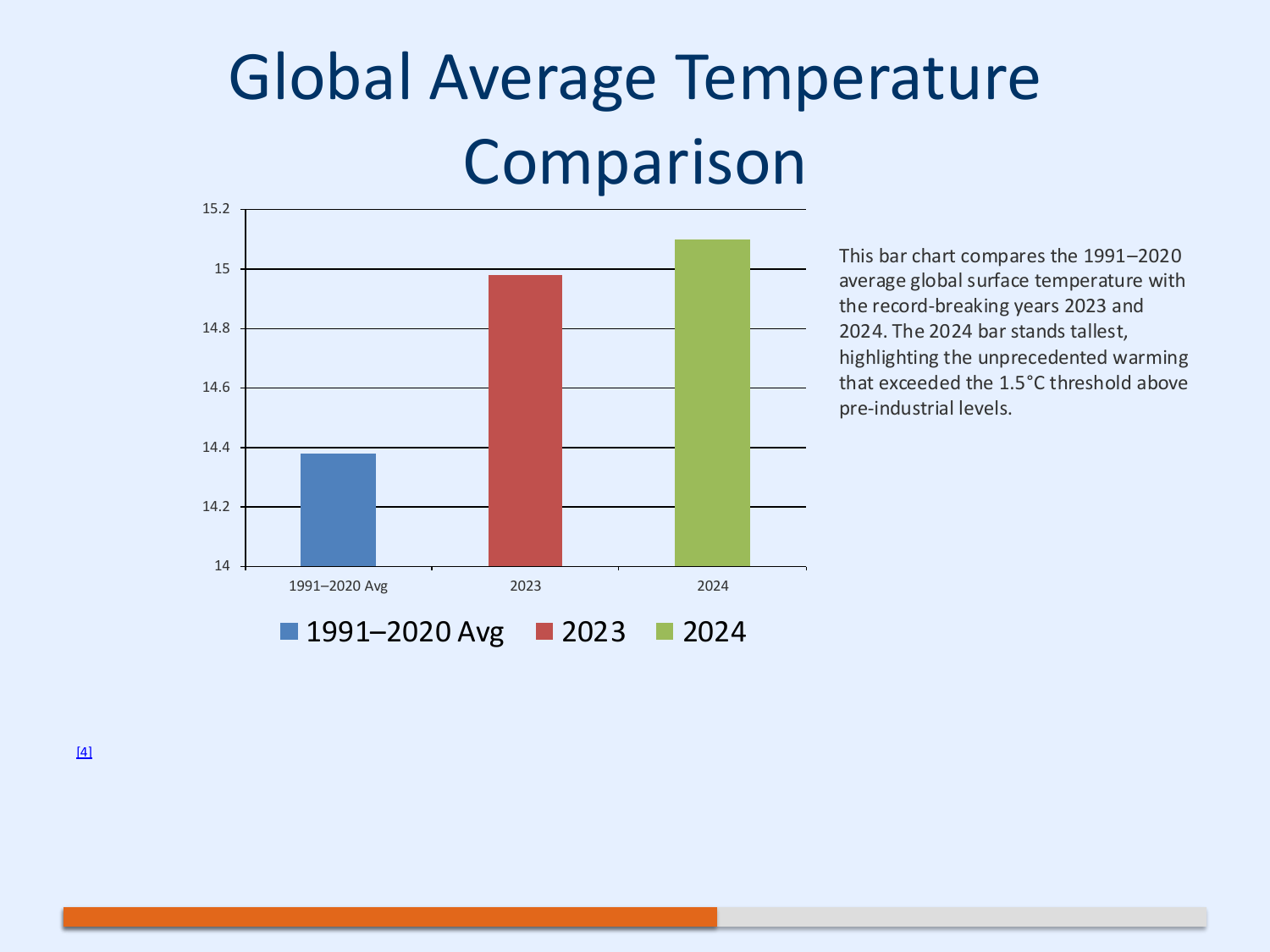}\\[0.2em]
            \includegraphics[page=7,width=0.85\linewidth]{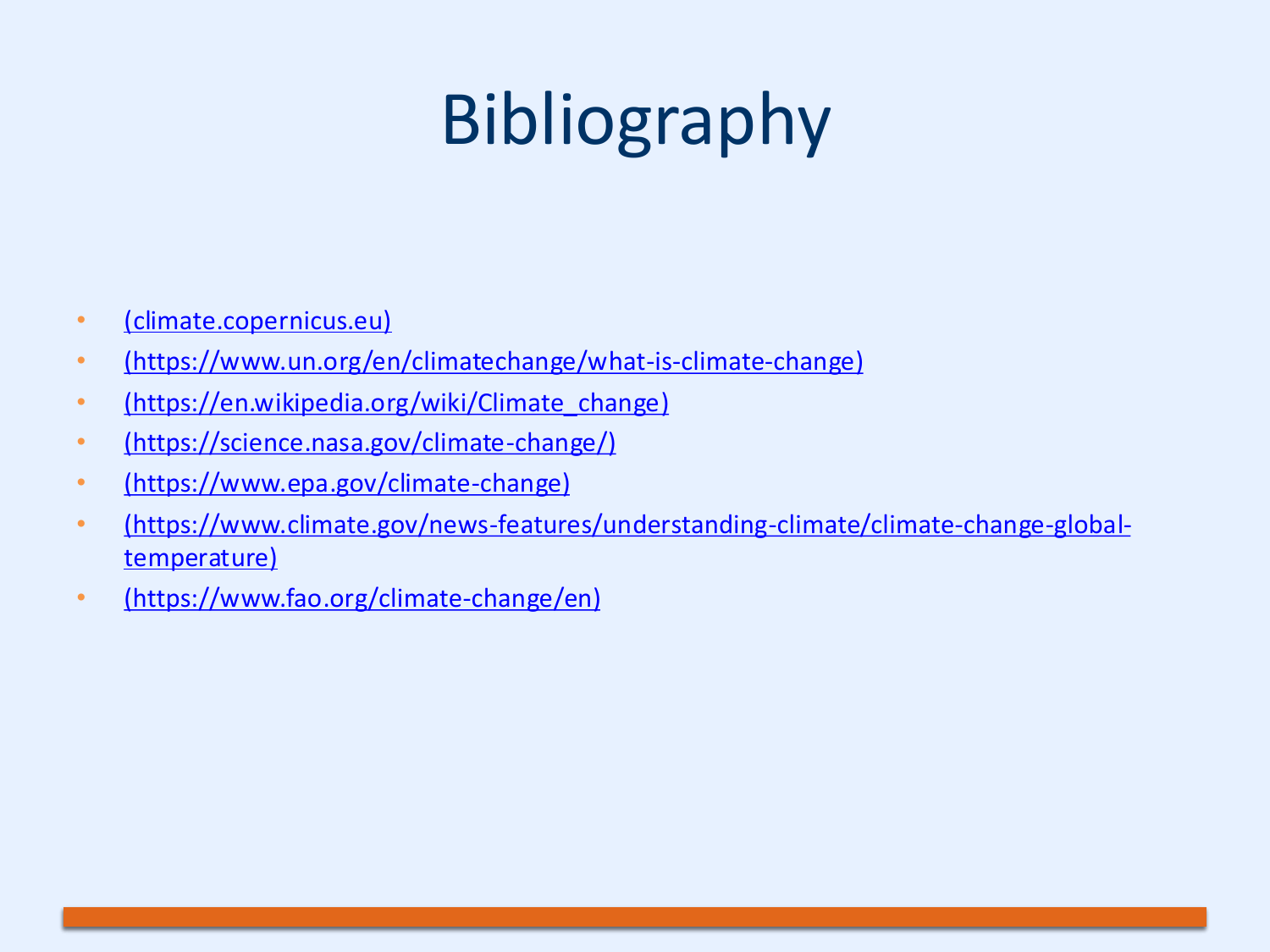}
        \end{minipage}

        \rule{\linewidth}{0.5pt}

        \textbf{\normalsize Consolidate Data 3 $\rightarrow$ 1}\\[0.2em]
       \begin{minipage}{0.48\linewidth}
          \centering
          \includegraphics[page=2,width=0.5\linewidth]{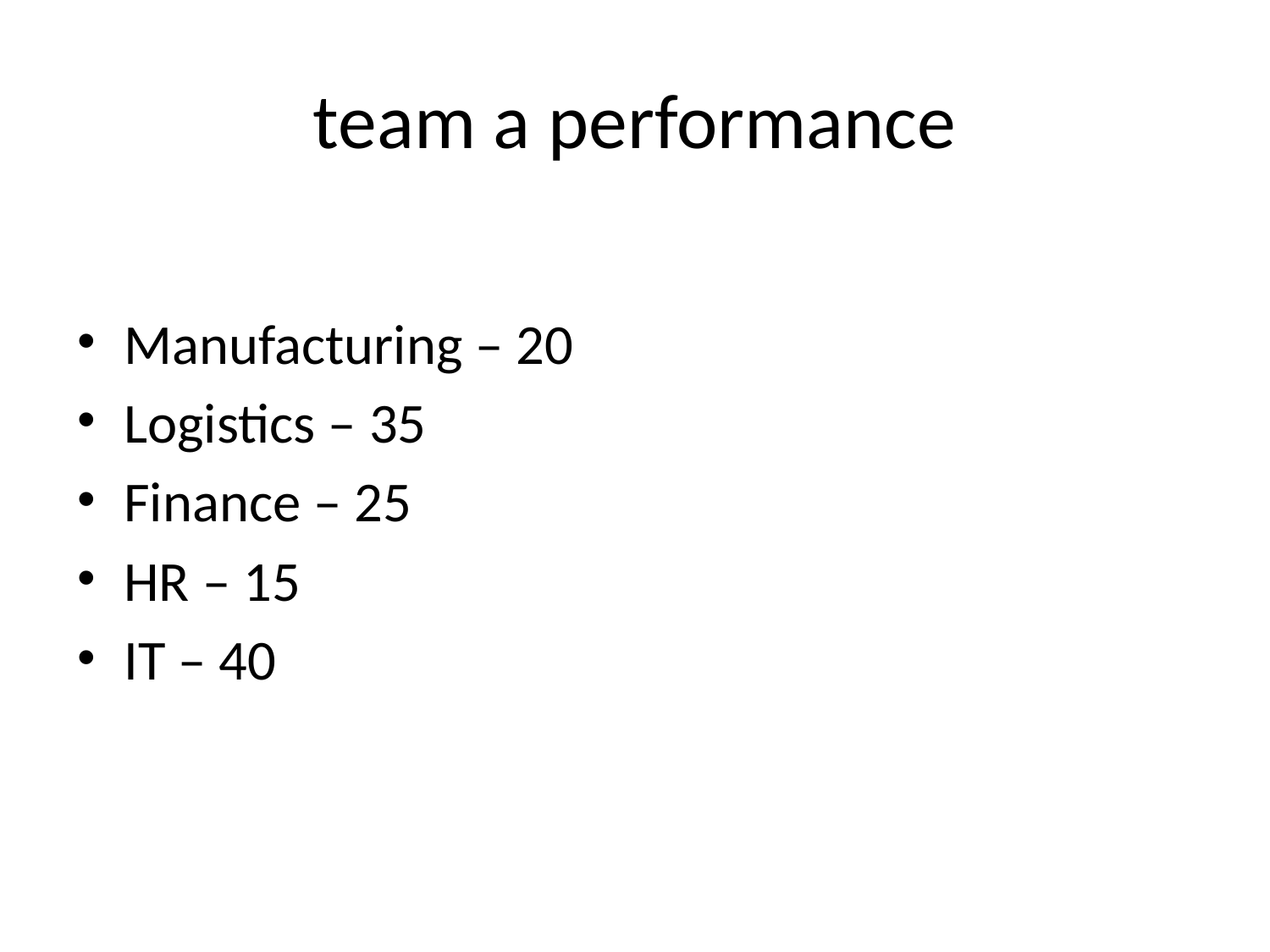}%
          \hspace{-0.15\linewidth}%
          \includegraphics[page=3,width=0.5\linewidth]{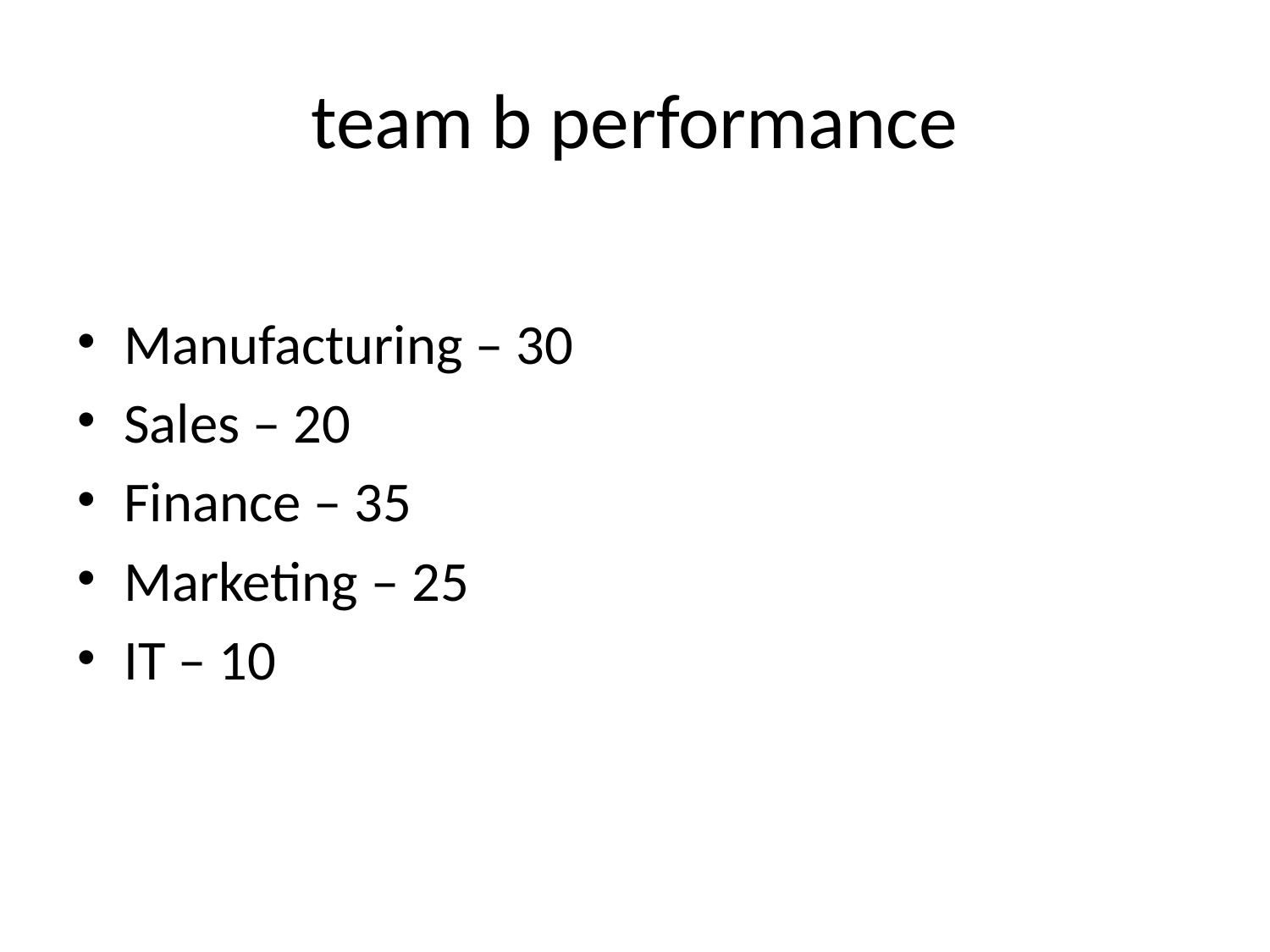}%
          \hspace{-0.15\linewidth}%
          \includegraphics[page=4,width=0.5\linewidth]{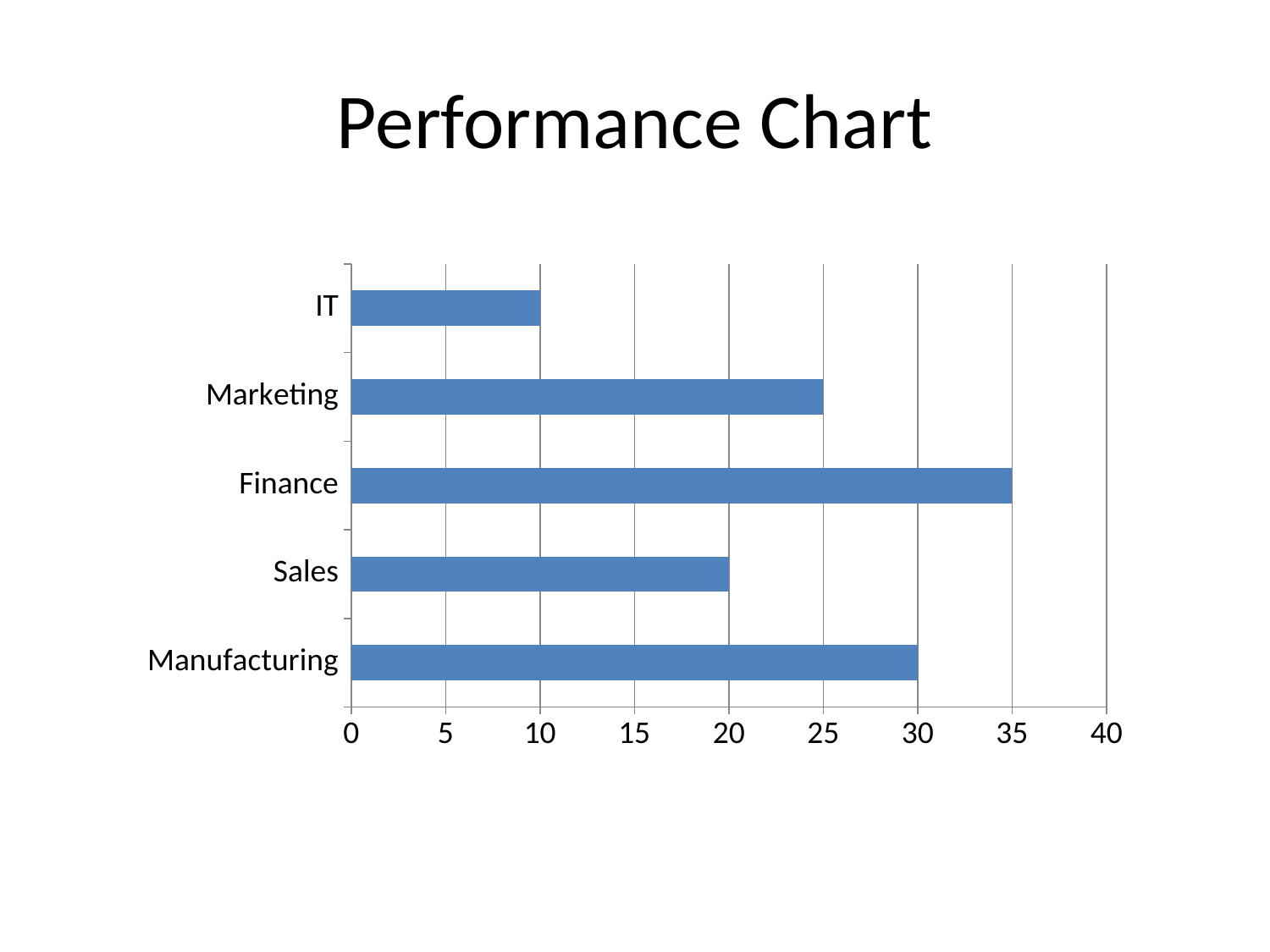}
        \end{minipage}%
        \hfill
        \begin{minipage}{0.48\linewidth}
          \centering
          \includegraphics[page=2,width=0.85\linewidth]{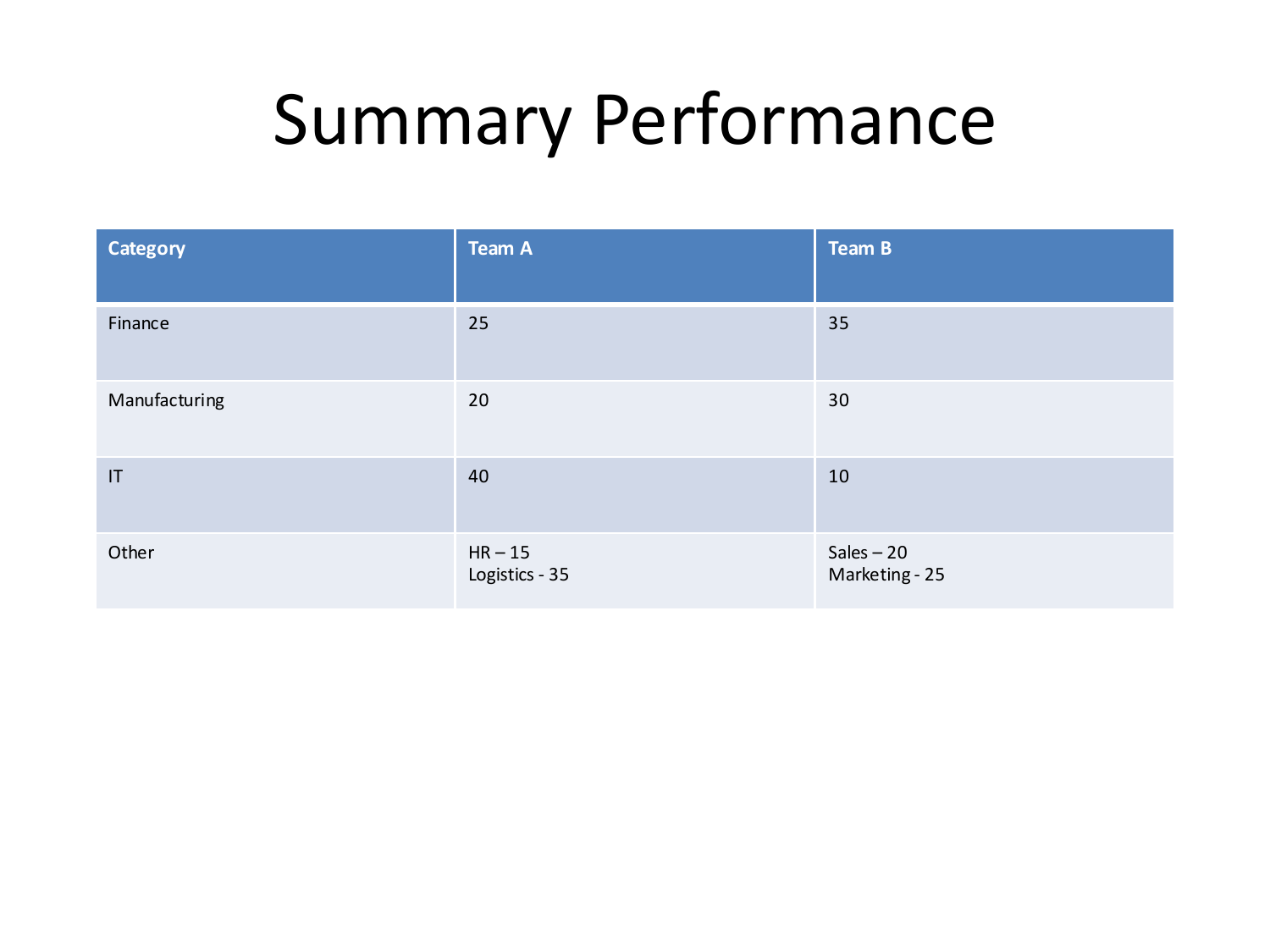}
        \end{minipage}

        \rule{\linewidth}{0.5pt}

        \textbf{\normalsize Cross-Slide Understanding}\\[0.2em]
        \begin{minipage}{0.48\linewidth}
            \centering
            \includegraphics[page=2,width=0.85\linewidth]{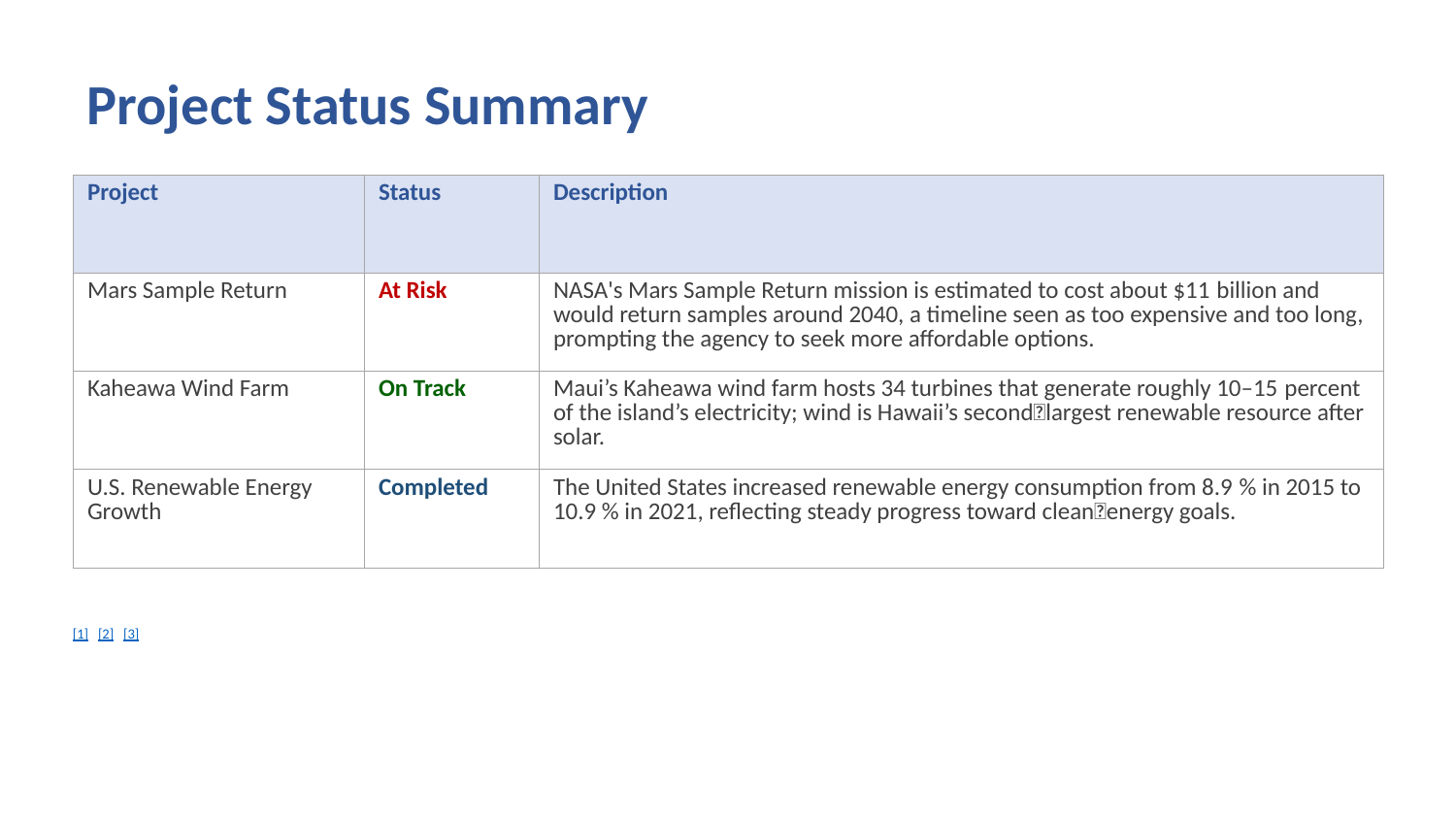}
        \end{minipage}%
        \hfill
        \begin{minipage}{0.48\linewidth}
            \centering
            \includegraphics[page=3,width=0.85\linewidth]{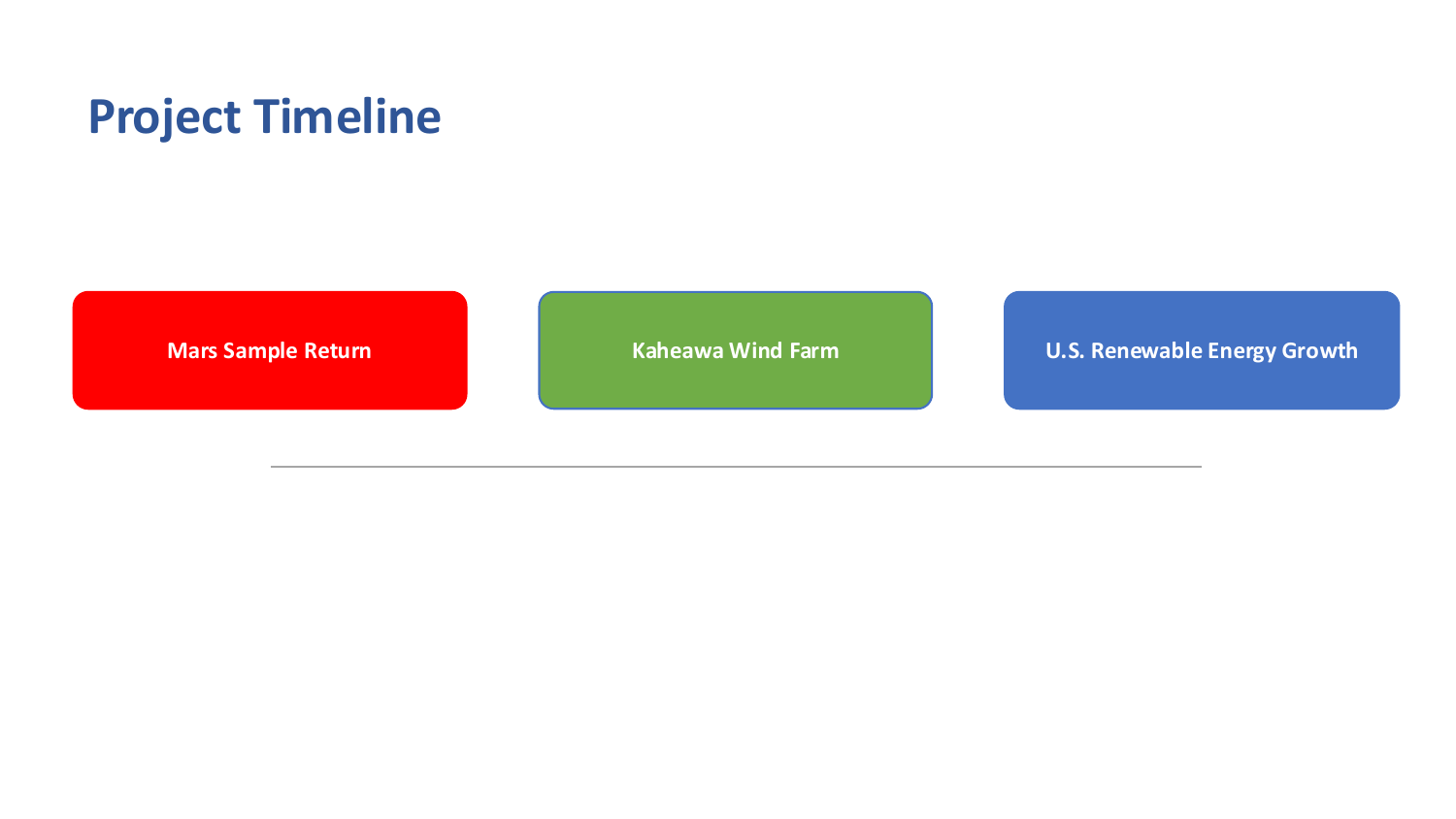}
        \end{minipage}

    \end{minipage}}
    \caption{Visualizing high-difficulty cases involving multi-edit cascades (Case 93), data consolidation (Case 74), and cross-slide understanding (Case 98).}
    \label{fig:appendix_cases_74_93_98}
\end{figure}

\begin{figure}[ht]
    \centering
    \colorbox{yellow!8}{%
        \begin{minipage}{0.98\linewidth}
            
            \textbf{\normalsize Prompt vs. Style Target Comparison (case 75)}\\[-.2em]
            \rule{0.95\linewidth}{0.5pt}\\[0.2em]
            \textcolor{blue!70!black}{\texttt{PROMPT}}:\\ \textit{"Please clean up this presentation. The slides are a mess. Align all the content to a consistent grid, ensure text boxes are vertically aligned, fix the chart legend so it doesn't overlap, and organize the pictures neatly with their captions below them."}\\[0.0em]
            
            \rule{0.95\linewidth}{0.5pt}\\[0.2em]
            \textcolor{blue!70!black}{\texttt{STYLE TARGET}} \section*{Slide 1:}
\begin{itemize}
    \item The three main text boxes (\texttt{'Overview'}, \texttt{'Details'}, \texttt{'Notes'}) must be arranged as columns, horizontally distributed with consistent spacing, and vertically top-aligned with each other.
    \item All elements on the slide (text boxes, table, picture) must be aligned \ldots
    \item The bulleted text within the \texttt{'Overview'}, \texttt{'Details'}, and \texttt{'Notes'} text boxes must be updated to describe the corrected state, as follows:
    \begin{itemize}
        \item \texttt{'Overview'} text: ``\textbullet{} This column snaps to grid.'', ``\textbullet{} Lines align to 8-pt baselines.'', ``\textbullet{} No overlaps or margin violations.''
        \item \texttt{'Details'} text: \ldots
    \end{itemize}
\end{itemize}

\section*{Slide 2:}
\begin{itemize}
    \item The chart's legend must be positioned so that it does not overlap with the chart's plot area. This can be achieved by moving the legend (e.g., to the bottom or right) or resizing the plot area.
    \item \ldots
    \item All remaining elements should be neatly arranged.
\end{itemize}

\section*{Slide 3:}
\begin{itemize}
    \item The slide title must be updated to ``Figures \& Captions (Aligned)''.
    \item The two pictures (\texttt{'Picture 9'} and \texttt{'Picture 11'}) must be \ldots
\end{itemize}

\section*{Global Constraints:}
\begin{itemize}
    \item The following properties should remain unchanged: Font styles (name, bold, italic), font sizes and colors, table data and styling, chart data and type, and the content of the pictures.
    \end{itemize}
            \end{minipage}%
        }
    
    \caption{Comparison showing the gap between a natural-language prompt and the detailed style target rubric (Case 75).}
    
    \label{fig:appendix_case75_example}
\end{figure}

\begin{figure*}[!t]
    \centering
    \definecolor{GeminiIF}{RGB}{217,65,2}   
    \definecolor{GPTIF}{RGB}{253,192,134}   
    \definecolor{GeminiVQ}{RGB}{31,90,180}  
    \definecolor{GPTVQ}{RGB}{166,206,227}   

    \ref{CommonLegend}
    

    \begin{minipage}[t]{0.45\textwidth}
        \centering
        \begin{tikzpicture}
            \begin{axis}[
                ybar,
                width=\linewidth,
                height=6.2cm,
                bar width=7pt,
                ymin=0, ymax=4.2,
                ymajorgrids,
                grid style={gray!18},
                enlarge x limits=0.12,
                symbolic x coords={PPTPilot,ChatGPT,GeminiCLI},
                xtick=data,
                xticklabels={PPTPilot,ChatGPT,Gemini\\CLI},
                xticklabel style={font=\small, text width=1.7cm, align=center},
                title={\textbf{PPTArena Benchmark Scores}},
                title style={yshift=-0.8ex},
            ]
                \addplot[fill=GeminiIF, draw=black!80] coordinates {
                    (PPTPilot,2.45) (ChatGPT,1.97) (GeminiCLI,1.92)
                };
                \addplot[fill=GPTIF, draw=black!80] coordinates {
                    (PPTPilot,2.36) (ChatGPT,2.07) (GeminiCLI,1.21)
                };
                \addplot[fill=GeminiVQ, draw=black!80] coordinates {
                    (PPTPilot,2.74) (ChatGPT,2.03) (GeminiCLI,2.15)
                };
                \addplot[fill=GPTVQ, draw=black!80] coordinates {
                    (PPTPilot,2.69) (ChatGPT,2.22) (GeminiCLI,1.98)
                };
            \end{axis}
        \end{tikzpicture}
    \end{minipage}%
    \hfill
    \begin{minipage}[t]{0.45\textwidth}
        \centering
        \begin{tikzpicture}
            \begin{axis}[
                ybar,
                width=\linewidth,
                height=6.2cm,
                bar width=7pt,
                ymin=0, ymax=4.2,
                ymajorgrids,
                grid style={gray!18},
                enlarge x limits=0.12,
                symbolic x coords={PPTPilot,ChatGPTAgent,MiniMaxAgent},
                xtick=data,
                xticklabels={PPTPilot,ChatGPT\\Agent,MiniMax\\Agent},
                xticklabel style={font=\small, text width=1.7cm, align=center},
                title={\textbf{Subset Evaluation Scores}},
                title style={yshift=-0.8ex},
                legend columns=4, 
                legend to name=CommonLegend, 
                legend style={
                    draw=none, 
                    fill=none, 
                    font=\footnotesize,
                    /tikz/every even column/.append style={column sep=0.5cm}
                }
            ]
                \addplot[fill=GeminiIF, draw=black!80] coordinates {(PPTPilot,2.04) (ChatGPTAgent,1.44) (MiniMaxAgent,0.92)};
                \addlegendentry{Gemini IF}

                \addplot[fill=GPTIF, draw=black!80] coordinates {(PPTPilot,1.71) (ChatGPTAgent,1.68) (MiniMaxAgent,1.04)};
                \addlegendentry{GPT-5.2 IF}

                \addplot[fill=GeminiVQ, draw=black!80] coordinates {(PPTPilot,2.60) (ChatGPTAgent,1.32) (MiniMaxAgent,0.80)};
                \addlegendentry{Gemini VQ}

                \addplot[fill=GPTVQ, draw=black!80] coordinates {(PPTPilot,1.54) (ChatGPTAgent,1.60) (MiniMaxAgent,0.84)};
                \addlegendentry{GPT-5.2 VQ}
            \end{axis}
        \end{tikzpicture}
    \end{minipage}

    \caption{\textbf{Average IF/VQ scores visualized by judge.} We compare Instruction Following (Orange) and Visual Quality (Blue) across agents. Darker bars represent the Gemini Judge; lighter bars represent the GPT-5.2 Judge. The legend at the top applies to both charts.}
    \label{fig:gemini_barplots}
\end{figure*}

\begin{figure*}[!t]
    \centering
    \small
    
    \begin{minipage}[t]{0.66\linewidth}
        \begin{minipage}[t]{0.48\linewidth}
            \centering
            \textbf{Create Layout and Add Captions}
            
            \includegraphics[width=\linewidth]{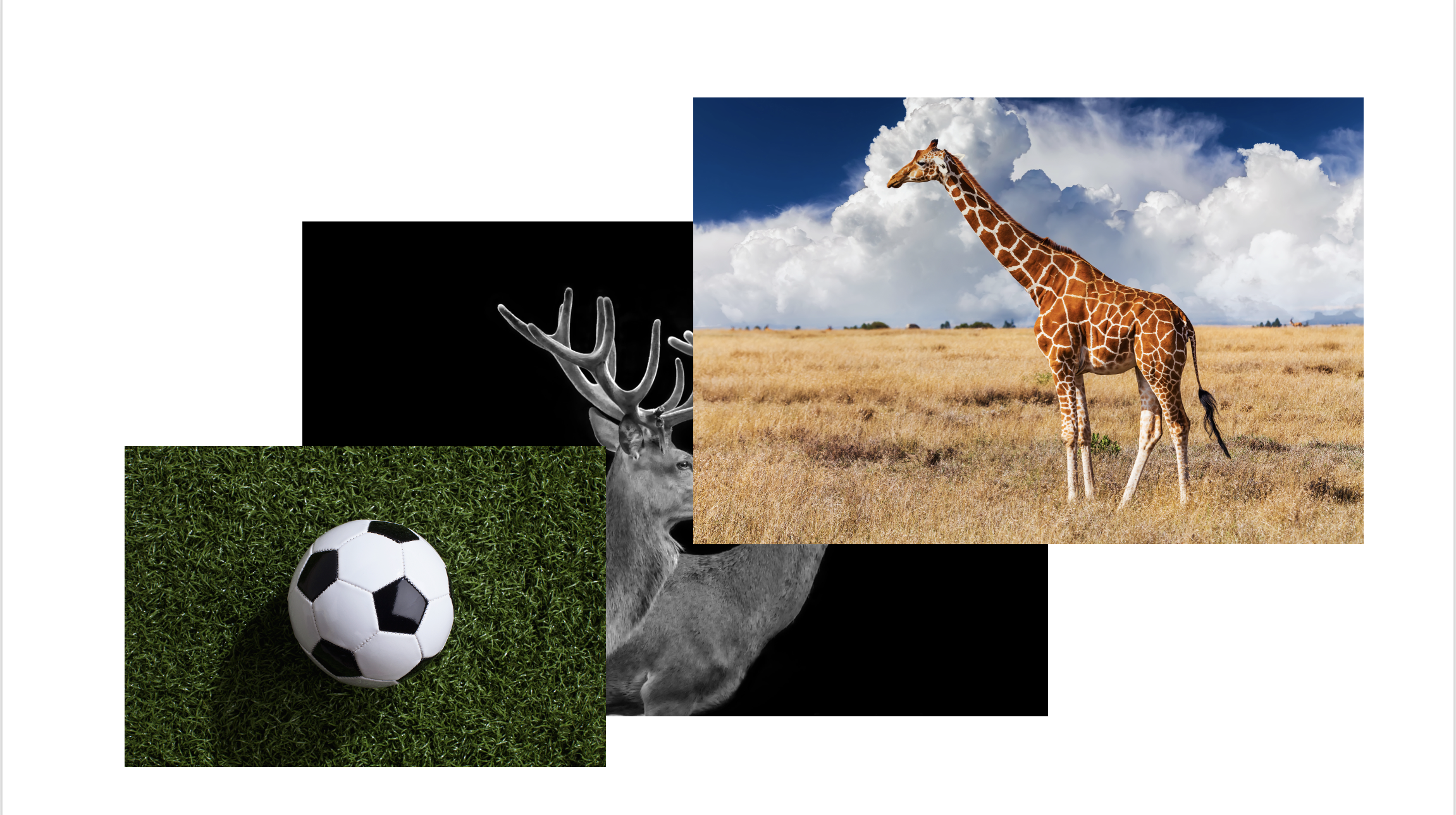}\\[-0.25em]
            \textit{\scriptsize Original}\\[0.4em]
            
            \includegraphics[width=\linewidth]{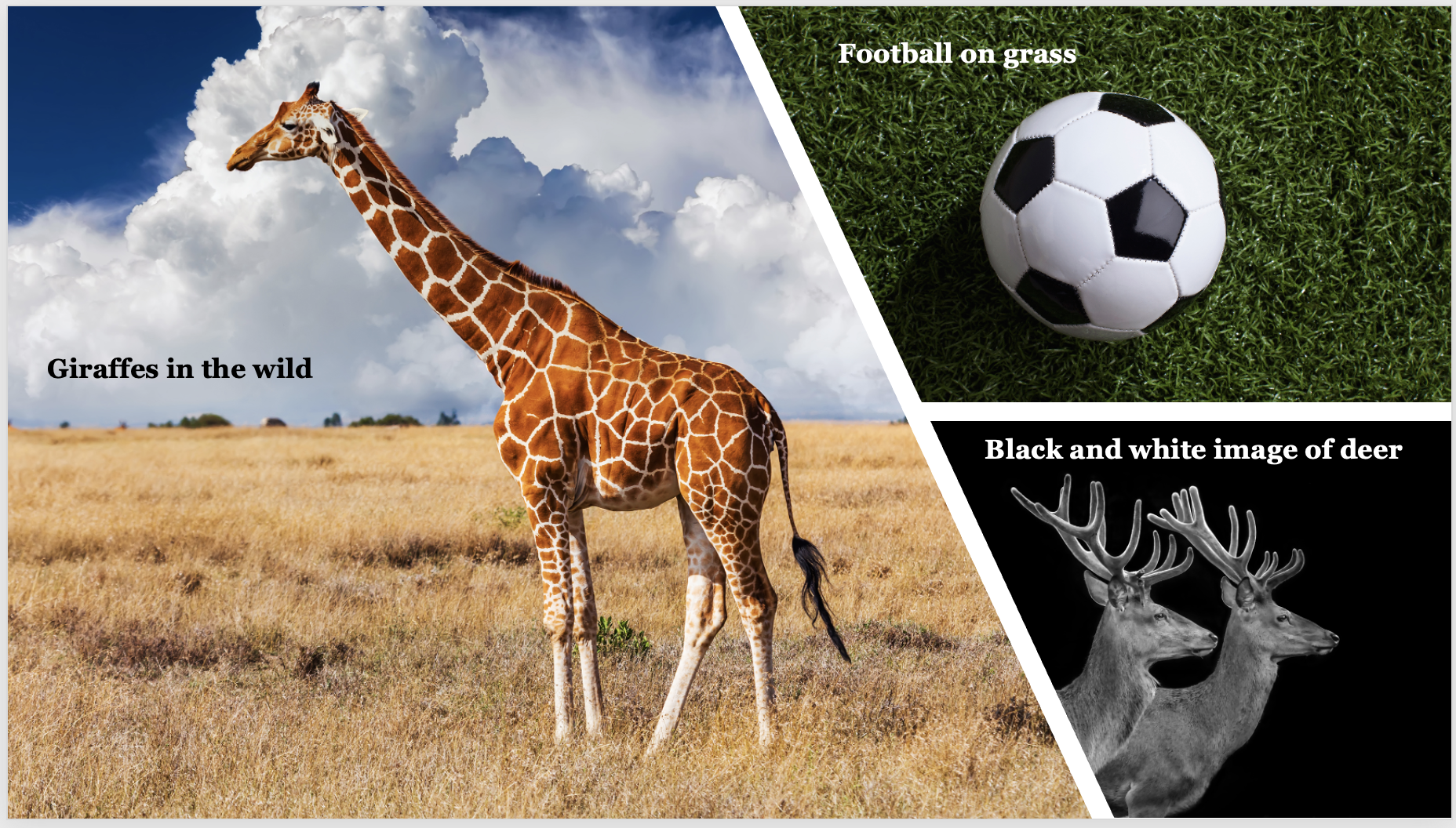}\\[-0.1em]
            \textit{\scriptsize Ground Truth}
        \end{minipage}
        \hfill
        \begin{minipage}[t]{0.48\linewidth}
            \centering
            \textbf{Correct Image Size \& Caption}
            
            \includegraphics[width=\linewidth]{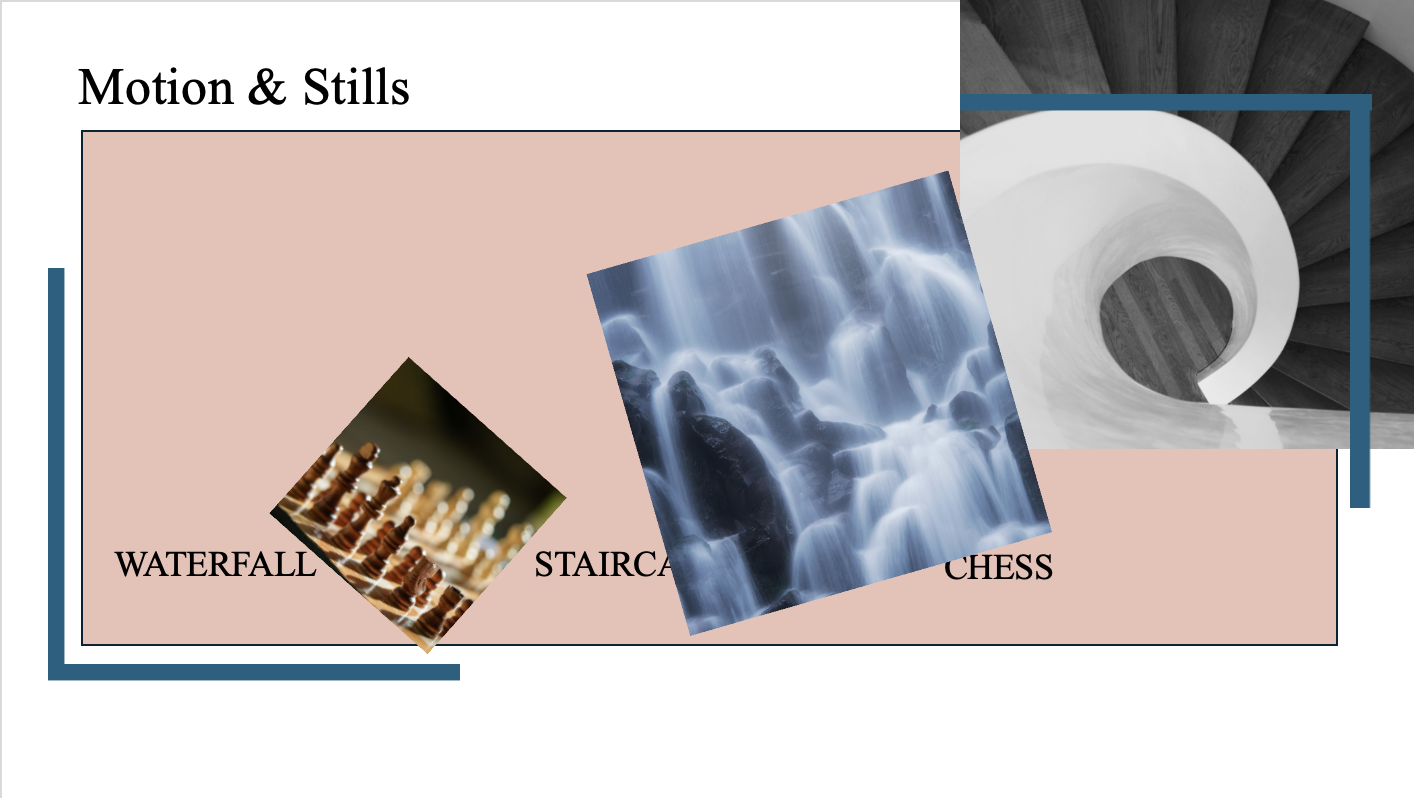}\\[-0.25em]
            \textit{\scriptsize Original}\\[0.4em]
            
            \includegraphics[width=\linewidth]{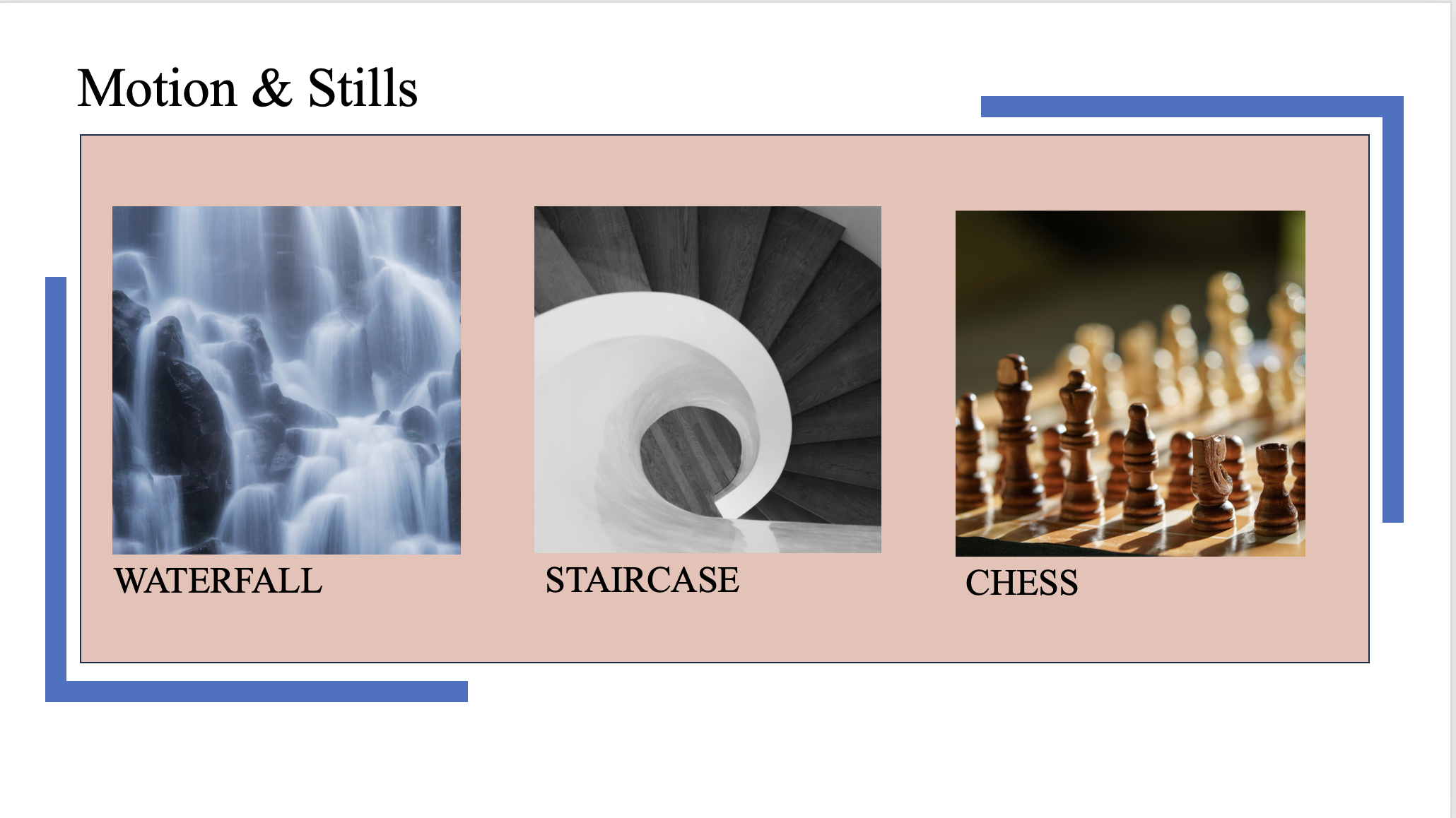}\\[-0.1em]
            \textit{\scriptsize Ground Truth}
        \end{minipage}
        
        
        \begin{minipage}[t]{0.48\linewidth}
            \centering
            \textbf{Fix Chart Theme \& Formatting}
            
            \includegraphics[width=\linewidth]{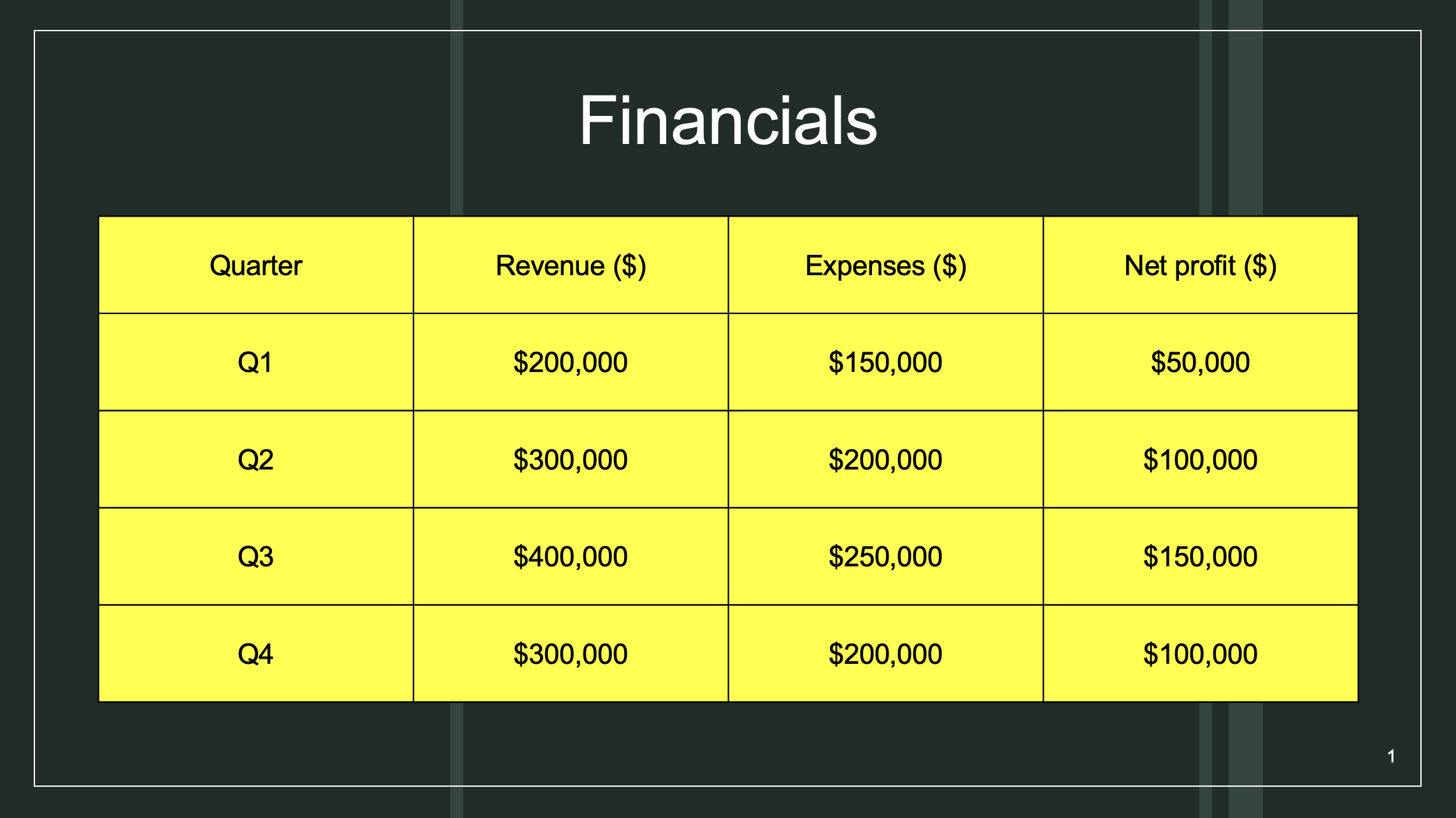}\\[-0.25em]
            \textit{\scriptsize Original}\\[0.4em]
            
            \includegraphics[width=\linewidth]{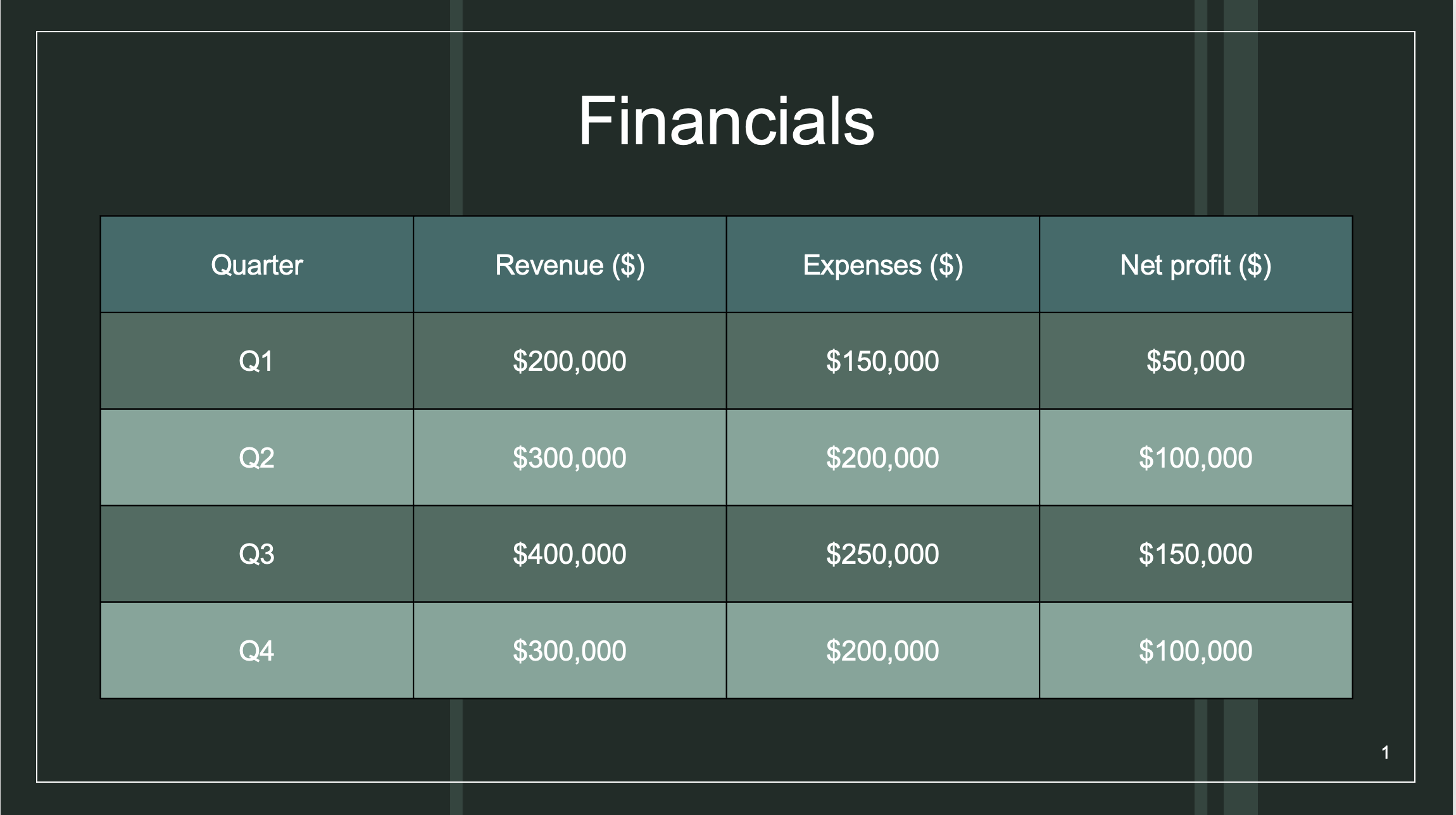}\\[-0.1em]
            \textit{\scriptsize Ground Truth}
        \end{minipage}
        \hfill
        \begin{minipage}[t]{0.48\linewidth}
            \centering
            \textbf{Organize Research Poster}
            
            \includegraphics[width=0.85\linewidth]{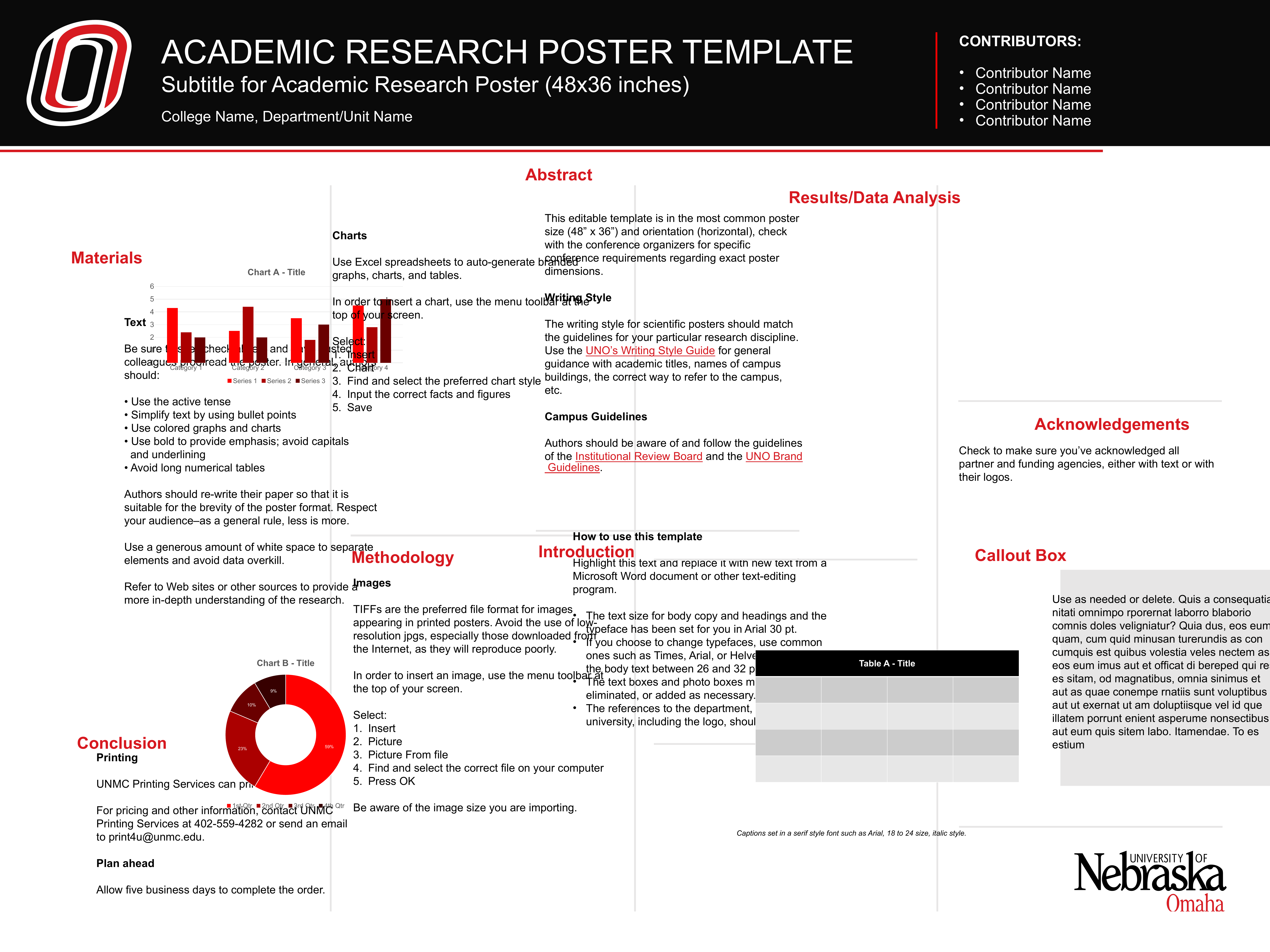}\\[-0.25em]
            \textit{\scriptsize Original}\\[0.4em]
            
            \includegraphics[width=0.85\linewidth]{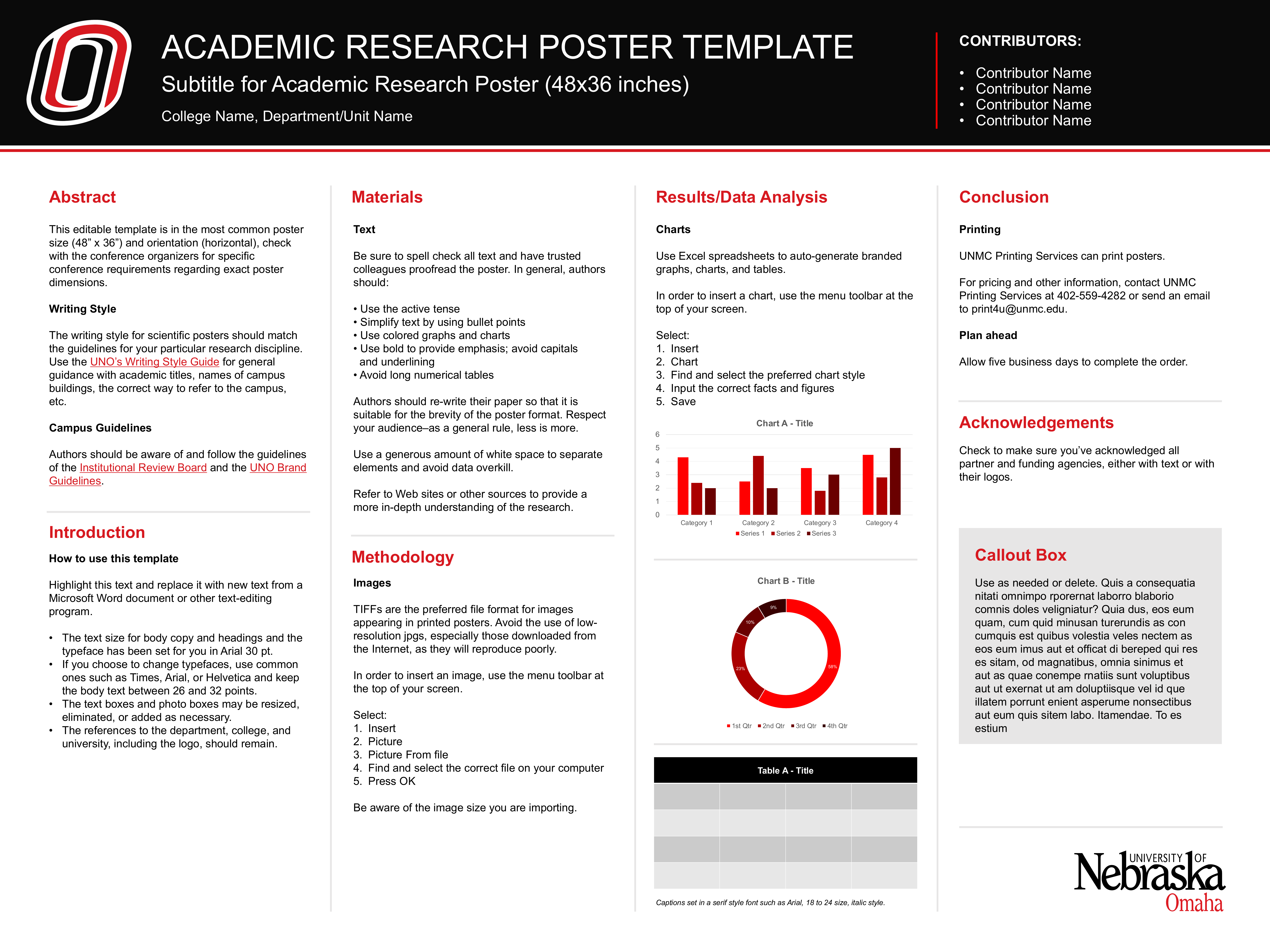}\\[-0.1em]
            \textit{\scriptsize Ground Truth}
        \end{minipage}
    \end{minipage}
    \hfill
    \begin{minipage}[t]{0.32\linewidth}
        \centering
        \textbf{Complex Collage Layout}

        \includegraphics[width=\linewidth]{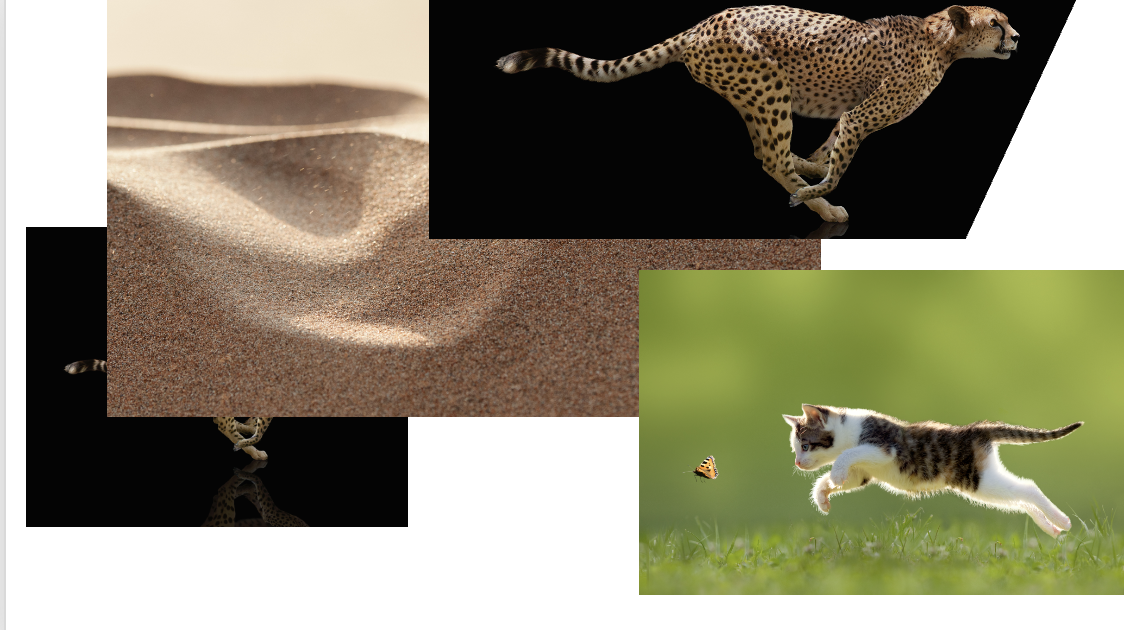}\\[0.2em]
        \includegraphics[width=\linewidth]{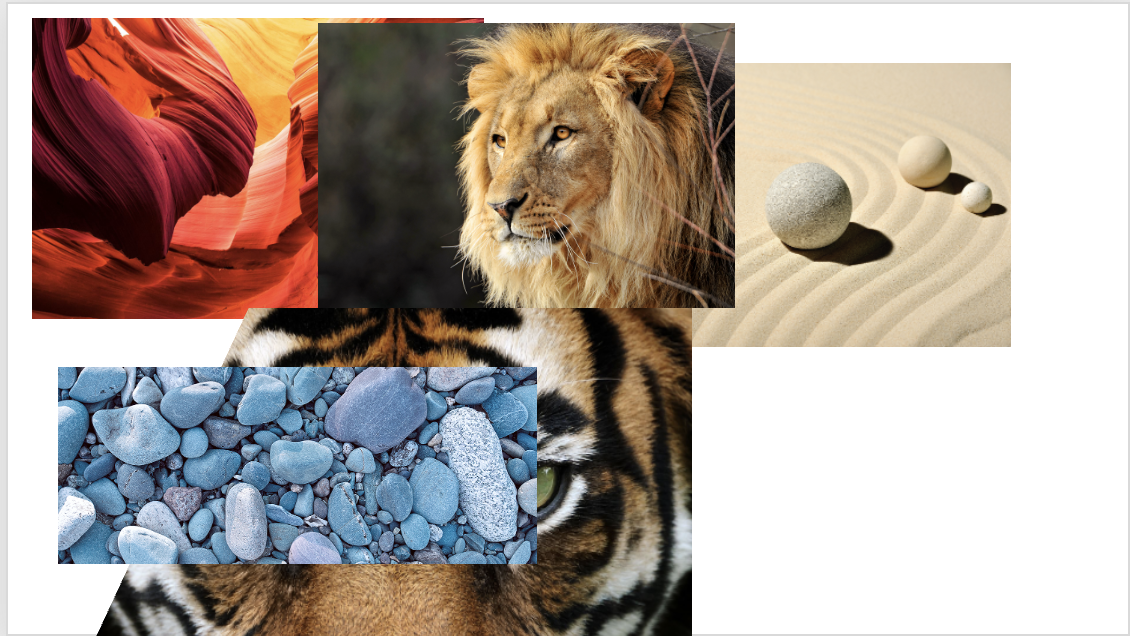}\\[-0.1em]
        \textit{\scriptsize Original Assets}
        
        
        \includegraphics[width=\linewidth]{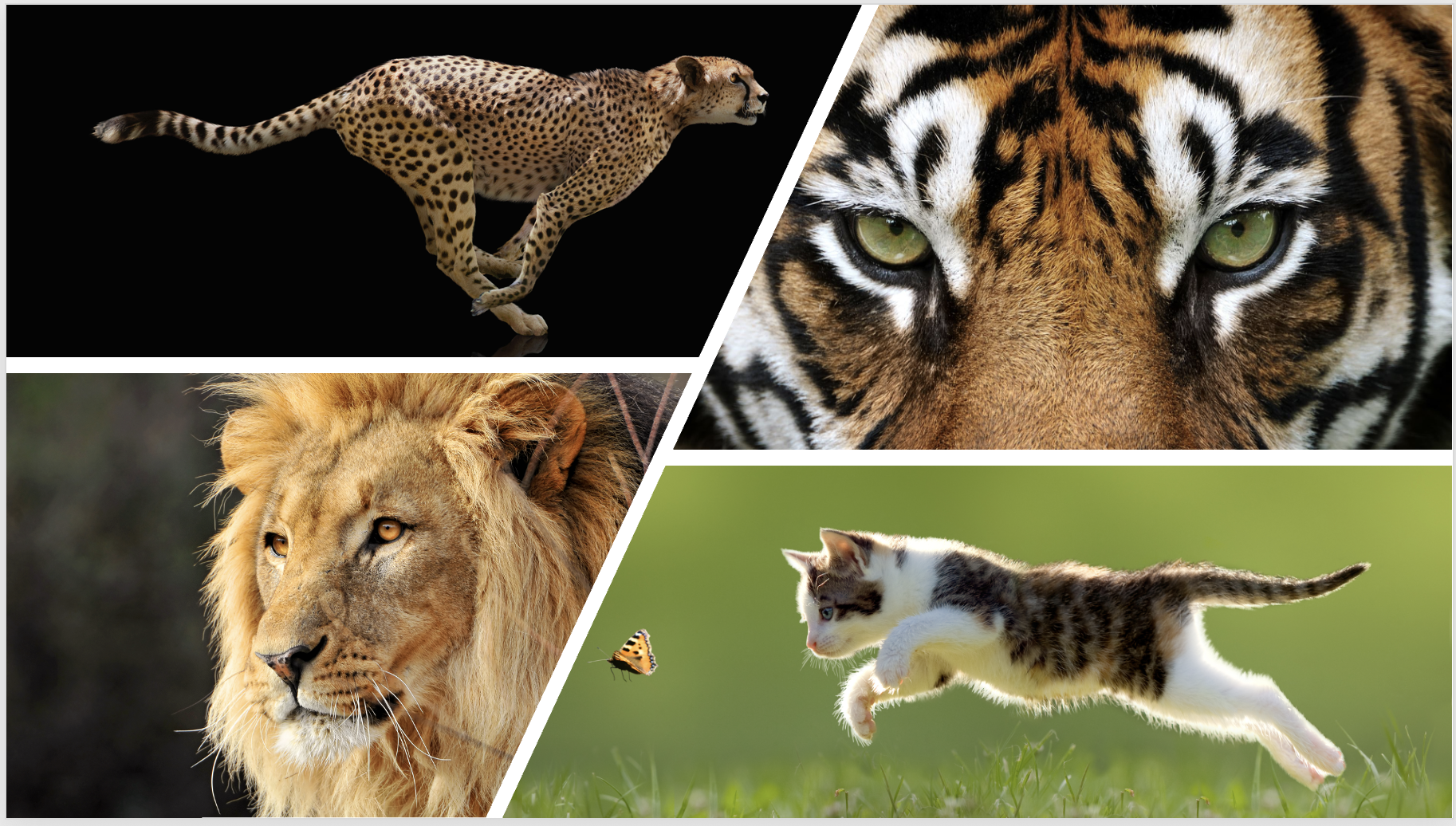}\\[0.2em]
        \includegraphics[width=\linewidth]{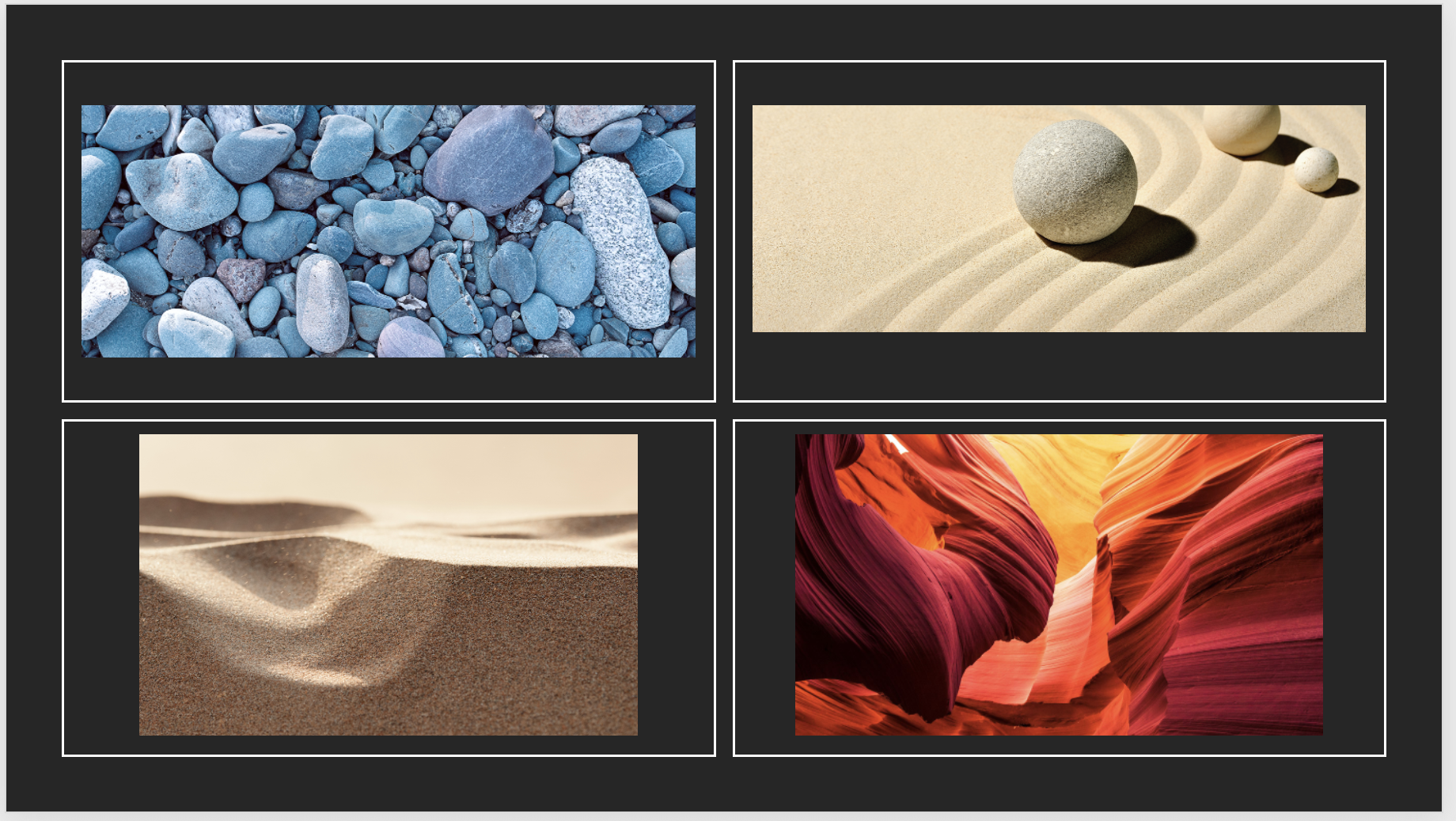}\\[-0.1em]
        \textit{\scriptsize Ground Truth}
    \end{minipage}

    \caption{Cases from our PPTArena Benchmark showing significant visual complexity. The left columns illustrate specific tasks (layout, sizing, charts, posters), while the right column demonstrates a complex multi-slide collage task where multiple assets are harmonized into a cohesive theme.}
    \label{fig:moreimages_cases}
\end{figure*}

\onecolumn

\section*{Appendix II Code Listings}
\addcontentsline{toc}{section}{Appendix II Code Listings}

\nolinenumbers

\noindent
\begin{lstlisting}[caption={PPTX to JSON Conversion Logic}, label={lst:pptx_to_json}, language=Python]
def pptx_to_json(filepath):
    """
    Converts a .pptx file to a comprehensive JSON representation.
    Captures all edit types: Content, Layout, Styling, Interactivity, and Structure.
    """
    try:
        prs = Presentation(filepath)
        
        # Presentation-level metadata
        presentation_data = {
            "filename": os.path.basename(filepath),
            "slide_width": prs.slide_width.pt if hasattr(prs, 'slide_width') else None,
            "slide_height": prs.slide_height.pt if hasattr(prs, 'slide_height') else None,
            "slides": []
        }
        
        # ... [Metadata logic omitted for brevity] ...
        
        for i, slide in enumerate(prs.slides):
            slide_data = {
                "slide_number": i + 1,
                "shapes": [],
                "notes": "",
                # Layout & Structure info
                "slide_layout": slide.slide_layout.name if hasattr(slide, 'slide_layout') else None,
                "slide_id": slide.slide_id if hasattr(slide, 'slide_id') else None,
            }
            
            # ... [Shape iteration and property extraction logic omitted for brevity] ...
            # ... [Captures: Background, Shapes, Text, Tables, Images, Charts, Groups] ...
            
            presentation_data["slides"].append(slide_data)
            
        return presentation_data
    except Exception as e:
        print(f"Error processing {filepath}: {e}")
        return None
\end{lstlisting}

\noindent
\begin{lstlisting}[caption={Smart Diff Construction}, label={lst:smart_diff}, language=Python]
def diff_pptx_json(ground_truth_json, prediction_json, initial_json=None):
    """
    Performs a deep comparison between ground_truth and prediction JSON structures.
    Returns a structured diff with only the differences, organized by slide and shape.
    """
    differences = []
    
    # ... [Helper functions: normalize_value, values_match, compare_dict, compare_list omitted] ...
    
    # Compare slides
    gt_slides = ground_truth_json.get("slides", [])
    pred_slides = prediction_json.get("slides", [])
    init_slides = initial_json.get("slides", []) if initial_json else []
    
    for slide_idx in range(max(len(gt_slides), len(pred_slides))):
        # ... [Slide comparison logic omitted] ...
        pass
    
    # Calculate similarity score
    total_properties = len(differences) + 100  # Baseline to avoid division by zero
    similarity_score = 1.0 - (len(differences) / total_properties)
    
    return {
        "has_differences": len(differences) > 0,
        "similarity_score": max(0.0, min(1.0, similarity_score)),
        "total_differences": len(differences),
        "differences": differences
    }
\end{lstlisting}

\clearpage

\noindent
\begin{lstlisting}[caption={Style Target Generation Prompt}, label={lst:style_prompt}]
You are an expert technical writer for presentation editing workflows.
Given ONLY the Original deck and the Ground Truth deck, produce actionable, specific instructions to convert Original into Ground Truth. Focus on content and structure inferred from JSON; do not reference any predictions.

--- ORIGINAL (JSON summary) ---
```json
{original_ppt_json_truncated}
```

--- GROUND TRUTH (JSON summary) ---
```json
{ground_truth_ppt_json_truncated}
```

Return a single JSON object with keys: overview_instructions (multi-sentence, stepwise where helpful), and notes (optional).
\end{lstlisting}
\endgroup

\clearpage

\nolinenumbers
\noindent
\begin{lstlisting}[caption={Full Visual Quality Judge Prompt}, label={lst:vq_prompt}]
You are a judge of VISUAL/CONTENT QUALITY and PRESERVATION.

CRITICAL UNDERSTANDING:
- The "Instruction" is what the model/editor received
- The "Style Target" is YOUR evaluation rubric - the model DID NOT see this
- Ground Truth is ONE valid example - accept other valid visual approaches
- Focus on SEMANTIC correctness: Are the visual elements correct? No overlap? Readable?
- Exact positions/sizes don't matter unless the Instruction explicitly requires them
- You will only receive Ground Truth and Prediction slide images; compare them directly slide-by-slide.
- Use any provided Style Target guidance to check required visual cues strictly.

FLEXIBILITY:
- Different layouts achieving the same goal are acceptable (e.g., list vs grid)
- Small position variations are fine if elements are clear and non-overlapping
- Theme colors may vary slightly as long as they're harmonious
- "Approximately centered" or "well-aligned" is acceptable without pixel-perfection

HARSH SCORING POLICY (very strict):
- Penalize any unintended visual change to non-requested content (fonts, sizes, colors, positions, shapes, charts, images, tables, or slide structure).
- Choose the lower score when uncertain between adjacent scores.

VISUAL_Quality score (0-5):
- 5: No unintended changes to non-requested content; fonts, sizes, colors, positions, and objects match Ground Truth; structure preserved.
- 4: Visually very close to the Ground Truth; only imperceptible or negligible differences (e.g., sub-pixel alignment); no style drift.
- 3: Minor but noticeable visual differences (e.g., slight font weight/size/spacing shifts) without breaking layout.
- 2: Clear deviations (e.g., wrong fonts/sizes/colors, noticeable position shifts) or small layout issues.
- 1: Major deviations (overlap, off-canvas, broken layout) but still legible.
- 0: Severely broken or unreadable slide.

Output a single JSON object with:
- visual_quality_score (0-5)
- visual_quality_reason (one sentence, specific evidence about visual differences)

--- User Instruction ---
{instruction_text}

--- Style Target (judge rubric, unseen by the model under evaluation) ---
{style_block}

You are given two labeled image sequences:
- Ground Truth: the correct target deck to match
- Prediction: the candidate deck produced by the system

CRITICAL: Ground Truth is ONE valid example, not the only correct answer.
Judge if Prediction achieves the SEMANTIC INTENT shown by Ground Truth.
Accept different valid layouts/arrangements that fulfill the instruction and style target.
Focus on: correctness, no overlap, readability, theme consistency.
Small position/size variations are acceptable if elements are clear.

Judge visual quality by comparing PREDICTION to GROUND TRUTH.

Return only:
- visual_quality_score (0-5)
- visual_quality_reason (one sentence, specific evidence about visual differences)
\end{lstlisting}
\linenumbers

\clearpage
\nolinenumbers
\noindent
\begin{lstlisting}[caption={Full Instruction Following Judge Prompt}, label={lst:if_prompt}]
You are a strict judge of INSTRUCTION FOLLOWING.

CRITICAL UNDERSTANDING:
- The "Instruction" is what the model/editor received (the user's request)
- The "Style Target" is YOUR evaluation rubric - the model DID NOT see this
- You will receive a FOCUSED DIFF showing only what changed between ground_truth and prediction
- Your job: Judge if the prediction's changes match the ground_truth's changes
- DO NOT compare prediction to initial - focus on whether prediction achieved ground_truth's outcome
- If the diff shows minimal differences, that's GOOD (high score)
- Ground Truth is ONE valid example, not the only correct answer

FLEXIBILITY:
- Accept different valid approaches (e.g., flags in a list vs rows is fine if they match the text)
- Exact positions/sizes don't matter unless the Instruction explicitly requires them
- Very small measurement variations ($\pm$1%) are acceptable for fonts/sizes due to rounding
- Z-order (layering) differences ARE significant and should be noted
- Focus on semantic properties: text content, font names, colors, structural changes, z-order

HARSH SCORING POLICY (very strict):
- Choose the lower score when uncertain between adjacent scores.
- For translation/summarization or other text edits requiring reasoning, semantic similarity is more important than exact wording.

INSTRUCTION_FOLLOWING score (0-5):
- 5: Every requested object/change exists and is exactly correct; nothing requested is missing or misapplied; no extra edits beyond the instruction.
- 4: All requested changes exist and are mostly correct; only a tiny inaccuracy that does not affect meaning.
- 3: Most requested changes exist but at least one is incomplete, incorrect, or missing detail.
- 2: Only some requested changes exist; notable misses or incorrect applications.
- 1: Requested changes largely not performed or substantially incorrect.
- 0: Contradicts or ignores the instruction entirely.

Output a single JSON object with:
- instruction_following_score (0-5)
- instruction_following_reason (one sentence, specific evidence comparing prediction to ground_truth)

--- USER INSTRUCTION (what the model received) ---
{instruction_part}

--- STYLE TARGET (your evaluation rubric - the model DID NOT see this) ---
{style_target_part}

--- SMART DIFF ANALYSIS (Prediction vs Ground Truth) ---
{formatted_diff}

CRITICAL COMPARISON INSTRUCTIONS:
The diff above shows ONLY the differences between prediction and ground_truth.
- If the diff shows "No differences" $\rightarrow$ Perfect match $\rightarrow$ Score 5
- If the diff shows differences in properties that the instruction requires $\rightarrow$ Score based on correctness
- Focus on whether prediction achieved the same semantic outcome as ground_truth

REMINDER: Judge if the prediction achieved the SEMANTIC INTENT of the Instruction.
The diff highlights what actually changed - use this to make an accurate judgment.
\end{lstlisting}

\end{document}